\documentclass{article}

\usepackage{arxiv}

\usepackage[utf8]{inputenc} 
\usepackage[T1]{fontenc}    
\usepackage{hyperref}       
\usepackage{url}            
\usepackage{booktabs}       
\usepackage{nicefrac}       
\usepackage{microtype}      

\usepackage{amsmath,amssymb,amsfonts}
\usepackage{algorithmic}
\usepackage{graphicx}

\usepackage{amsmath}
\usepackage{multirow}
\usepackage{booktabs}
\usepackage{color}
\usepackage{colortbl}
\usepackage{breqn}
\usepackage{xcolor}

\newcommand\R{\mathbb{R}}

\newcommand\numberthis{\addtocounter{equation}{1}\tag{\theequation}}

\newcommand{\Th}{\boldsymbol{\theta}}
\newcommand{\ggamma}{\boldsymbol{\gamma}}
\newcommand{\ppi}{\boldsymbol{\pi}}

\newcommand{\h}{\mathbf{h}}
\newcommand{\q}{\mathbf{q}}
\newcommand{\s}{\mathbf{s}}
\renewcommand{\u}{\mathbf{u}}
\renewcommand{\v}{\mathbf{v}}
\newcommand{\y}{\mathbf{y}}

\newcommand{\tr}{\text{Tr}}
\newcommand{\argmin}{\text{argmin}}

\newcommand{\beq}{\begin{equation}}
\newcommand{\eeq}{\end{equation}}

\title{Two-block vs. Multi-block ADMM: \\ An empirical evaluation of convergence}

\author{
  Andre Goncalves\thanks{Corresponding author} \\
  Lawrence Livermore National Laboratory\\
  Livermore, CA, USA\\
  \texttt{andreric.goncalves@gmail.com} \\
   \And
 Xiaoli Liu\\
  College of Humanities and Sciences\\
  Northeast Normal University\\
  Changchun, China \\
  \texttt{neuxiaoliliu@gmail.com} \\
   \And
   Arindam Banerjee \\
   Department of Computer Science and Engineering\\
   University of Minnesota, Twin Cities, USA \\
   \texttt{banerjee@cs.umn.edu}
}

\begin{document}
\maketitle

\begin{abstract}
Alternating Direction Method of Multipliers (ADMM) has become a widely used optimization method for convex problems, particularly in the context of data mining in which large optimization problems are often encountered. ADMM has several desirable properties, including the ability to decompose large problems into smaller tractable sub-problems and ease of parallelization, that are essential in these scenarios. The most common form of ADMM is the two-block, in which two sets of primal variables are updated alternatingly. Recent years have seen advances in multi-block ADMM, which update more than two blocks of primal variables sequentially. In this paper, we study the empirical question: {\em Is two-block ADMM always comparable with sequential multi-block ADMM solving an equivalent problem?} In the context of optimization problems arising in multi-task learning, through a comprehensive set of experiments we surprisingly show that multi-block ADMM consistently outperformed two-block ADMM on optimization performance, and as a consequence on prediction performance, across all datasets and for the entire range of dual step sizes. Our results have an important practical implication: rather than simply using the popular two-block ADMM, one may considerably benefit from experimenting with multi-block ADMM applied to an equivalent problem.	
\end{abstract}

\keywords{Alternating Direction Method of Multipliers \and  Two-block ADMM  \and Multi-block ADMM \and Convergence Analysis \and Nonconvex optimization \and Multitask Learning}

\section{Introduction}
\label{sec:intro}

The past decade has seen wide ranging applications of the alternating direction method of multipliers (ADMM) for different learning problems \cite{Boyd2011,Wang2014,ye2011split,wahlberg2012admm}. The success of ADMM in the context of data mining can be attributed to its ability to decouple large problems into smaller tractable sub-problems, simplicity of the updates (sometimes in closed form) for these individual sub-problems, ease of parallelization, among others. ADMM has also been widely applied to unconstrained non-smooth problems, such as the lasso and fused lasso, by introducing additional variables (lifting) thereby decoupling smooth and non-smooth terms \cite{Yang2011,wahlberg2012admm}.

The most popular version of ADMM is the two-block ADMM which considers the following problem:
\beq
\min_{x,z}~ f(x) + g(z) \quad \text{s.t.} \quad Ax + B z = c~,
\eeq
where $x \in \R^n, z \in \R^m, A \in \R^{d \times n}, B \in \R^{d \times m}, c \in \R^d$ \cite{Deng2017,He2012,hu2013efficient,das2013online}.
The formulation is called two-block because of the two sets of variables $(x,z)$ which are updated
alternatingly in the primal updates of ADMM. Recent years have seen the use of multi-block ADMM
in some settings  \cite{Deng2017,Wang2014b,Chen2016,wang2015global}, which considers more than two sets of variables which are sequentially
updated in the primal. The literature has also looked at potential challenges in doing naive multi-block ADMM, and several approaches to doing it with theoretical guarantees of convergence have been developed \cite{wang2015convergence,lin2015sublinear,wang2015global}.

Two-block ADMM is by far the most widely used variant of ADMM. In practice, often two-block ADMM is used without much thought regarding alternative multi-block ADMM options. For convex problems, they are supposed to give the same solution after all, with suitable choices of (dual) step sizes. Typically, a multi-block problem can be solved as a two-block problem with
suitable variable and constraint grouping, and similarly two-block problems can be converted into multi-block by suitable lifting \cite{Deng2017,Wang2014b,wang2013solving}.
Our work is motivated by a desire to understand why two-block ADMM is preferred. A part of the reason may be the seemingly negative theoretical results
associated with convergence of multi-block ADMM \cite{Chen2016}. In many cases, such negative theoretical results focused on a specific problem or operating regimes where multi-block ADMM diverges. 
In practice, most optimization methods have such weak spots in terms of struggling in a type of problem or operating regime, e.g., the simplex algorithm for linear programs \cite{klee1970good}, stochastic gradient
descent (SGD) for deep networks \cite{kingma2014adam}, etc. In many cases, such algorithms are empirically successful and it takes a while to improve our theoretical understanding of its behavior,
e.g., smoothed analysis of the simplex algorithm \cite{Spielman2004}, convergence of SGD on non-convex problems \cite{du_gradient_2018,du_gradient_2018-1,zou_stochastic_2018,brutzkus_sgd_2017}, etc. Such perspective motivates our empirical work on two-block vs.~multi-block ADMM comparisons.

In particular, in this paper we study the empirical question: {\em Is two-block ADMM always comparable with multi-block ADMM solving an equivalent problem?} 
For both two-block and multi-block ADMM, we consider linearized ADMM \cite{Wang2014} which has considerably computational benefits and keep the (dual) step size as the only hyper-parameter. Also, for multi-block ADMM,
we use the direct extension from two-block ADMM which sequentially updates the primal variables \cite{Boyd2011}.    
We perform experiments using optimization problems arising in multi-task learning (MTL), with application focus on modeling Alzheimer's disease progression, Parkinson's disease assessment, and air quality prediction. For each problem, we conduct a direct comparison between two-block and multi-block ADMM by varying one hyper-parameter---the dual step-size in ADMM (see Section~\ref{sec:admm} and \ref{sec:expts} for details)-- up to 5 orders of magnitude, running each method for 1,000 iterations, and for 10,000 iterations in some cases, and repeating each experiment 10 times to account for any inherent variability. The approaches are evaluated based on different evaluation measures including primal residual convergence and prediction performance on validation set (see Section~\ref{sec:expts}). \footnote{The experimental setup and code will be available in {\tt github}, and a reference will be added in the final version.}

Our results are surprising: multi-block ADMM empirically outperformed two-block ADMM on optimization performance, and as a consequence on prediction performance, across all datasets and for the entire range of dual step sizes. For optimization performance, multi-block ADMM achieves primal residuals which are a few orders of magnitude smaller than that achieved by two-block ADMM, implying a better convergence to the feasible set. Note that primal iterates in any form of ADMM are always outside the feasible set, so having a really small primal residual is important to ensure good convergence, and multi-block ADMM illustrates desirable performance here. Multi-block ADMM also achieves smaller primal objective, which is especially interesting when primal residuals are small. Finally, the models estimated using multi-block ADMM consistently outperforms the corresponding two-block versions on prediction performance evaluated on validation sets.

The take-away from the results reported is the following: if a situation calls for ADMM based optimization, rather than using the popular two-block version by default, one needs to try out equivalent multi-block versions and compare their performance. The results reported do not imply that multi-block ADMM will outperform two-block ADMM on every problem, but that one needs to at least consider multi-block as a serious alternative rather than defaulting to the two-block version due to its popularity. Thus, the contribution of the work is empirical, hoping to change the way our community uses ADMM based algorithms.

The rest of the paper is organized as follows. We review related work in Section~\ref{sec:related}. We present the MTL model in Section~\ref{sec:flsgl} and discuss two-block and multi-block ADMM for learning the model in Section~\ref{sec:admm}. We present detailed results comparing the algorithms in Section~\ref{sec:expts} and conclude in Section~\ref{sec:conc}.


\section{Related Work}
\label{sec:related}

The success of ADMM with two blocks has encouraged the natural extension of ADMM 
to solve problems with multiple blocks \cite{wang2013solving,Wang2014b,Hong2014,Hong2017,Deng2017}. 
Theoretical results \cite{Chen2016} have shown that without additional conditions, the direct extension of ADMM with more than two blocks generally fails to converge. But for certain specific classes of problems and/or conditions, researchers have been able to show convergence guarantees for variants of multi-block ADMM. The existing variants of multi-block ADMM can be mainly grouped into two categories: Gauss-Seidel ADMM and Jacobi ADMM.

The Gauss-Seidel ADMM (GSADMM) \cite{Hong2017,Hong2014} seeks to solve the primal updates in a cyclic block coordinate fashion. 
In \cite{He2012}, a back substitution step was added, which facilitated the convergence proof for ADMM with multiple blocks. In \cite{Hong2014}, a block successive upper bound minimization method of multipliers (BSUMM) is proposed to solve the problem. In \cite{Lin2016}, the authors proved that multi-block ADMM can achieve $\epsilon$-optimal solutions within $O(1/\epsilon^2)$ iterations by solving a modified version of the original problem, and whitin $O(1/\epsilon)$ iterations when applied to certain  \textit{sharing} problems conditioned on properties of the augmented Lagrangian function. Theoretical convergence of some of these methods need fairly strict conditions, e.g., certain local error bounds need to hold and/or the step size needs to be sufficiently small or decreasing.  

Another broad approach to multi-block ADMM is Jacobi ADMM \cite{wang2013solving,Deng2017,parikh2014}, which solves 
the problem in a parallel block coordinate fashion. In \cite{wang2013solving,parikh2014}, the problem
is solved by using two-block ADMM with splitting variables (sADMM). \cite{Deng2017} considers a proximal
Jacobian ADMM (PJADMM) by adding proximal terms. A randomized block coordinate variant
of ADMM named RBSUMM was proposed in \cite{Hong2014}. Recently, \cite{Hong2016} demonstrated that ADMM is convergent regardless of the number of variable blocks for certain nonconvex {\it consensus} and {\it sharing} problems.

Although general theoretical aspects of multi-block ADMM are still not well understood, it has been observed that modified versions of ADMM, including the two-block instance, though with convergence guarantee, often perform slower than the multi-block ADMM with no convergent guarantee \cite{Lin2016}. It is thus of great interest to further study the practical aspects of convergence of multi-block ADMM. Aiming to address the lack of numerical experiments, we systematically investigate the convergence and performance of two-block and multi-block ADMM in the context of multitask learning, which has significantly benefited from advances in ADMM-type of methods.

\section{Temporally Smooth Multi-task Learning (TS-MTL)}
\label{sec:flsgl}
Consider a multi-task learning problem over $T$ tasks, where each task corresponds to a time point $t = 1,\ldots,T$.
For each time point $t$, we consider a regression task based on data $(\y_t,X_t)$, where $X_t \in \R^{n \times p}$ denotes the matrix of covariates,  $p$ is the number of covariates and $n$ is the number of samples, shared across all the tasks, and $\y_t \in \R^n$ is the matrix of responses. Let $\Theta \in \R^{p \times T}$ denote the regression parameter matrix over all tasks, so that column $\Th_t \in \R^p$ corresponds to the parameters for the task in time step $t$. This multitask learning formulation is based on our recent work \cite{Liu:2018:MAD}.
%
%

We pose the multi-task learning problem for $\Theta$ such that two goals are accomplished: each $\theta_t$ accomplishes low regression error for each task $t$, and ``nearby'' $\theta_t$ are coupled to be similar, since the ``nearby'' tasks are temporarily related. The notion of ``nearby'' needs to be suitably defined. 
In this paper, we adopt an approach inspired by non-parametric regression~\cite{Wasserman2006}, which has been recently shown to be effective for TS-MTL problems. In particular, we model a local approximation to $\theta_t$ in terms of the other $\theta_{\ell}, \ell \neq t$ as
\begin{align*}
\hat{\Th}_t = \sum_{\substack{\ell=1 \\ \ell \neq t}}^T w_{\ell,t} \Th_{\ell}, ~\forall t, 
w_{\ell,t} = \frac{\exp\left(-\frac{(\ell-t)^2}{\sigma^2}\right)}{\sum_{\substack{\ell'=1\\ \ell'\neq t}}^T \exp\left(-\frac{(\ell'-t)^2}{\sigma^2}\right)}~, \forall \ell\neq t, 
\label{eq:gaussLap}
\end{align*}
where $\sigma \geq 0$ is a constant bandwidth parameter. Based on such an approximation, our TS-MTL formulation focus on encouraging sparsity of the residual
\beq
\Th_t - \hat{\Th}_t = \Th_t - \sum_{\substack{\ell=1 \\ \ell \neq t}}^T w_{\ell,t} \Th_{\ell} ~, ~\forall t~.
\eeq
The TS-MTL problem can now be posed as the following unconstrained problem:
\begin{equation}
\begin{split}
\underset{\Theta,\Gamma}{\text{min}}~ \sum_{t=1}^T & \|\y_t - X_t \Th_t\|^2 + R_{\lambda_2}^{\lambda_1} (\Theta) + \lambda_3  \sum_{t=1}^T \| \Th_t - \sum_{\substack{\ell=1 \\ \ell \neq t}}^T w_{\ell,t} \Th_{\ell} \|_1~,
\label{eq:Formulation}
\end{split}
\end{equation}
where $R_{\lambda_2}^{\lambda_1}(\Theta)$ is the combination of lasso and group lasso penalties, also known as the sparse group Lasso penalty, which allows simultaneous joint feature selection for all tasks and selection of a specific set of features for each task \cite{FoGLasso}. In particular,
\beq
R_{\lambda_2}^{\lambda_1} (\Theta) = \lambda_1 \| \Theta \|_1 + \lambda_2 \| \Theta \|_{2,1}~,
\eeq
where $\| \Theta \|_1$ is the Lasso penalty and $\| \Theta \|_{2,1} = \sum_{j=1}^p \| \Th_j \|, \Th_j \in \R^T$ is the group Lasso penalty considering groups across time for each feature $j$, encouraging the regression models at different time points to share a common set of features. In the formulation, $\lambda_1, \lambda_2, \lambda_3 > 0$ are the regularization parameters which are fixed, and will be chosen using cross validation.

\section{ADMM for Learning TS-MTL Models}
\label{sec:admm}
The unconstrained optimization problem in \eqref{eq:Formulation} can be difficult to optimize directly due to the non-smooth terms. In this section, we consider two formulations, respectively solved as a two-block and multi-block ADMM, which introduce additional variables and associated linear constraints. While the use of two-block or multi-block ADMM for solving the problem is not especially novel, through extensive experiments in Section~\ref{sec:expts} we will show that the two formulations and associated ADMM algorithms behave surprisingly differently in practice.


We start by noting that the optimization problem in \eqref{eq:Formulation} can be formulated as the following linearly constrained optimization problem:
\begin{equation}
\begin{aligned}
& \underset{\Theta,\Gamma,Q,\Pi}{\text{min}}
& & \sum_{t=1}^T \frac{1}{2} \|\y_t - X_t \Th_t\|^2 + R_{\lambda_2}^{\lambda_1} (Q) + \lambda_3 \| \Pi \|_1\\
& \text{subject to}
& &  \Theta - Q = 0,~\q_t - \sum_{\substack{\ell=1 \\ \ell\neq t}}^T w_{|t-\ell|} \q_{\ell} - \ggamma_t = 0~\forall t,~ \Gamma - \Pi = 0~.
\end{aligned}
\label{eq:constrained2}
\end{equation}

Note that there are four variables $\Theta,\Gamma,Q,\Pi \in \R^{p \times T}$ in the formulation.
The problem can also be formulated as
\begin{equation}
\begin{aligned}
& \underset{\Theta,\Gamma,Q,\Pi}{\text{min}}
& & ~~\sum_{t=1}^T \frac{1}{2} \|\y_t - X_t \Th_t\|^2 + R_{\lambda_2}^{\lambda_1} (Q) + \lambda_3 \| \Pi \|_1 \\
& \text{subject to}
& &  \Theta - Q = 0,~ \Th_t - \sum_{\substack{\ell=1 \\ \ell\neq t}}^T w_{|t-\ell|} \Th_{\ell} - \ggamma_t = 0 ~\forall t,~ \Gamma - \Pi = 0~.
\end{aligned}
\label{eq:constrained3}
\end{equation}

The only difference between \eqref{eq:constrained2} and \eqref{eq:constrained3} is the form of the constraint associated with temporal smoothness residuals:
\beq
\ggamma_t = \q_t - \sum_{\substack{\ell=1 \\ \ell\neq t}}^T w_{|t-\ell|} \q_{\ell}~
 \quad \text{vs.} \quad \ggamma_t = \Th_t - \sum_{\substack{\ell=1 \\ \ell\neq t}}^T w_{|t-\ell|} \Th_{\ell} ~,
\label{eq:ts_const}
\eeq
but the constraints are equivalent since $\Theta = Q$.



\subsection{Linearized Two-Block ADMM}
\label{ssec:lin2b-admm}
The augmented Lagrangian $L_{\rho} = L_{\rho}(\Theta, \Gamma, Q, \Pi, S, U, V)$ for \eqref{eq:constrained2} is given by:


		\begin{align*}
		L_{\rho} = & \sum_{t=1}^T \frac{1}{2} \|\y_t - X_t \Th_t\|^2 + R_{\lambda_2}^{\lambda_1} (Q) + \lambda_3 \| \Pi \|_1  +  \tr(S^T (\Theta - Q))  \\
		& + \frac{\rho}{2} \| \Theta - Q \|^2 +   \tr(V^T (\Gamma - \Pi)) + \frac{\rho}{2} \| \Gamma - \Pi \|^2\\
		& + \sum_{t=1}^T \left\{ \u_t^T \left(\q_t - \sum_{\substack{\ell=1 \\ \ell\neq t}}^{T} z_{t\ell} \right) + \right.  \left. \frac{\rho}{2} \left\| \q_t - \sum_{\substack{\ell=1 \\ \ell\neq t}}^{T}z_{t\ell} \right\|^2 \right\} \\
		\end{align*}

\noindent where $z_{t\ell} =  w_{|t-\ell|} \q_{\ell} - \ggamma_t$, and $S, U, V$ represent Lagrange multipliers.
The optimization problem can be solved as a two-block ADMM with two sets of primal variables: $Z_1 = (\Theta,\Gamma)$, since $\Theta$ and $\Gamma$ can be updated in parallel, and $Z_2 = (Q,\Pi)$ since $Q$ and $\Pi$ can be updated independently.

Let
$h(\Theta) = \frac{\rho}{2} \sum_{t=1}^T  \| \Th_t - \sum_{\substack{\ell=1 \\ \ell\neq t}}^T w_{|t-\ell|} \Th_{\ell} - \ggamma_t \|^2$.
While the original objective function does not have terms with interactions between $\Th_t$ and $\Th_{\ell}, \ell \neq t$, the augmented Lagrangian do have such terms, and such interactions are captured in $h(\Theta)$. In order to decouple the $\Th_t$ updates, we will perform the ADMM updates by suitably linearizing $h(\Theta)$ around the current iterate $\Theta^k$. Let $\h_t^k = \nabla_{\Th_t} h(\Theta^k)$ denote the gradient with respect to $\Th_t$. Recent work on Bregman ADMM \cite{Wang2014} and related work on inexact ADMM \cite{Yang2011,Boyd2011} have shown that ADMM updates with such linearization continue to work with the same rates of convergence.


{\bf Update $\Th_t^{k+1}$:} The update involves solving the following quadratic objective, which can be done efficiently using Cholesky decomposition as discussed in \cite{Boyd2011}:
\beq
\Th_t^{k+1} = \underset{\Th_t}{\argmin}~ \frac{1}{2} \|\y_t - X_t \Th_t\|^2 + (\s_t^k)^T \Th_t  + \frac{\rho}{2} \| \Th_t - \q_t \|^2 ~.
\label{eq:theta_twob}
\eeq

{\bf Update $\ggamma_t^{k+1}$:} With $c_t^k = (\ppi_t^k + \q_t^{k} - \sum_{\substack{\ell=1 \\ \ell \neq t}}^T w_{|t-\ell|} \q_{\ell}^{k} )$, the update for $\ggamma_t$ needs to solve:
\beq
\ggamma_t^{k+1} = \underset{\ggamma_t}{\argmin}~ \frac{\rho}{2} \left\| \ggamma_t -  c_t^k \right\|^2 - (\u_t^k - \v_t^k)^T \ggamma_t ~.
\label{eq:gamma_twob}
\eeq

{\bf Update $Q$:} The update for $Q$ is based on linearization and we need to compute the proximal operator for $R_{\lambda_1}^{\lambda_2}(\cdot)$:
\begin{align*}
Q^{k+1}  =  \underset{Q}{\argmin}~ &  \Bigg( \Bigg. \frac{\rho}{2} \| Q - \Theta^{k+1} \|^2 - \tr((S^k + H^k + A^k)^T Q) + \frac{\rho_1}{2} \| Q - Q^k \|^2 + R_{\lambda_2}^{\lambda_1} (Q)~ \Bigg. \Bigg), \numberthis
\label{eq:q_twob}
\end{align*}
where $A^k$ is such that $\tr((A^k)^T Q) = \sum_{t=1}^T (\u_t^k)^T (\q_t - \sum_{\ell \neq t} w_{|t-\ell|} \q_{\ell})$, and $\rho_1 > 0$ is a suitably chosen constant. In particular, since $h(Q)$ is smooth and has Lipschitz continuous gradients say with constant $\nu$ under 2-norm, it suffices to have $\rho_1 \geq 2 \nu$ \cite{Wang2014}.
The expression can be simplified to get in the form of a proximal operator computation for $R_{\lambda_1}^{\lambda_2}(Q)$.

{\bf Update $\Pi$:} For the update $\Pi$, we need to compute the proximal operator for $L_1$-norm:
\begin{equation}
\Pi^{k+1} = \underset{\Pi}{\text{argmin}}~ \frac{\rho}{2} \| \Pi - \Gamma^{k+1} \|^2 - \tr((V^k)^T \Pi) + \lambda_3 \| \Pi \|_1~.
\label{eq:pi_twob}
\end{equation}


{\bf Dual Updates $S, U, V$:} Following standard ADMM dual updates, the updates for the dual variables for our setting are as follows:
\begin{align}
S^{k+1} &  = S^k + \rho (\Theta^{k+1} - Q^{k+1}) \\
\u_t^{k+1} & = \u_t^k + \rho(\q_t^{k+1} - \sum_{\substack{\ell=1 \\ \ell \neq k}}^T w_{|t-\ell|} \q_{\ell}^{k+1} - \ggamma_t^{k+1})~, ~\forall t \\
V^{k+1} & = V^k + \rho (\Gamma^{k+1} - \Pi^{k+1})~.
\end{align}

\subsection{Linearized Multi-Block ADMM}
\label{ssec:linmb-admm}
The augmented Lagrangian $L_{\rho} = L_{\rho}(\Theta, \Gamma, Q, \Pi, S, U, V)$ for \eqref{eq:constrained3} is given by:

\begin{align*}
L_{\rho} = & \sum_{t=1}^T \frac{1}{2} \|\y_t - X_t \Th_t\|^2 + R_{\lambda_2}^{\lambda_1} (Q) + \lambda_3 \| \Pi \|_1 + \tr(S^T (\Theta - Q)) \\
				& +  \frac{\rho}{2} \| \Theta - Q \|^2 +  \tr(V^T (\Gamma - \Pi)) + \frac{\rho}{2} \| \Gamma - \Pi \|^2 \\
				& + \sum_{t=1}^T \left\{ \u_t^T \left(\Th_t - \sum_{\substack{\ell=1 \ell\neq t}}^{T} z_{t\ell}\right) + \right. \left.  \frac{\rho}{2} \left\| \Th_t - \sum_{\substack{\ell=1 \ell\neq t}}^{T} z_{t\ell} \right\|^2 \right\}~,
\end{align*}
where $z_{t\ell} = w_{|t-\ell|} \Th_{\ell} - \ggamma_t$.

The optimization problem can be solved as a multi-block ADMM with 3 sets of primal variables: $Z_1 = \Theta$, $Z_2 = \Gamma$, and $Z_3 = (Q,\Pi)$. Unlike the two-block case earlier, $\Theta$ and $\Gamma$
here are coupled, and hence the updates need to be sequential, i.e., updates to $\Theta$, followed by
updates to $\Gamma$, followed by updates to $(Q,\Pi)$ which can be done in parallel. Note that one can stack $(\Theta,\Gamma)$ into a long vector and update them jointly by solving a high-dimensional quadratic
objective. We do not consider such a two-block formulation because we have already considered a two-block
formulation which is arguably simpler, since the quadratic objective there only involves $\Theta$.
As before, we work with a linearized version of $h(\Theta)$ for the updates.


{\bf Update $\Th_t^{k+1}$:} The update involves solving the following unconstrained quadratic objective:
\begin{align*}
\Th_t^{k+1} = \underset{\Th_t}{\argmin} ~ & \Bigg( \Bigg. \frac{1}{2} \|\y_t - X_t \Th_t\|^2 + (\s_t^k)^T \Th_t  + \frac{\rho}{2} \| \Th_t - \q_t \|^2     + (\u_t^k + \h_t^k)^T \Th_t + \frac{\rho_1}{2} \| \Th_t - \Th_t^k \|^2 \Bigg. \Bigg) ~ \numberthis \label{eq:theta_mblock},
\end{align*}
where $\rho_1 > 0$ is a suitably chosen constant.
In particular, since $h(\Theta)$ is smooth and has Lipschitz continuous gradients with constant $\nu$ under 2-norm, it suffices to have $\rho_1 \geq 2 \nu$ \cite{Wang2014}.

{\bf Update $\ggamma_t^{k+1}$:} With $c_t^{k+1} = (\ppi_t^k + \Th_t^{k+1} - \sum_{\substack{\ell =1 \\ \ell \neq t}}^{T} w_{|t-\ell|} \Th_{\ell}^{k+1} )$, the update for $\ggamma_t$ needs to be solve:
\beq
\ggamma_t^{k+1} = \underset{\ggamma_t}{\argmin}~ \frac{\rho}{2} \left\| \ggamma_t - c_t^{k+1} \right\|^2 - (\u_t^k - \v_t^k)^T \ggamma_t ~.
\label{eq:gamma_mblock}
\eeq

{\bf Update $Q$:} The update for $Q$ effectively needs to compute the proximal operator
 for $R_{\lambda_1}^{\lambda_2}(\cdot)$ \cite{Yu2013,Yu2013b}:
\beq
Q^{k+1} =  \underset{Q}{\argmin}~ \frac{\rho}{2} \| Q - \Theta^{k+1} \|^2 - (S^k)^T Q + R_{\lambda_2}^{\lambda_1} (Q)~.
\label{eq:q_mblock}
\eeq

{\bf Update $\Pi$:} The update for $\Pi$ needs to solve:
\beq
\Pi^{k+1} = \underset{\Pi}{\argmin}~ \frac{\rho}{2} \| \Pi - \Gamma^{k+1} \|^2 - \tr((V^k)^T \Pi) + \lambda_3 \| \Pi \|_1~.
\label{eq:pi_mblock}
\eeq

{\bf Dual Updates for $S, U, V$:} The updates for the dual variables are as follows:
\begin{align}
S^{k+1} &  = S^k + \rho (\Theta^{k+1} - Q^{k+1}) \\
\u_t^{k+1} & = \u_t^k + \rho(\Th_t^{k+1} - \sum_{\substack{\ell=1 \\ \ell \neq k}}^T w_{|t-\ell|} \Th_{\ell}^{k+1} - \ggamma_t^{k+1})~, \forall t \\
V^{k+1} & = V^k + \rho (\Gamma^{k+1} - \Pi^{k+1})~.
\end{align}

\subsection{Two- vs.~multi-block ADMM}

From the presented two-block and multi-block ADMMs definition, we notice that the only difference is the fact that multi-block ADMM has an additional block due to the coupling of $\Theta$ and $\Gamma$ variables. Table~\ref{fig:side_by_side} shows a side-by-side comparison of both ADMM variants.  In the two-block version, $\Theta$ and $\Gamma$ are update independently, and as such, $\boldsymbol{\gamma}_{t}^{k+1}$ is updated based on the previous value of the lifting variable $\mathbf{q}_t^{k+1}$, while on the multi-block version $\boldsymbol{\theta}_{t}^{k+1}$ uses the already updated primal variable $\boldsymbol{\theta}_{t}^{k+1}$.  From this perspective, the multi-block ADMM performs a block coordinate descent on the primal variables $\Theta$ and $\Gamma$, while the two-block carry out a block update.


\begin{table}[htb]
	\caption{Side-by-side comparison between the two-block and multi-block ADMM.  Two variables in the same block means that they can be updated independently.}
	\centering
	\begin{tabular}{|c|c|c}
		\multicolumn{2}{c}{Two-block}  \\ \hline
		\multirow{2}{*}{$Z_{1} = (\Theta, \Gamma)$} &  $\Theta^{k+1} = $ Eq.~(\ref{eq:theta_twob})  \\
		&  $\Gamma^{k+1} = $ Eq.~(\ref{eq:gamma_twob}) \\ \hline
		\multirow{2}{*}{$Z_{2} = (Q, \Pi)$} &  $Q^{k+1} = $ Eq.~(\ref{eq:q_twob})  \\
		&  $\Pi^{k+1} = $ Eq.~(\ref{eq:pi_twob})\\
		
		\hline
	\end{tabular}
	\begin{tabular}{|c|c|c|}
		\multicolumn{2}{c}{Multi-block}\\ \hline
		$Z_{1} = (\Theta)$ &  $\Theta^{k+1} = $ Eq.~(\ref{eq:theta_mblock}) \\  \hline
		$Z_{2} = (\Gamma)$ &  $\Gamma^{k+1} = $ Eq.~(\ref{eq:gamma_mblock}) \\  \hline
		\multirow{2}{*}{$Z_{3} = (Q, \Pi)$} &  $Q^{k+1} = $ Eq.~(\ref{eq:q_mblock}) \\
		&  $\Pi^{k+1} = $ Eq.~(\ref{eq:pi_mblock}) \\
		\hline
	\end{tabular}
	\label{fig:side_by_side}
\end{table}

\section{Experimental Results}
\label{sec:expts}

In this section, we present experimental analysis to evaluate convergence and performance aspects of multi-block ADMM and two-block ADMM in solving TS-MTL formulation for three multitask learning problems: ($i$) Alzheimer's disease progression, ($ii$) Parkinson's disease assessment, and ($iii$) Air quality prediction.  Before presenting the datasets, we discuss the experimental setup and evaluation metrics used for all three datasets.

{\bf Experimental methodology:}
We randomly split the data into training and test sets using a proportion of 70\% of the data for training and 30\% to evaluate the models. From the training set, 20\% of the data is used as validation set. Ten independent executions were perform to account for variability in the data.
For hyper-parameter selection we consider a grid of regularization parameter values, where each regularization parameter  varied from $ 10^{-1} $ to $ 10^{3} $ in log scale. For this selection, the dual step size $\rho$ was set to 1, which is a commonly used value \cite{Boyd2011}. The data was z-scored before applying regression methods. As the dual step size parameter $\rho$ plays an important role at the convergence of ADMM-type of methods, we considered a wide range of values in our experiments, once the methods regularization parameters are defined.


{\bf Evaluation metrics:} For the quantitative performance evaluation, we computed the Root Mean Squared Error (rMSE) between the predicted and target clinical scores for each time point (task). For aggregated performance over all time points, the normalized mean squared error (nMSE) \cite{argyriou2008convex,cFSGL} is used. The rMSE and nMSE are defined as follows:  $\text{rMSE}(Y_{t},\hat{Y}_{t}) = (\frac{1}{n_t}\sum_{i=1}^{n_t} (Y_t^i - \hat{Y}_t^i)^2)^{1/2}$, and $\text{nMSE}(Y,\hat{Y}) = (\sum_{t=1}^k \frac{\|Y_t-\hat{Y}_t \|_2^2}{\sigma(Y_t)})/(\sum_{t=1}^k n_t)$,
%
where  $ Y $ and $ \hat{Y} $ are the ground truth cognitive scores and the predicted cognitive scores, respectively, and $n_t$ is the number of samples of the $t$-th time point. Smaller values for nMSE and rMSE  represent better regression performance. The average (avg) and standard deviation (std) of performance measures across multiple runs on different splits of data are shown as avg $\pm$ std for the predictive performance experiment.

We show nMSE performance of the methods on the validation set over time in order to correlate the reduction in the primal residuals with the actual  generalization performance of the multitask regressors. We report the primal residual as the sum of the squares of the three primal residuals defined by the constraints in Eqs.~(\ref{eq:constrained2}) and (\ref{eq:constrained3}). It is an aggregated measure of how much the current solution violates the constraints.

\subsection{Alzheimer's disease progression}

We used the dataset from ADNI \footnote{http://adni.loni.usc.edu}, which is a multi-site study that aimed to improve clinical trials for the prevention and treatment of Alzheimer disease (AD).  The study gathered and analyzed thousands of structural MRI brain scans, genetic profiles, and biomarkers in blood and cerebrospinal fluid that are used to measure the progress of disease or the effects of treatment.  For this study, we used data from the first phase of ADNI project, called ADNI-1. Subjects perform an initial screening, referred to as \textit{baseline} (BL), in which demographics information and an initial brain scan are collected. After approval, the subject is scheduled to attend a calendar of follow-up visits that are denoted by the duration starting from the baseline. We use the notation, for example,  Month 6 (M6) to denote the time point half year after the baseline. Currently, ADNI has up to Month 48 follow-up data available for some patients. Table~\ref{tab:adni_sample_size} presents the number of patients at each time step (follow-up visits) used in this study.

\begin{table}[htb]
	\caption{ADNI: Number of patients (samples) at each time step (task).  M6, M12, M24, M36, and M48 correspond to the number of months after the baseline screening was taken. Number of patients decrease as many drop out the study.}
	\centering
	\begin{tabular}{cccccc} \toprule
		\multirow{2}{*}{Baseline} & \multicolumn{5}{c}{Months after baseline} \\ \cline{2-6}
		& M6 & M12 & M24 & M36 & M48 \\
		\hline
		788 & 718 & 662 & 532 & 345 & 91 \\
		\bottomrule
	\end{tabular}
	\label{tab:adni_sample_size}
\end{table}

At each visit, subject's cognitive function is measured by its performance on a set of neuro-psychological tests of most cognitive domains
such as memory, verbal communication, reasoning, coordination of movement, and planning of tasks. Following ADNI guidelines, a few cognitive scores are taken at each visit, including \cite{CORNLIN,High-Order}: Alzheimer's Disease Assessment Scale (ADAS-Cog), Mini-Mental State Exam (MMSE), and Rey Auditory Verbal Learning Test (RAVLT-TOTAL). 

Using the MRI brain scans from each visit, the multi-task learning problem involves accurately predicting a given cognitive score over multiple time steps, i.e., each task focuses on modeling a given cognitive score at a given time step, and different tasks focus on different time steps for the same cognitive score. Instead of using raw structural MRI images, we used the features from cortical reconstruction and volumetric segmentations. Details of the feature extraction procedure are available at: \url{http://adni.loni.usc.edu/methods/mri-tool/}.  Note that modeling different cognitive scores are treated as separate multitask learning problems, and the tasks correspond to the time points for a given cognitive score. \\ 

\noindent\textbf{Convergence results on ADNI}

\begin{figure*}[htb]
	\centering
		\begin{tabular}{cccc}
			$\boldsymbol{\rho}=0.01$ & $\boldsymbol{\rho}=0.1$ & $\boldsymbol{\rho}=1$ & $\boldsymbol{\rho}=10$ \\ \hline
			\includegraphics[scale=0.2]{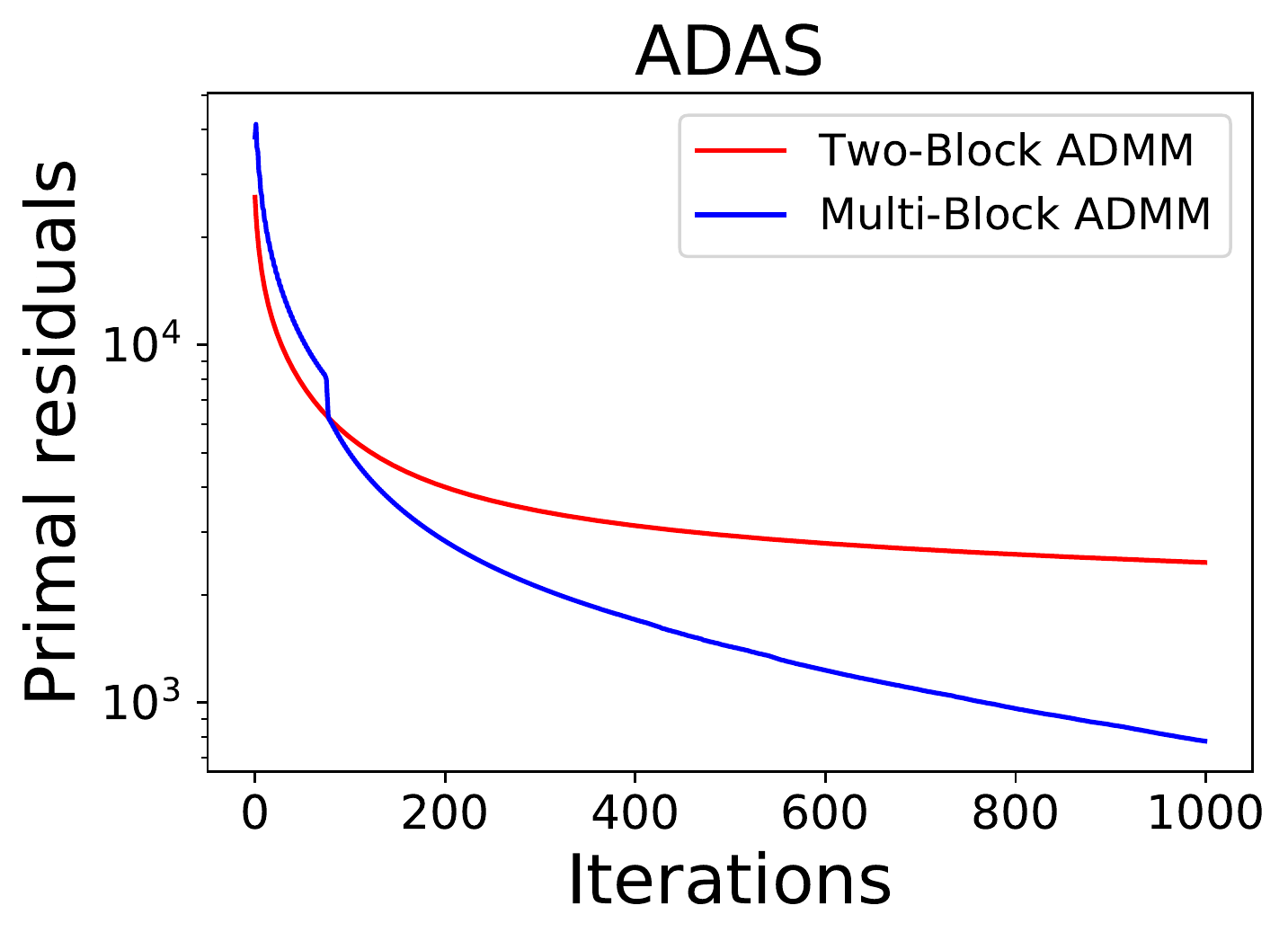}&
			\includegraphics[scale=0.2]{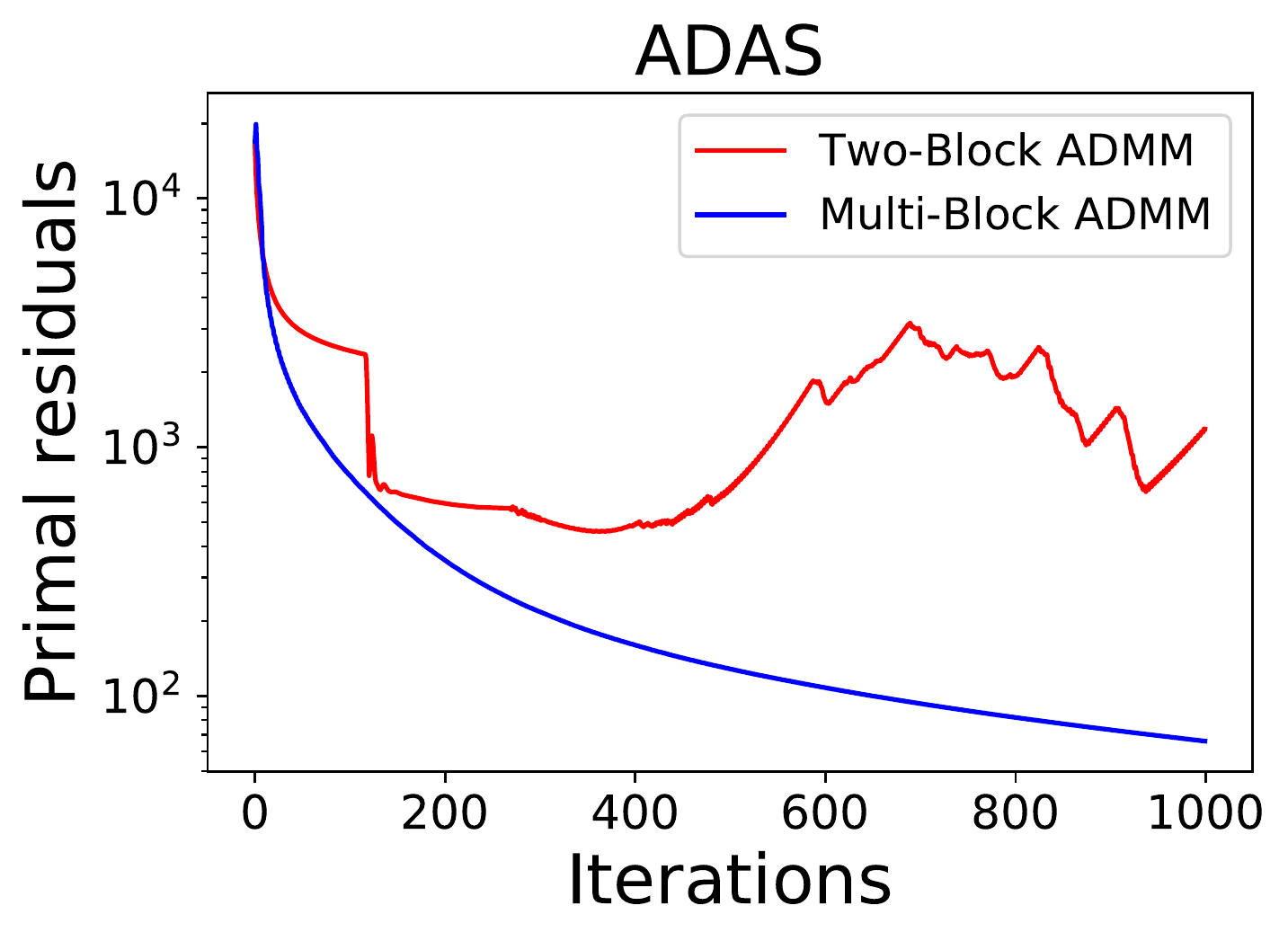}&
			\includegraphics[scale=0.2]{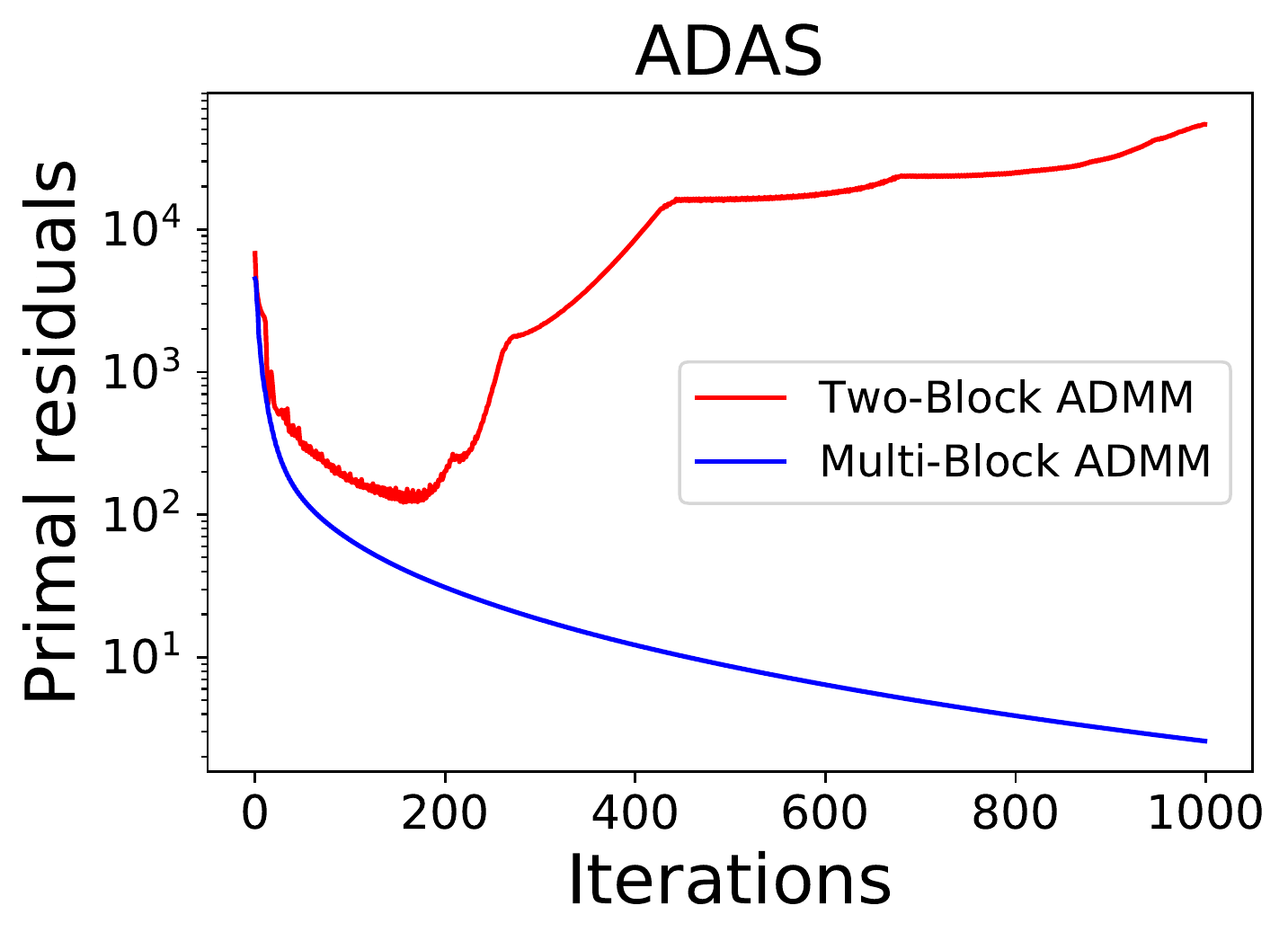}&
			\includegraphics[scale=0.2]{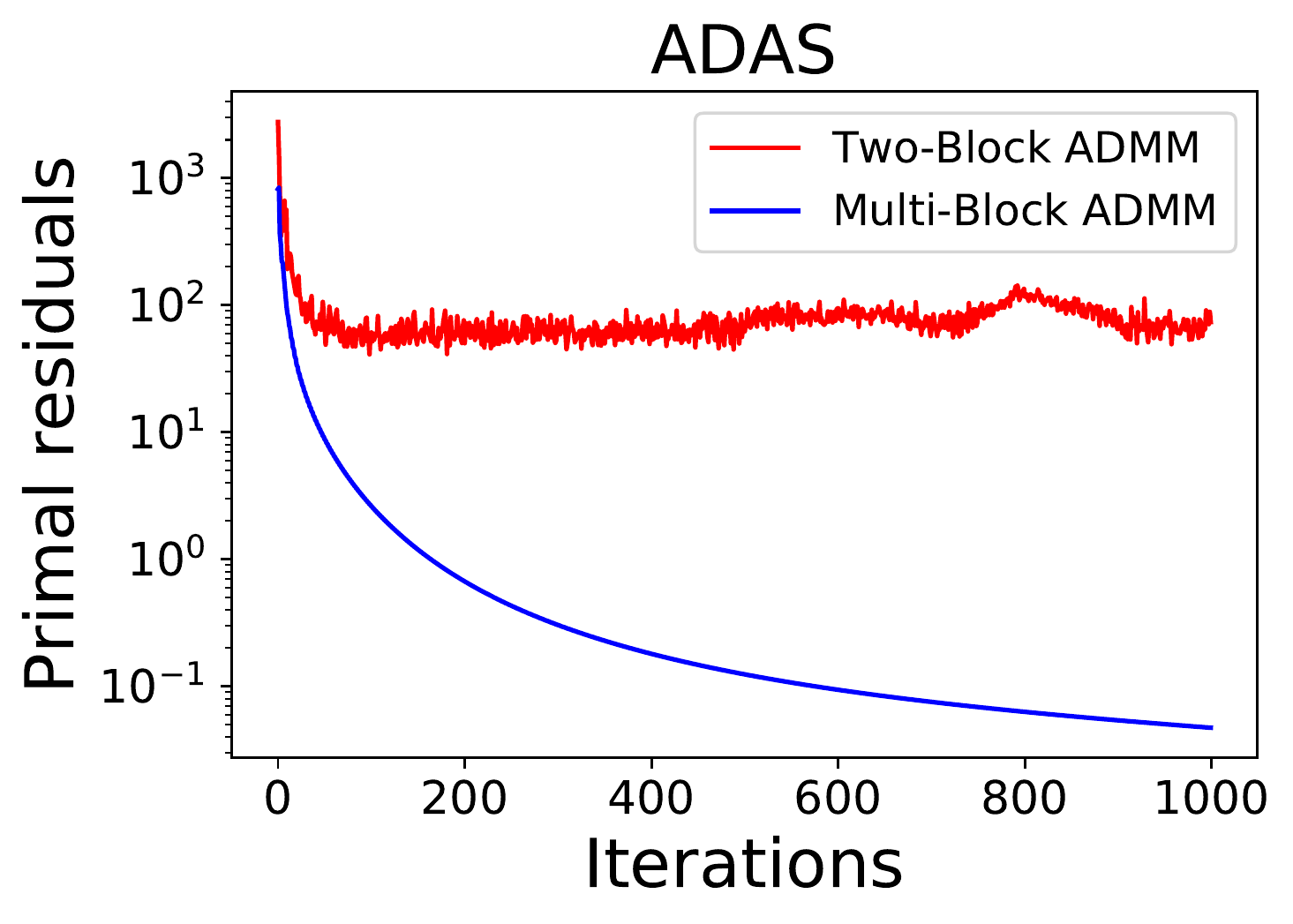}\\
			
			\includegraphics[scale=0.2]{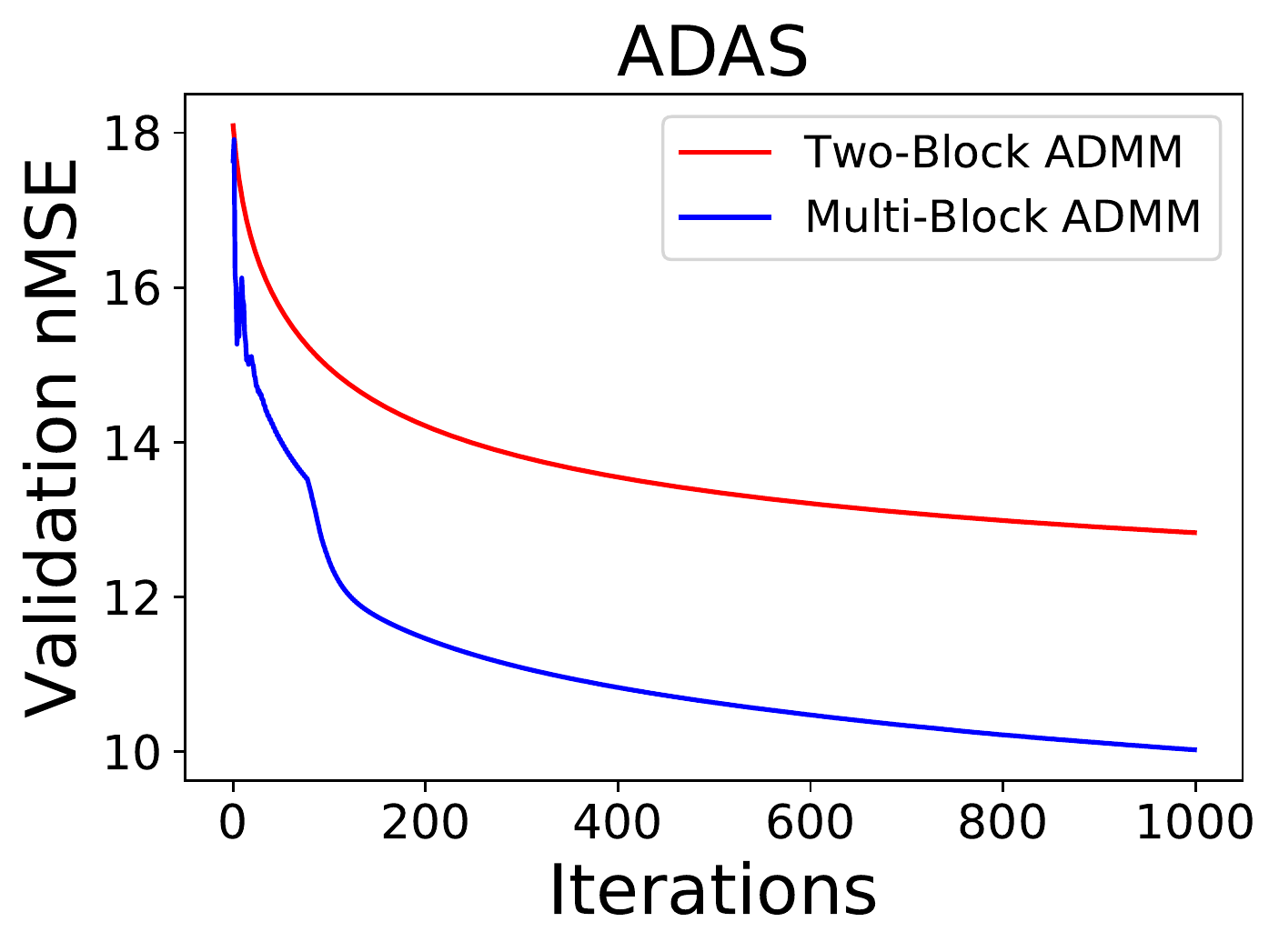}&
			\includegraphics[scale=0.2]{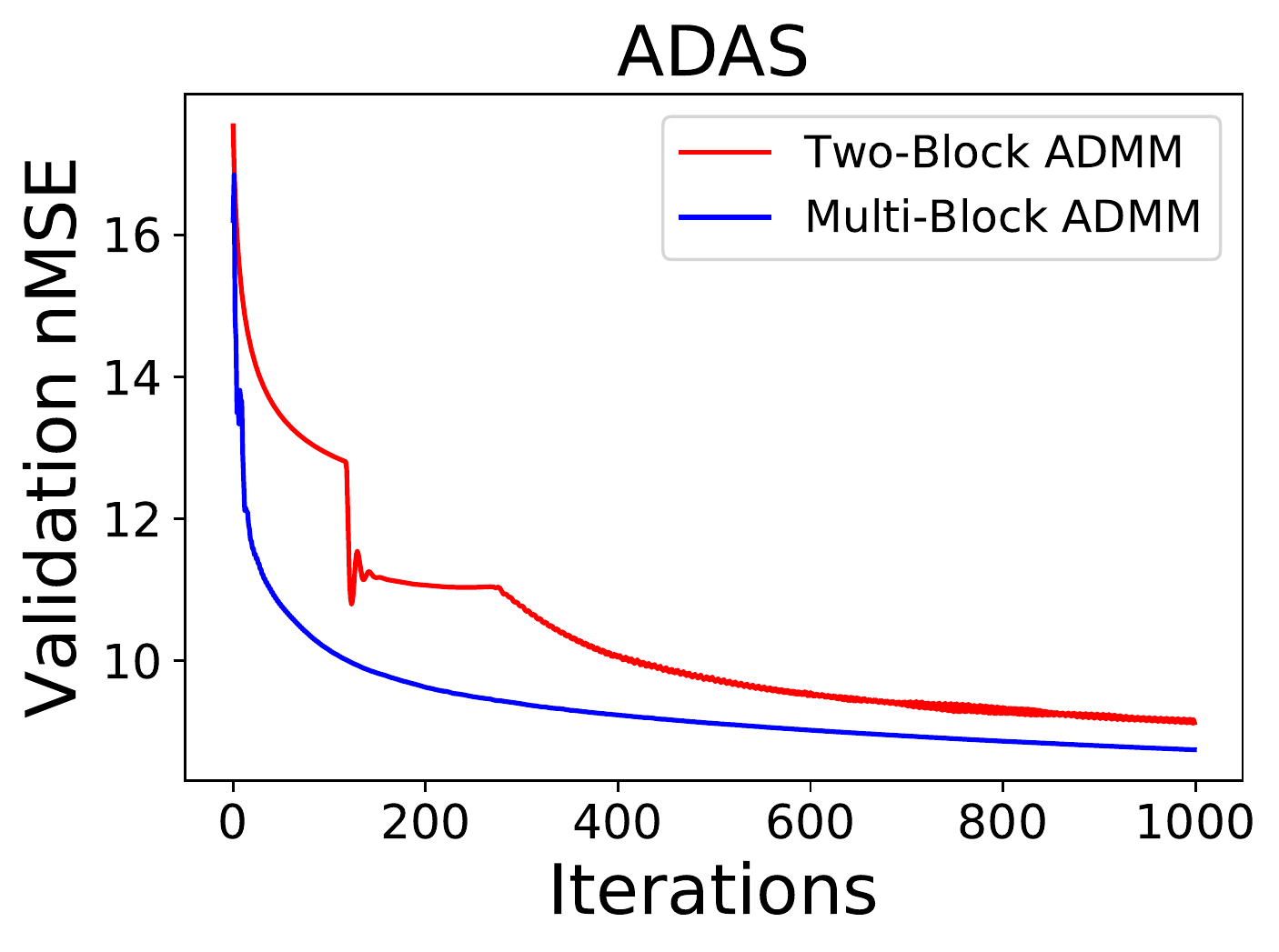}&
			\includegraphics[scale=0.2]{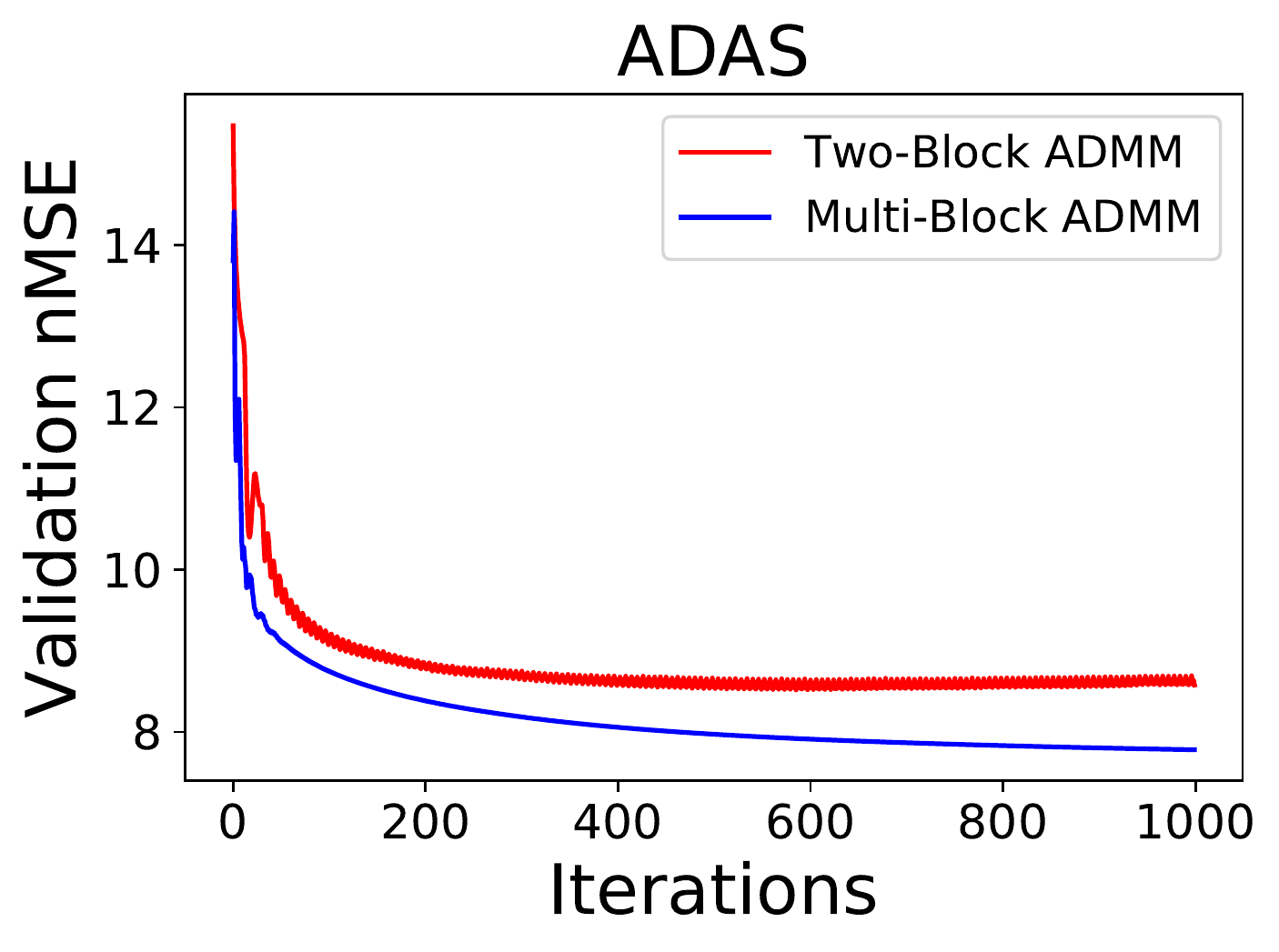}&
			\includegraphics[scale=0.2]{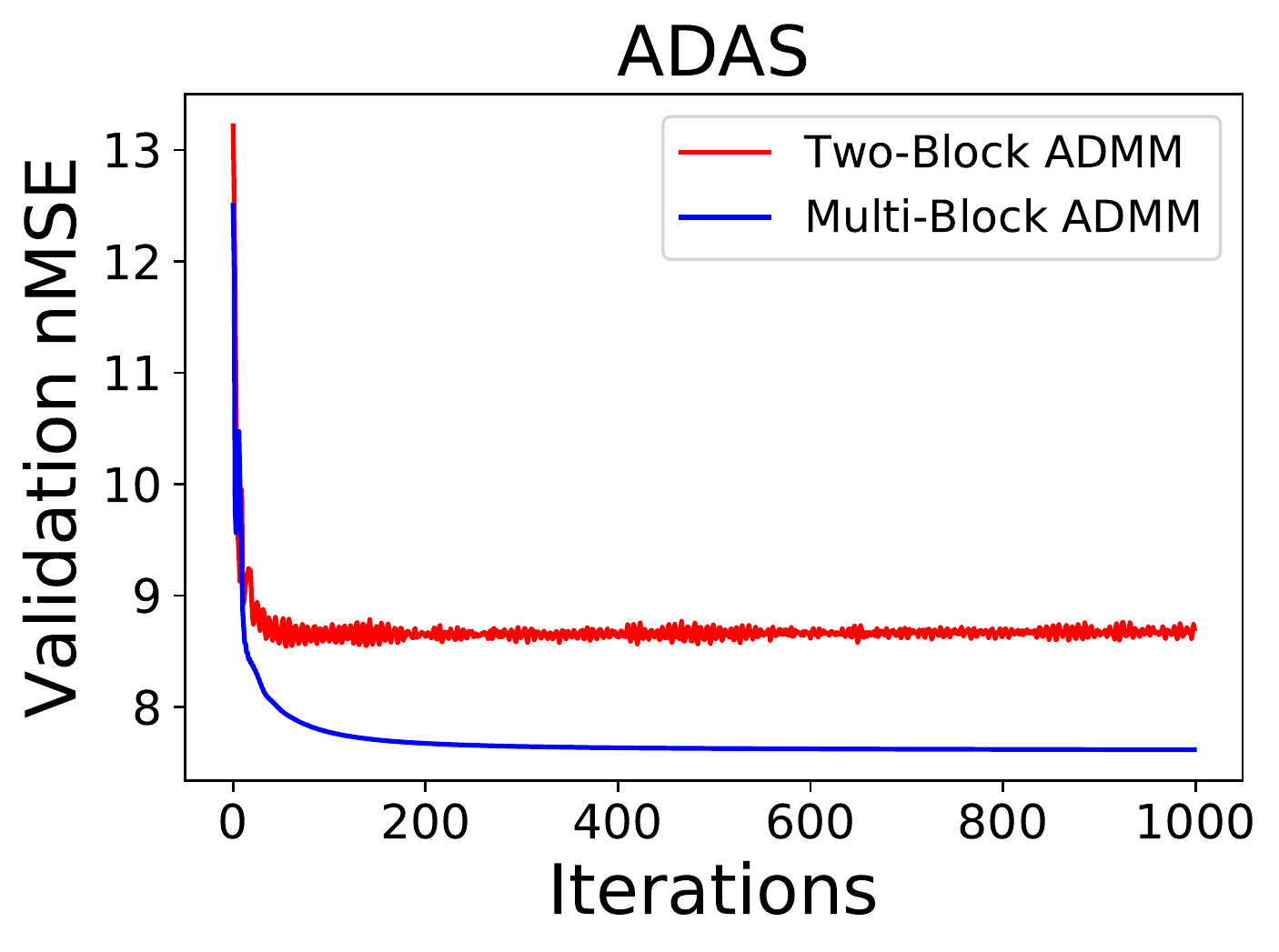}\\
			
			\includegraphics[scale=0.2]{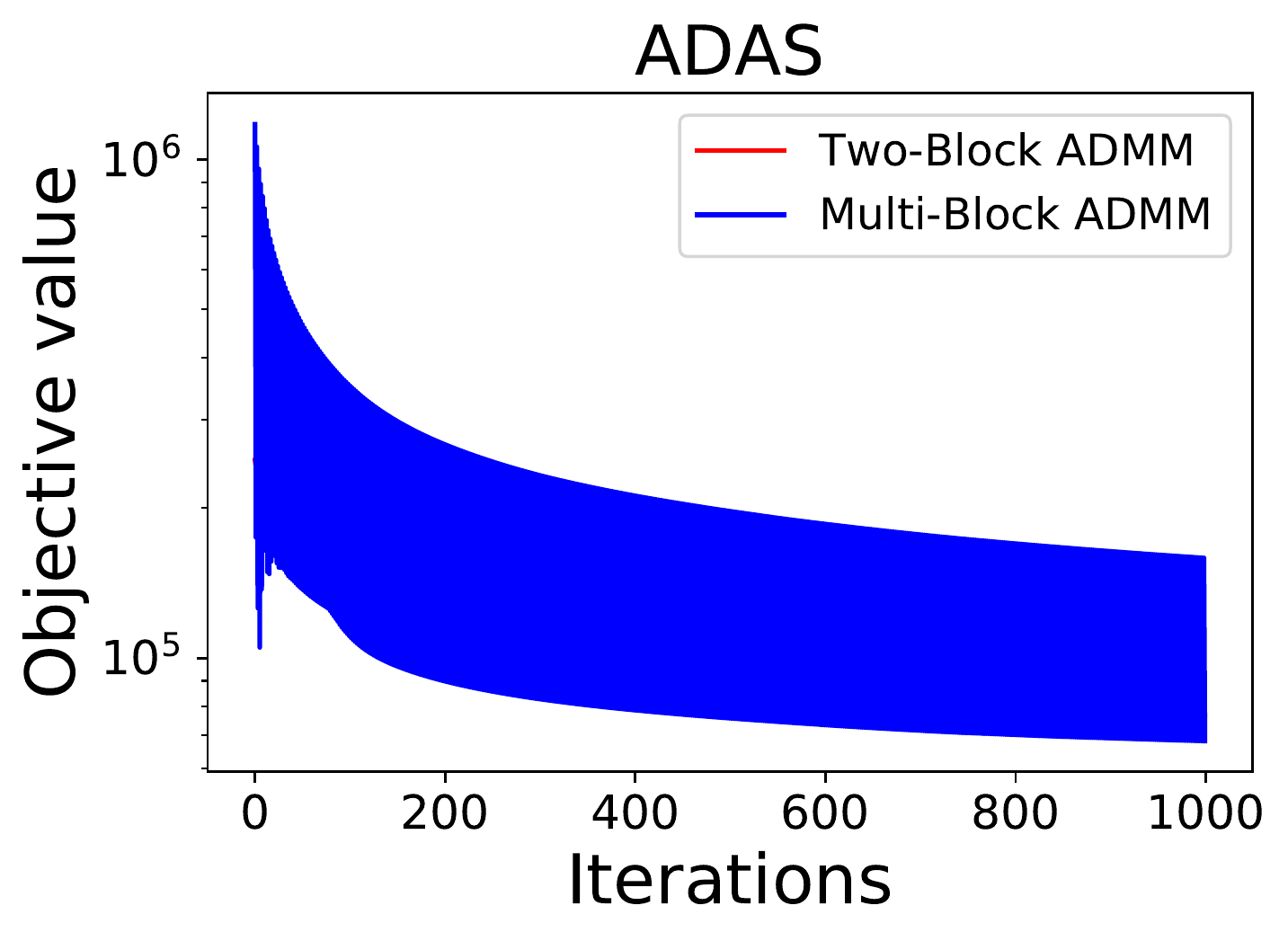}&
			\includegraphics[scale=0.2]{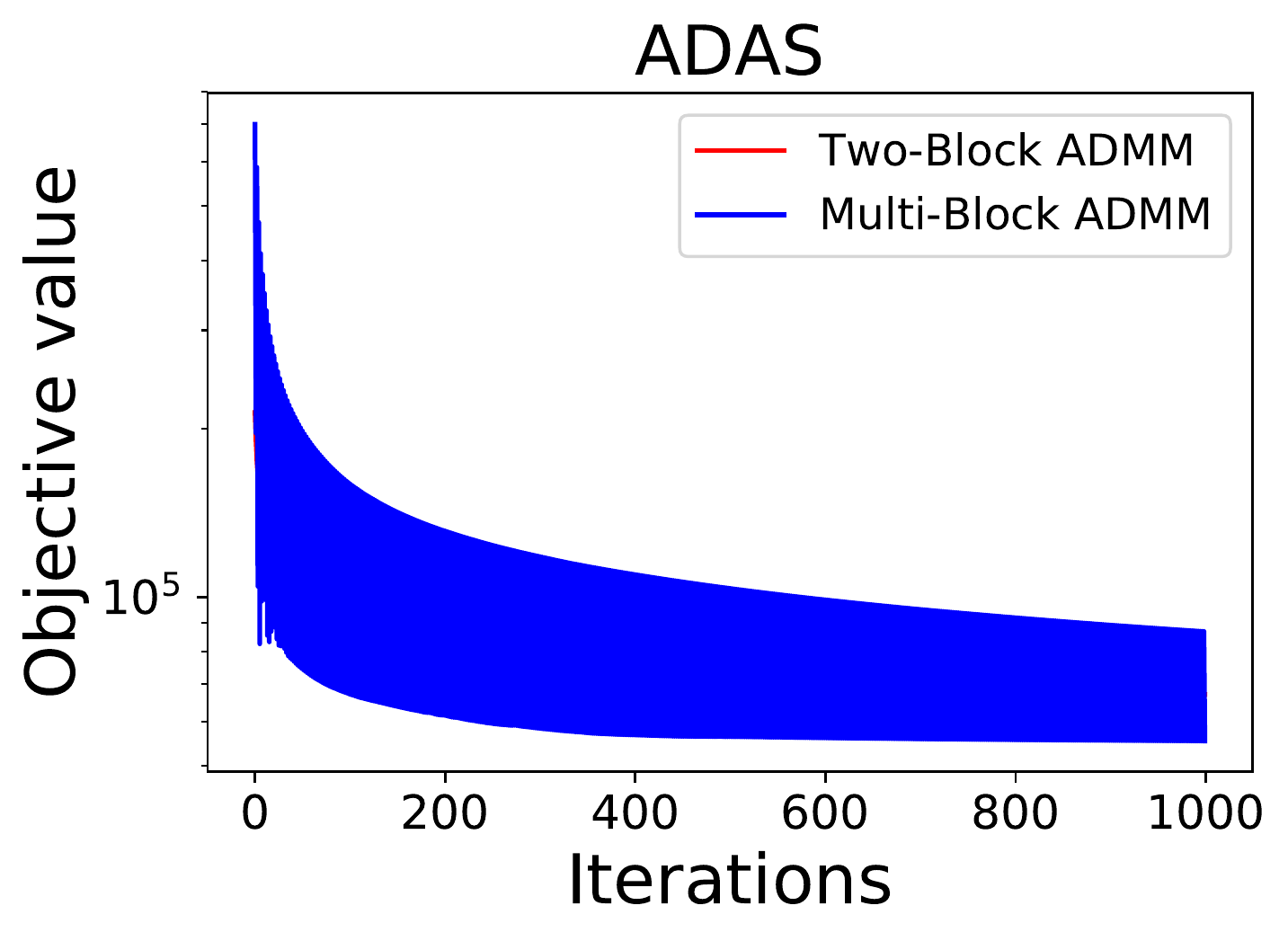}&
			\includegraphics[scale=0.2]{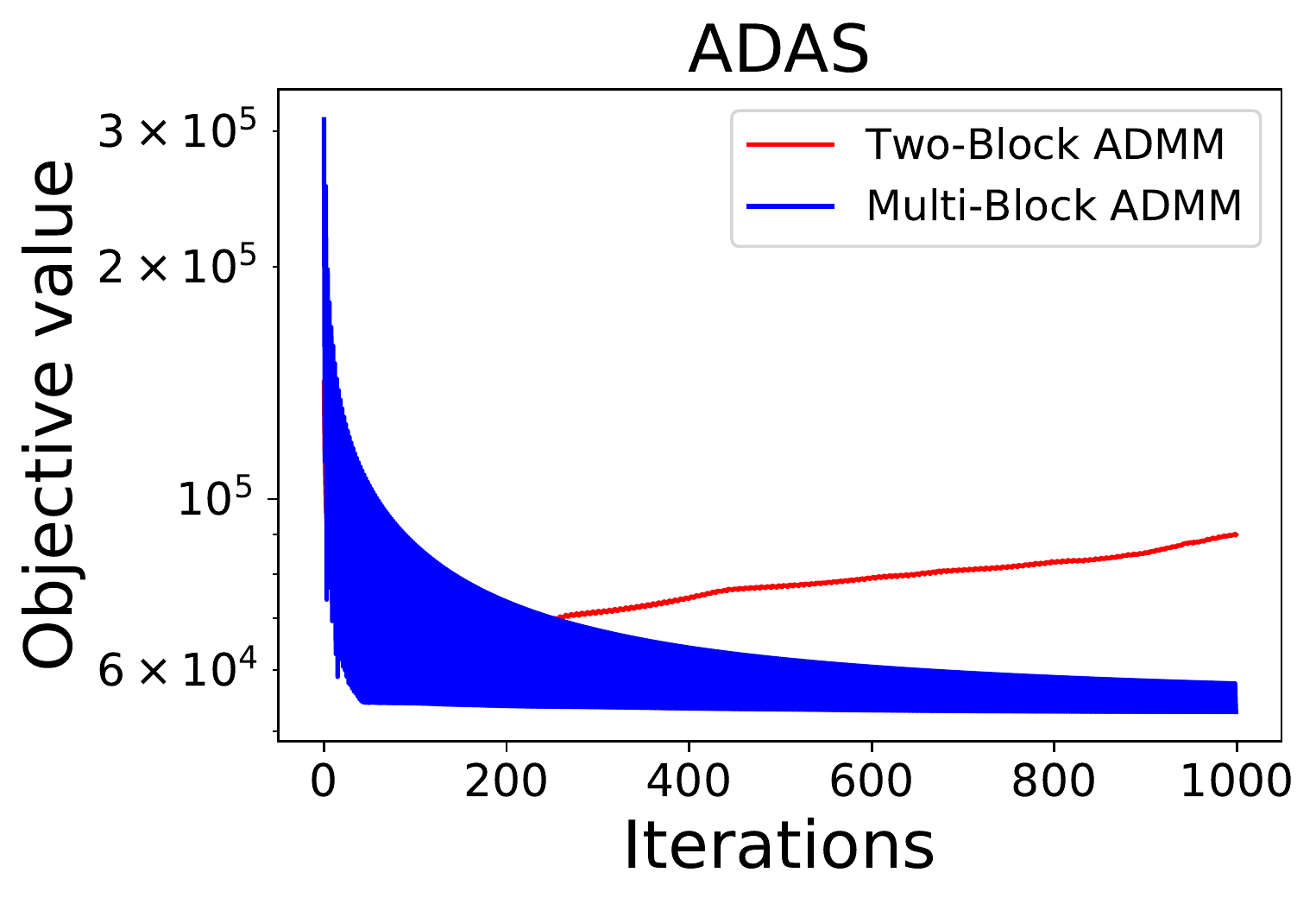}&
			\includegraphics[scale=0.2]{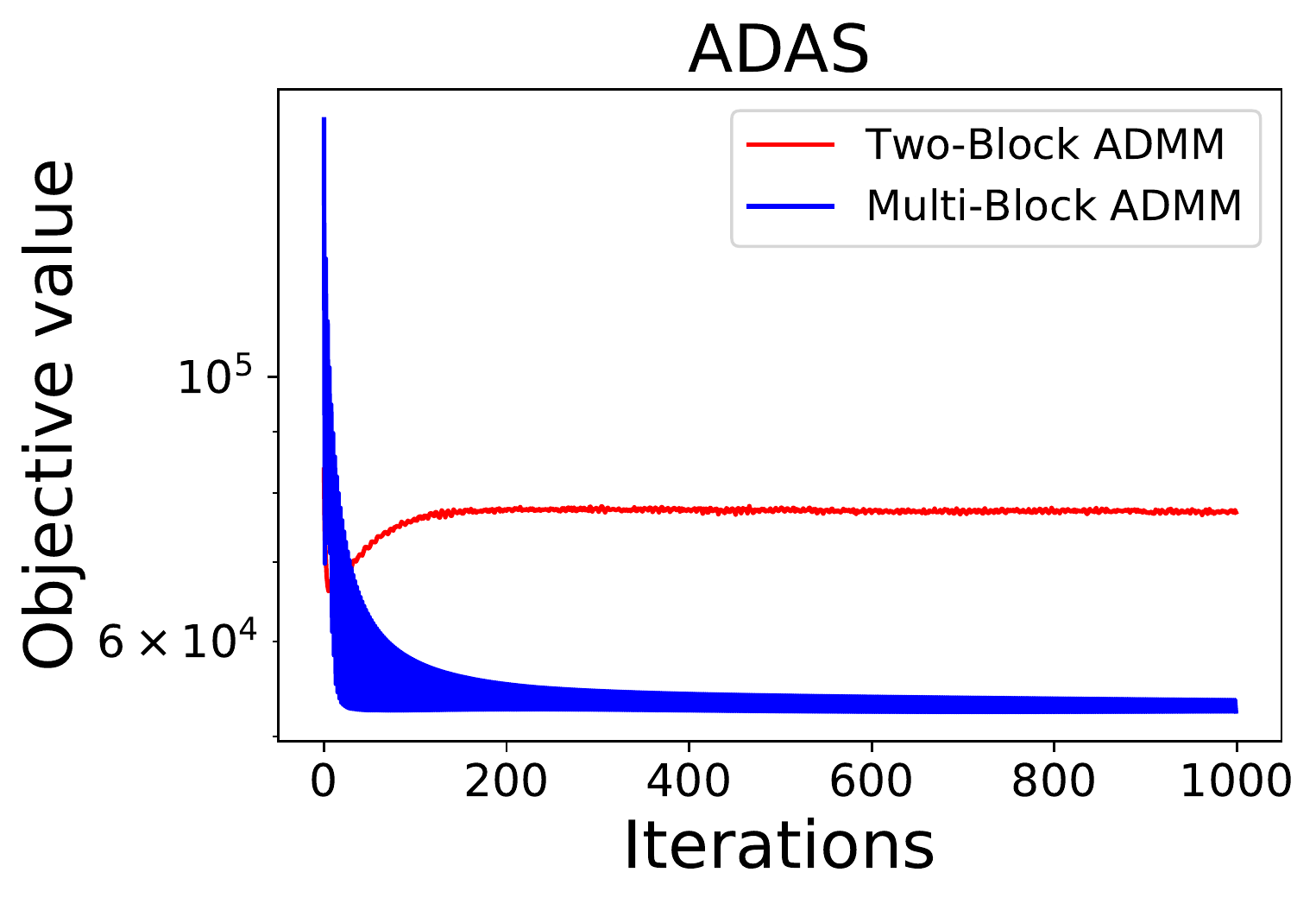}\\ 
			
	\end{tabular}
	\caption{Primal residual, validation nMSE performance, and primal objective computed on ADAS cognitive scores from the ADNI dataset considering several values for the dual step size $\rho$. Plot shows the average curves for 10 independent runs. In the Multi-block ADMM, the primal residual has a smoother behavior and mostly monotonically decreases over time,  while two-block ADMM shows a more unstable  convergence. Better convergence of multi-block ADMM leads to a lower nMSE in the validation set.}%
	\label{fig:adni1_convergence}
\end{figure*}

Figure~\ref{fig:adni1_convergence} shows the convergence of the two-block and multi-block ADMMs for the ADAS cognitive score. Due to space limitations we do not present the curves for MMSE and RAVLT-TOTAL, as they showed similar behaviors. As we can see, the multi-block ADMM has a smooth and steady convergence than the two-block version for all the dual step sizes tested. On the other hand, the two-block starts with a convergence rate similar to multi-block, but the primal residual increases ($\rho=0.1$ and $\rho=1$) or get stuck and oscillate in a region of the optimization space ($\rho=10$). The superior  convergence of multi-block ADMM reflects the better generalization of the trained multitask learning regressors, measured by their performance on the validation set. In all the step sizes investigated, notably, the multi-block ADMM achieved solutions with higher generalization capacity.

It is worth noting that not all variation in the primal residual directly affect the validation performance. This can be clearly observed for $\rho=0.1$ and $\rho=1$, where increments in the primal residual after a period of smooth decreasing do not drop the nMSE performance, which in fact continued to decrease.

\begin{figure}[htb]
	\centering
	\begin{tabular}{ll}
		\includegraphics[scale=0.25]{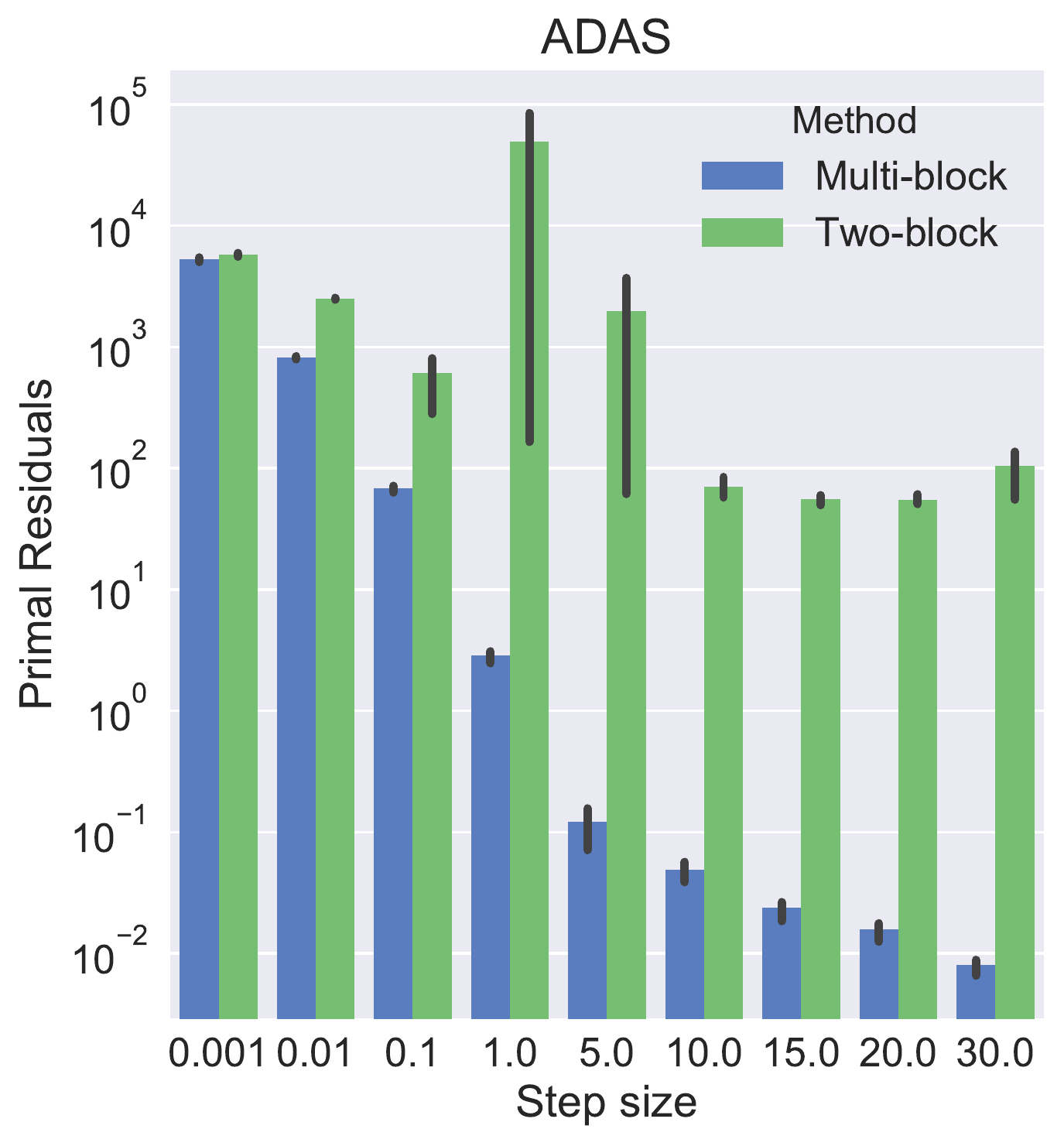}
		\includegraphics[scale=0.25]{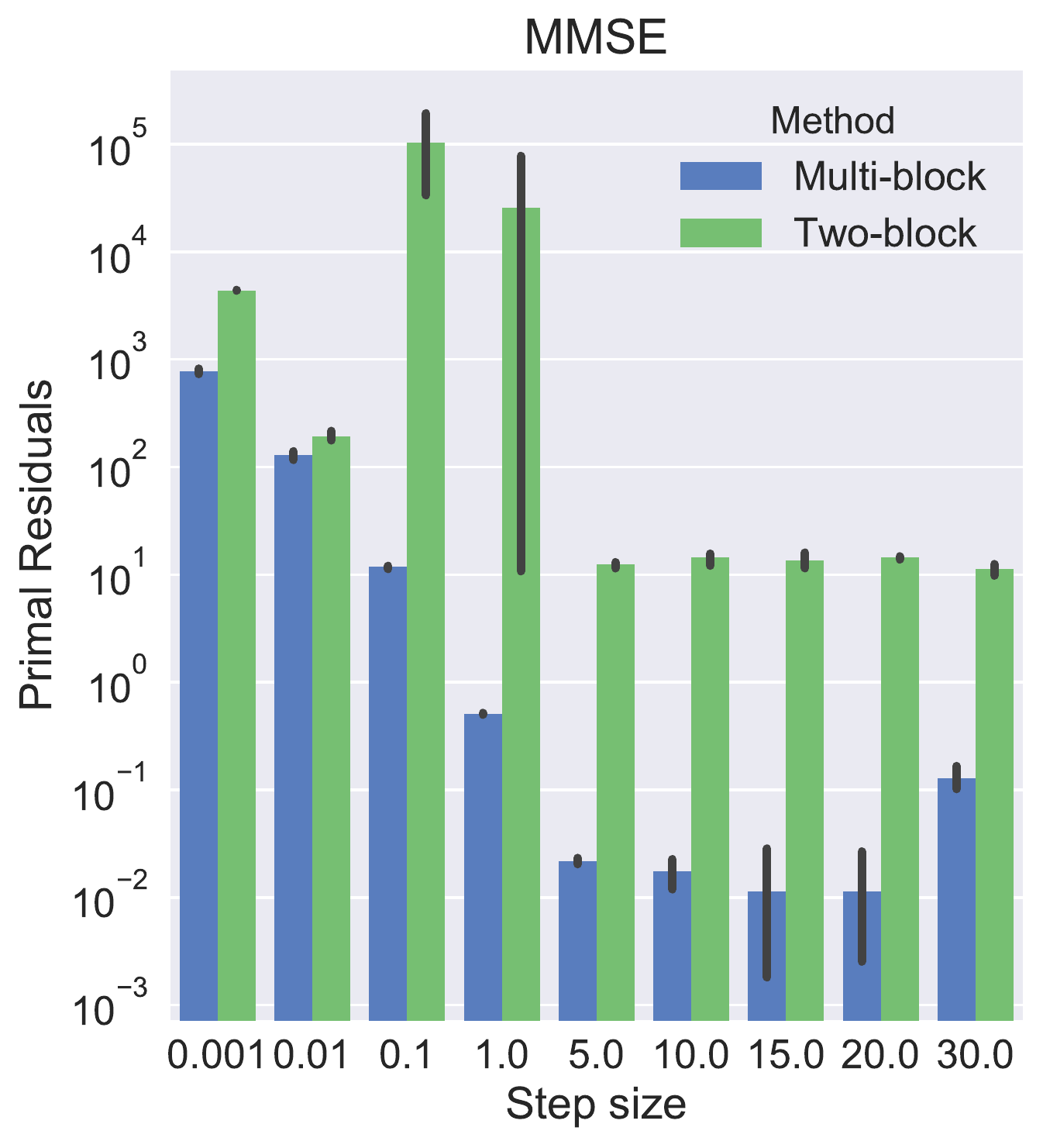}
		\includegraphics[scale=0.25]{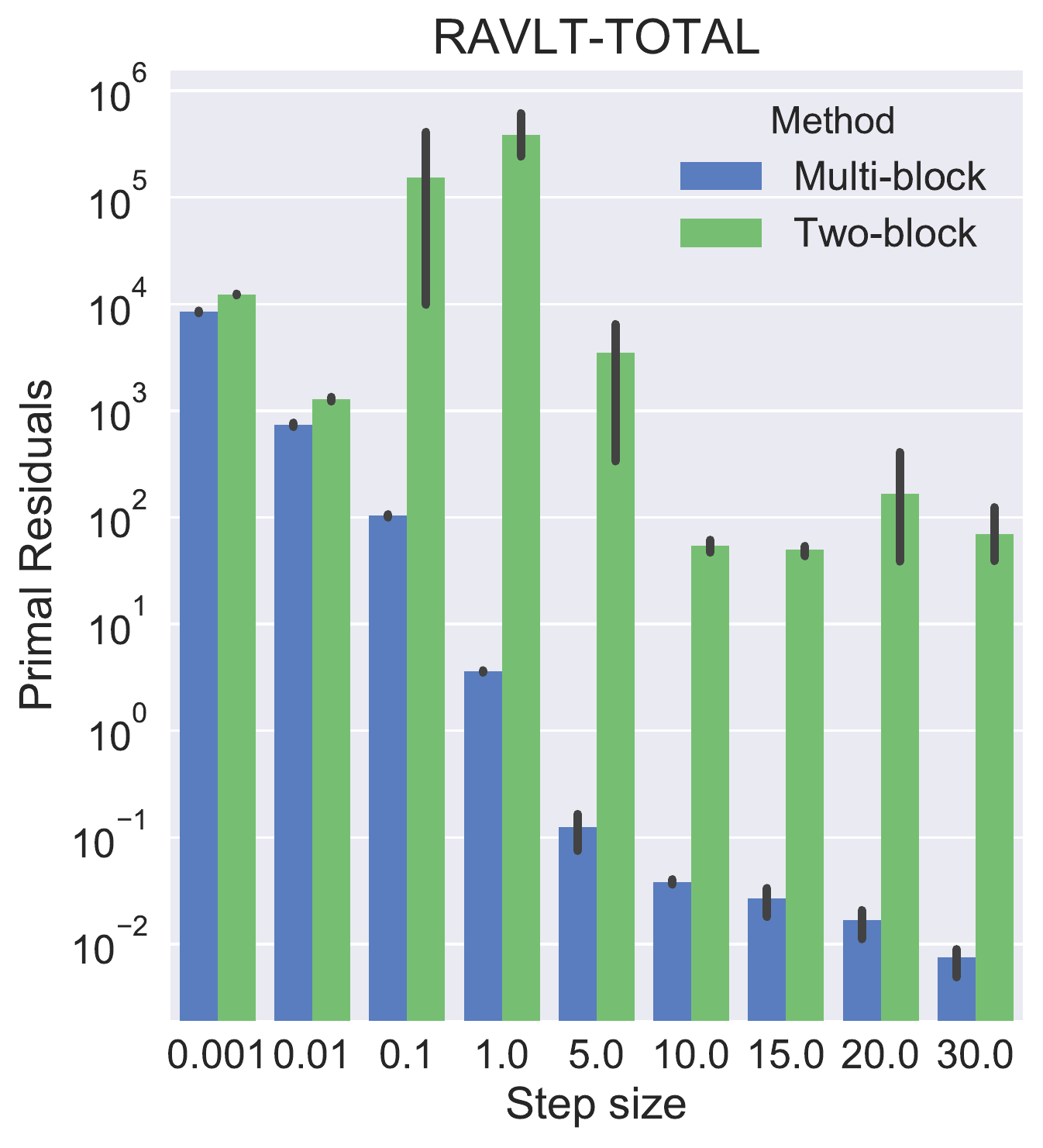} \\
		\includegraphics[scale=0.25]{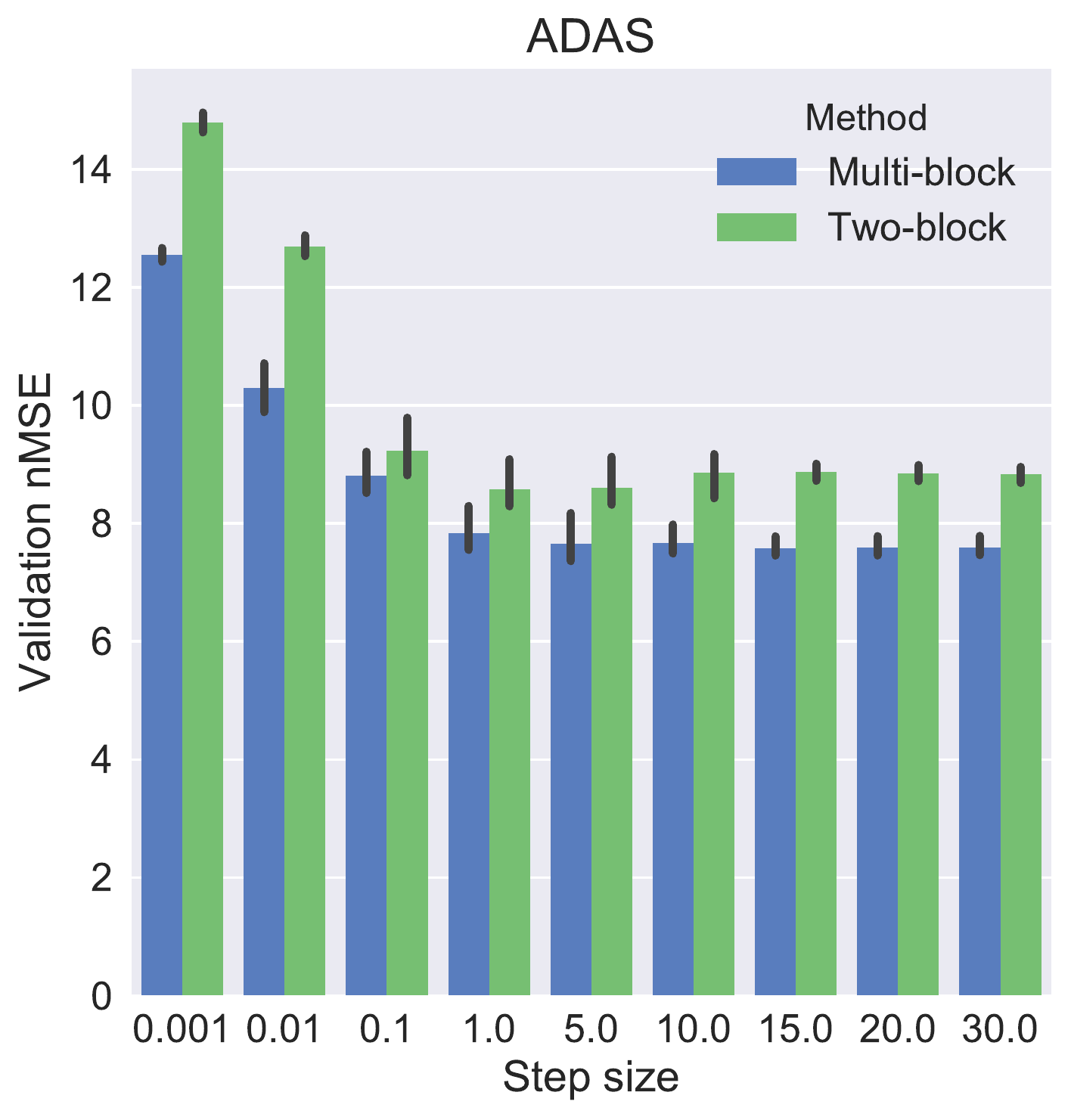}
		\includegraphics[scale=0.25]{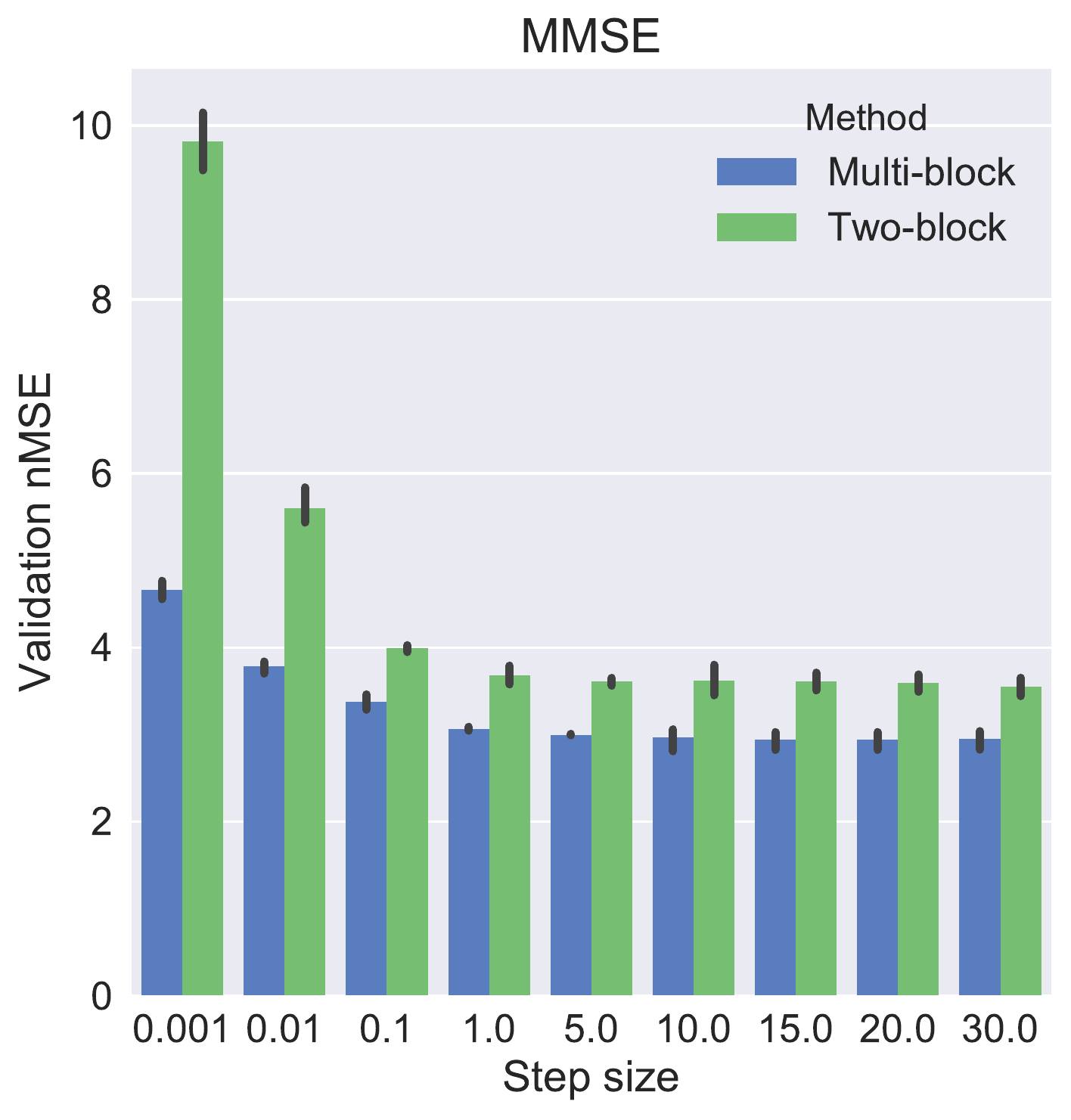} 		
		\includegraphics[scale=0.25]{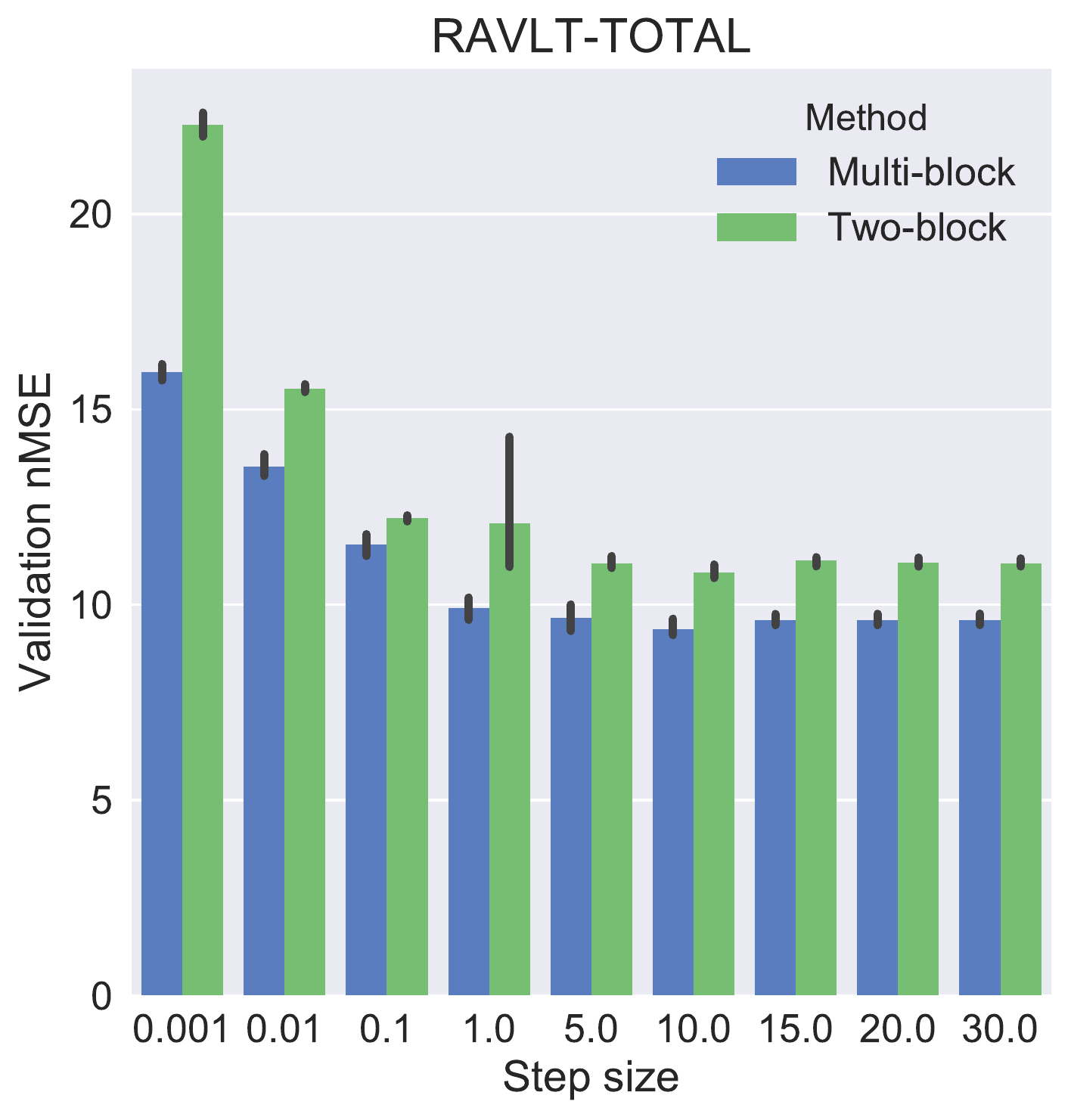} \\
	\end{tabular}
	\caption{Comparison of two-block and multi-block ADMM for TS-MTL formulation on ADNI-2. Validation nMSE and primal residual are computed as the average over the last 100 iterations of the ADMM. Deviation bars show the variation over 10 independent executions of the methods with different train and test data split.}%
	\label{fig:concise_convergence_adni1}
\end{figure}

A condensed visualization of the convergence and validation performance of the methods is presented in Figure~\ref{fig:concise_convergence_adni1}. For each step size $\rho$, we computed and compared the the average of primal residual and validation nMSE over the last 100 iterations out of the 1,000 iterations used for training the model.  The deviation bars show the variation over 10 independent runs. We extended the range of values for dual step size $\rho \in $  [0.001, 0.01, 0.1, 1, 10, 20, 30]. For ADNI, the maximum step size was set to 30, as beyond this value we have empirically observed that the two-block ADMM diverges for this particular problem. For all cases, the multi-block outperforms the two-block ADMM.\\

\begin{figure}[htb]
	\centering
	\scalebox{1}{
		\begin{tabular}{ccc}
			$\boldsymbol{\rho}=0.001$ & $\boldsymbol{\rho}=0.01$ & $\boldsymbol{\rho}=0.1$ \\ \hline
			\includegraphics[scale=0.25]{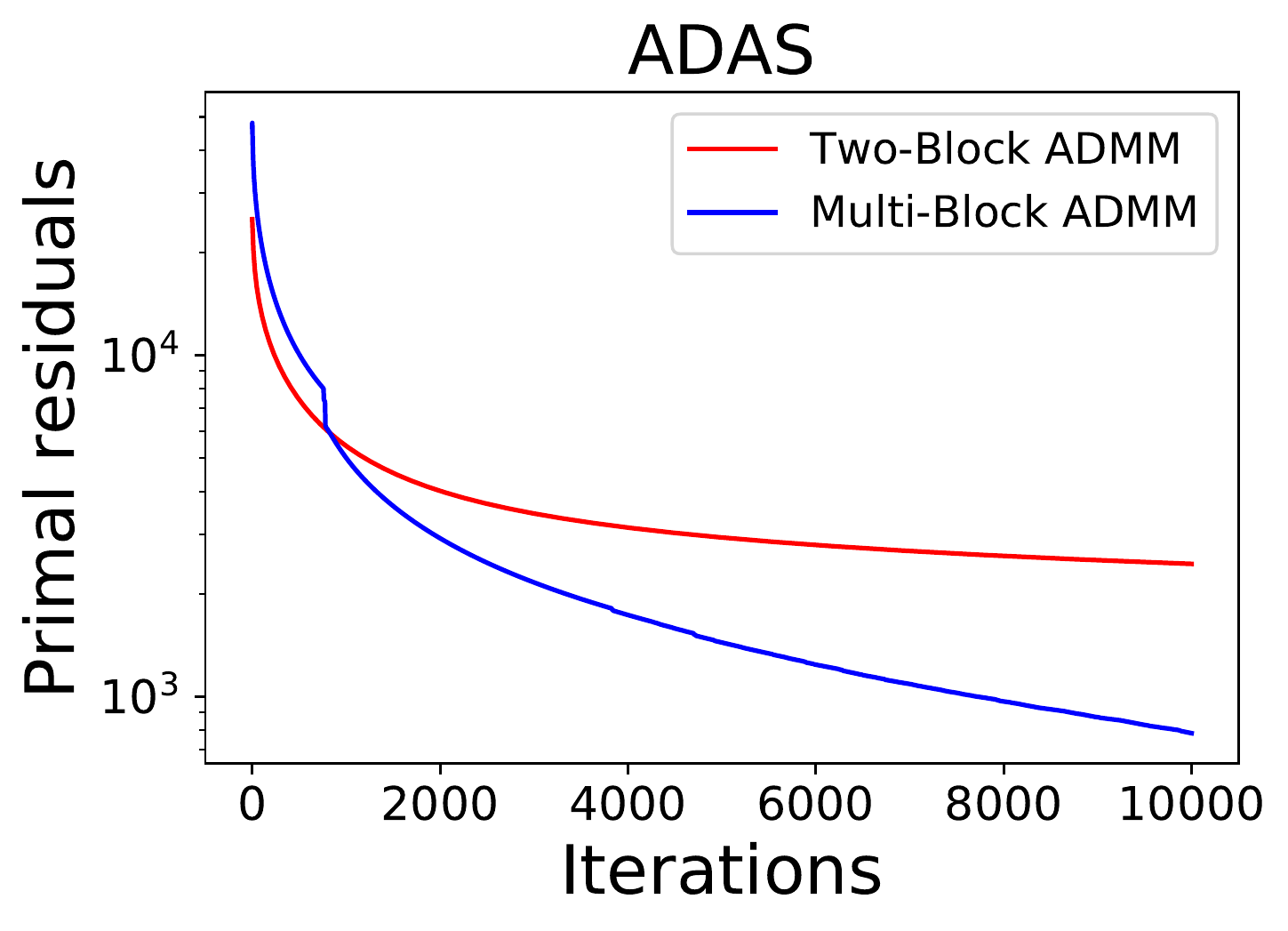}&
			\includegraphics[scale=0.25]{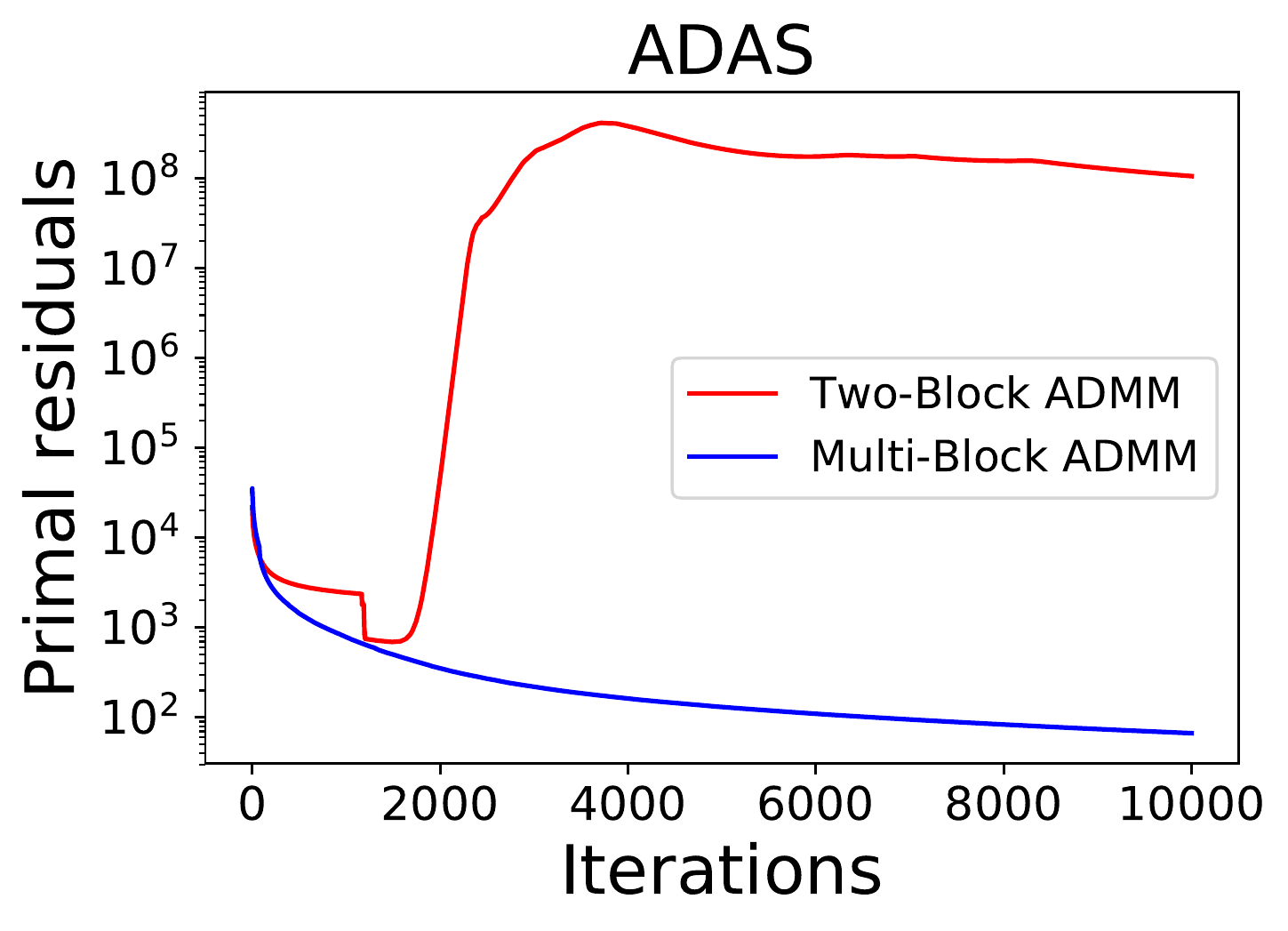}&
			\includegraphics[scale=0.25]{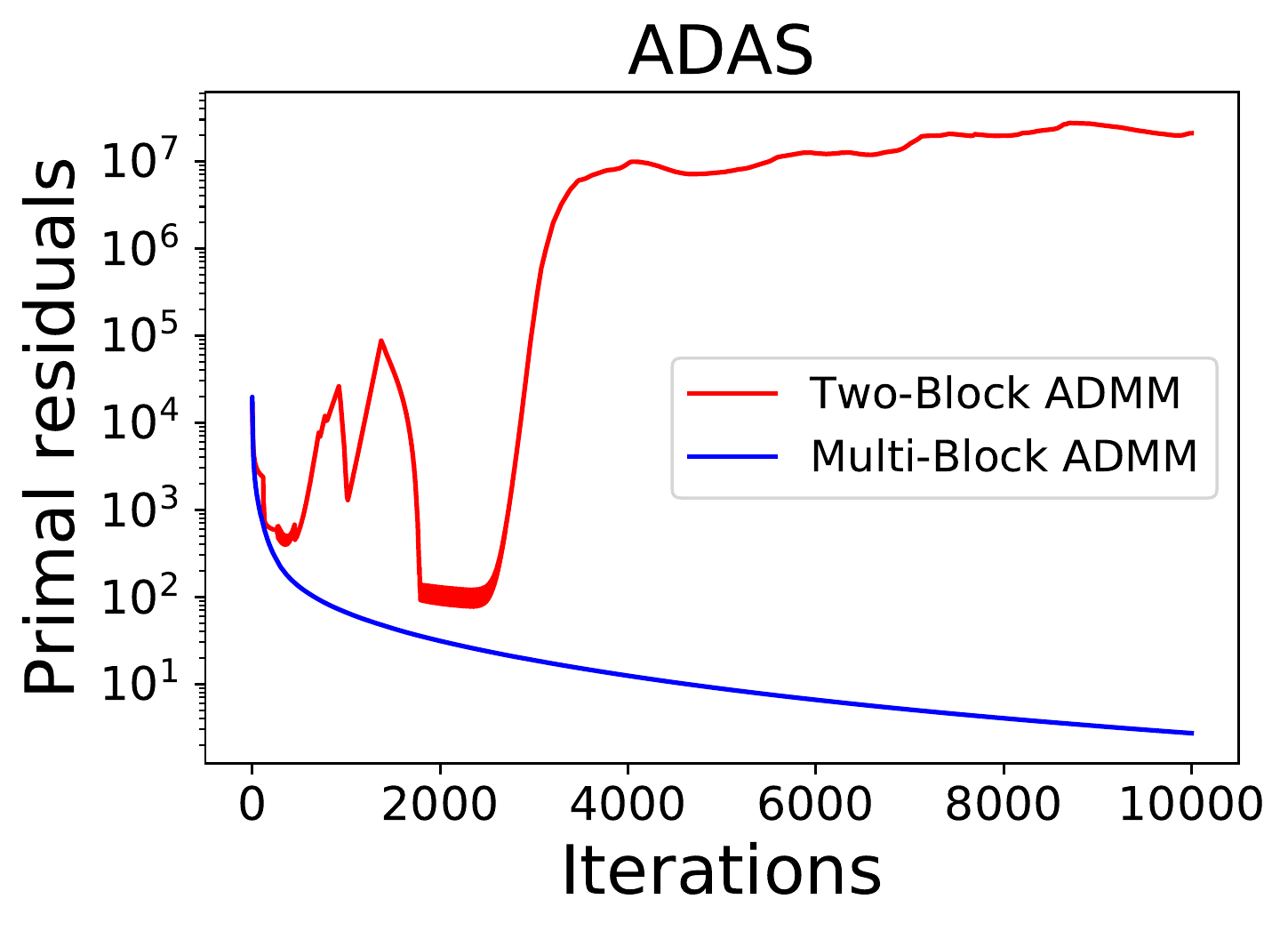}\\
			
			\includegraphics[scale=0.25]{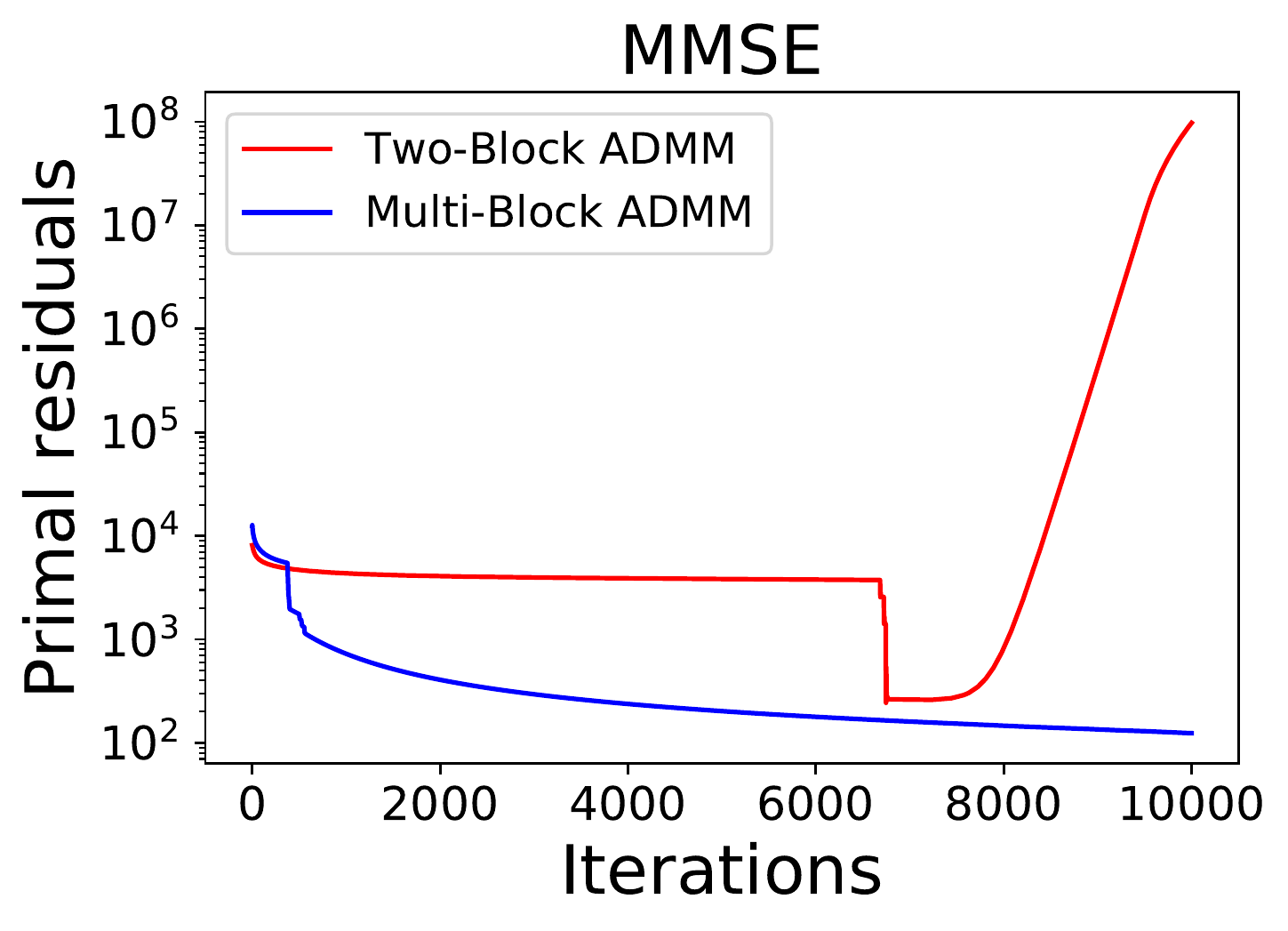}&
			\includegraphics[scale=0.25]{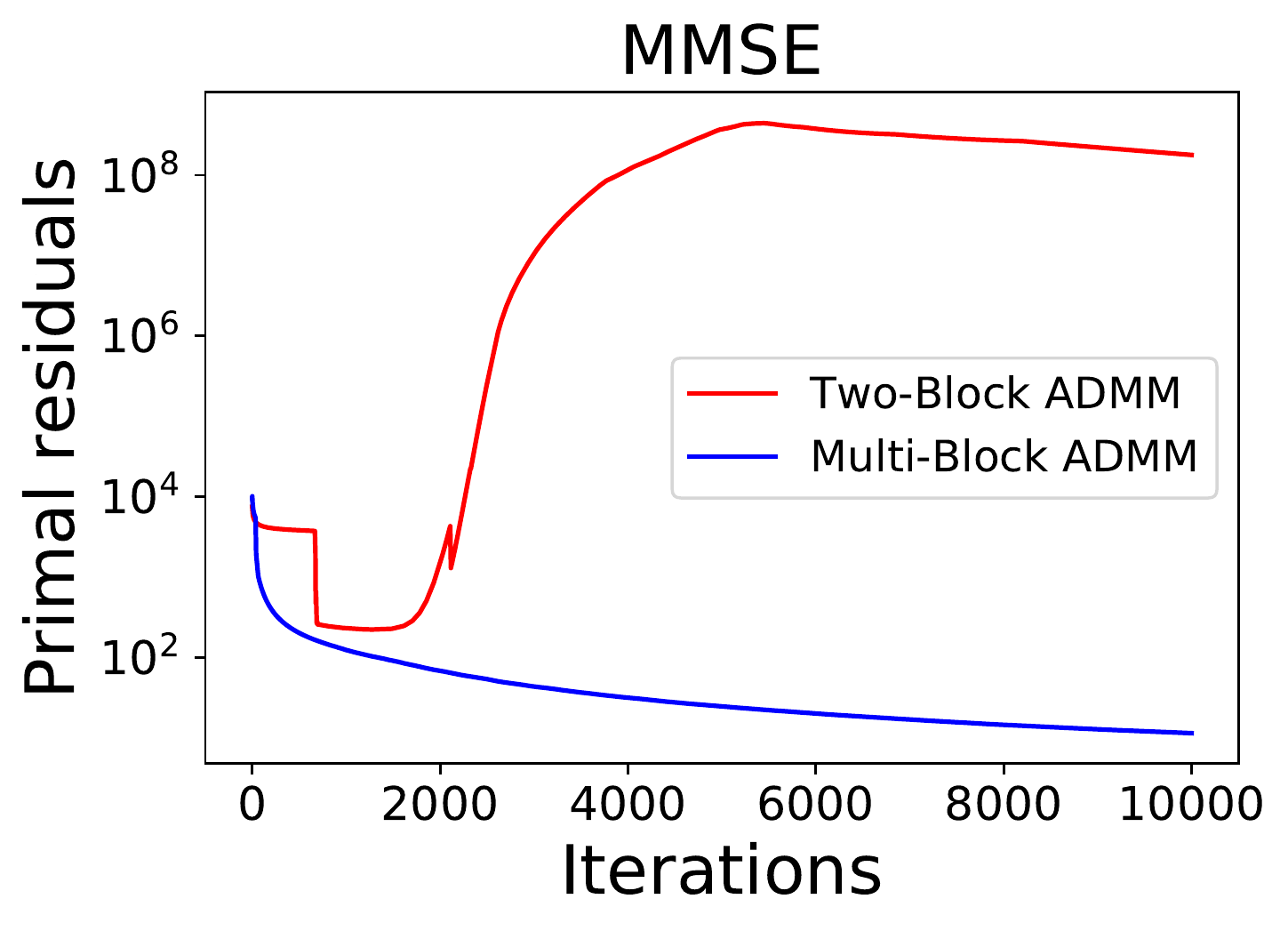}&
			\includegraphics[scale=0.25]{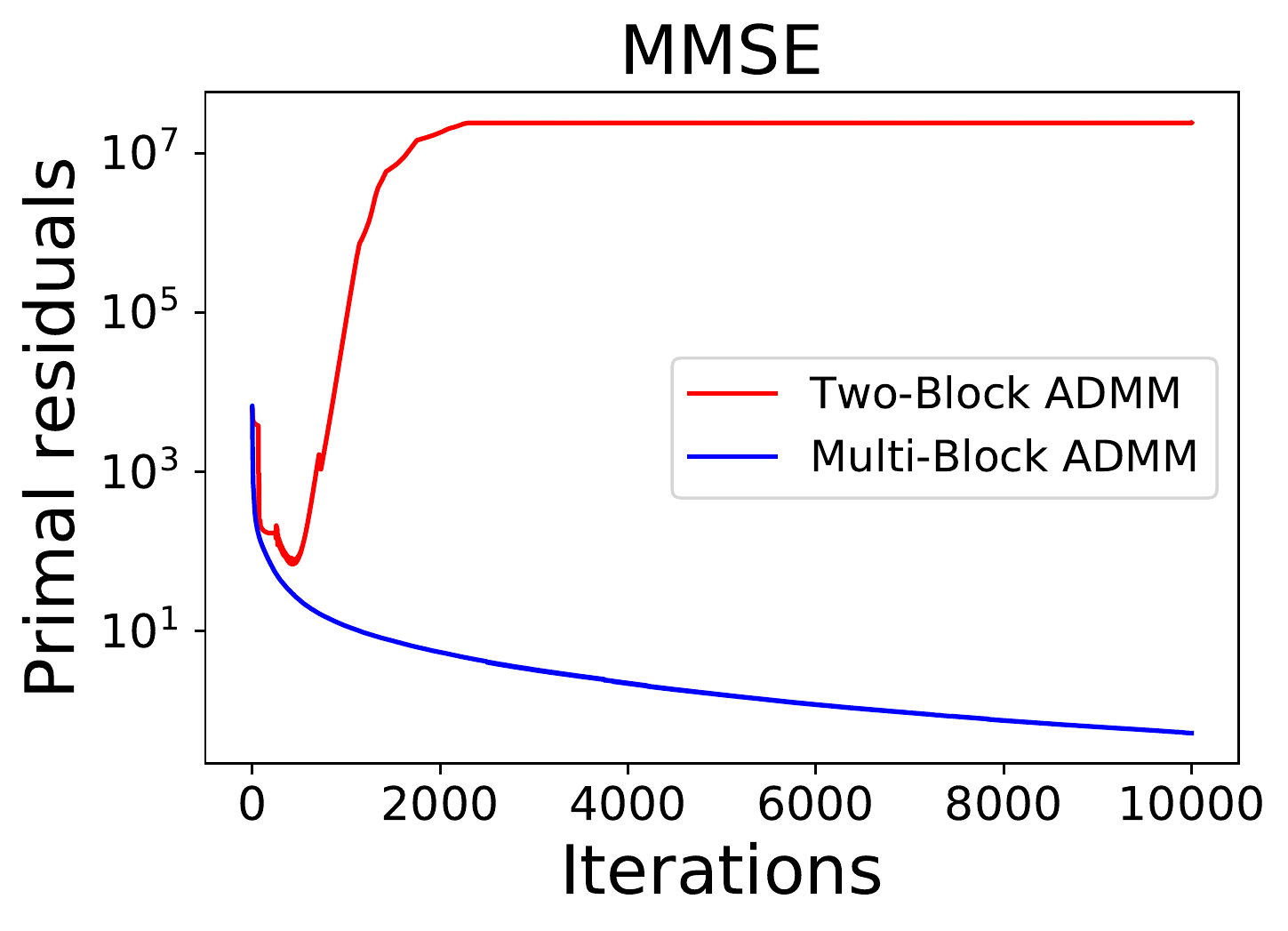}\\
			
			\includegraphics[scale=0.25]{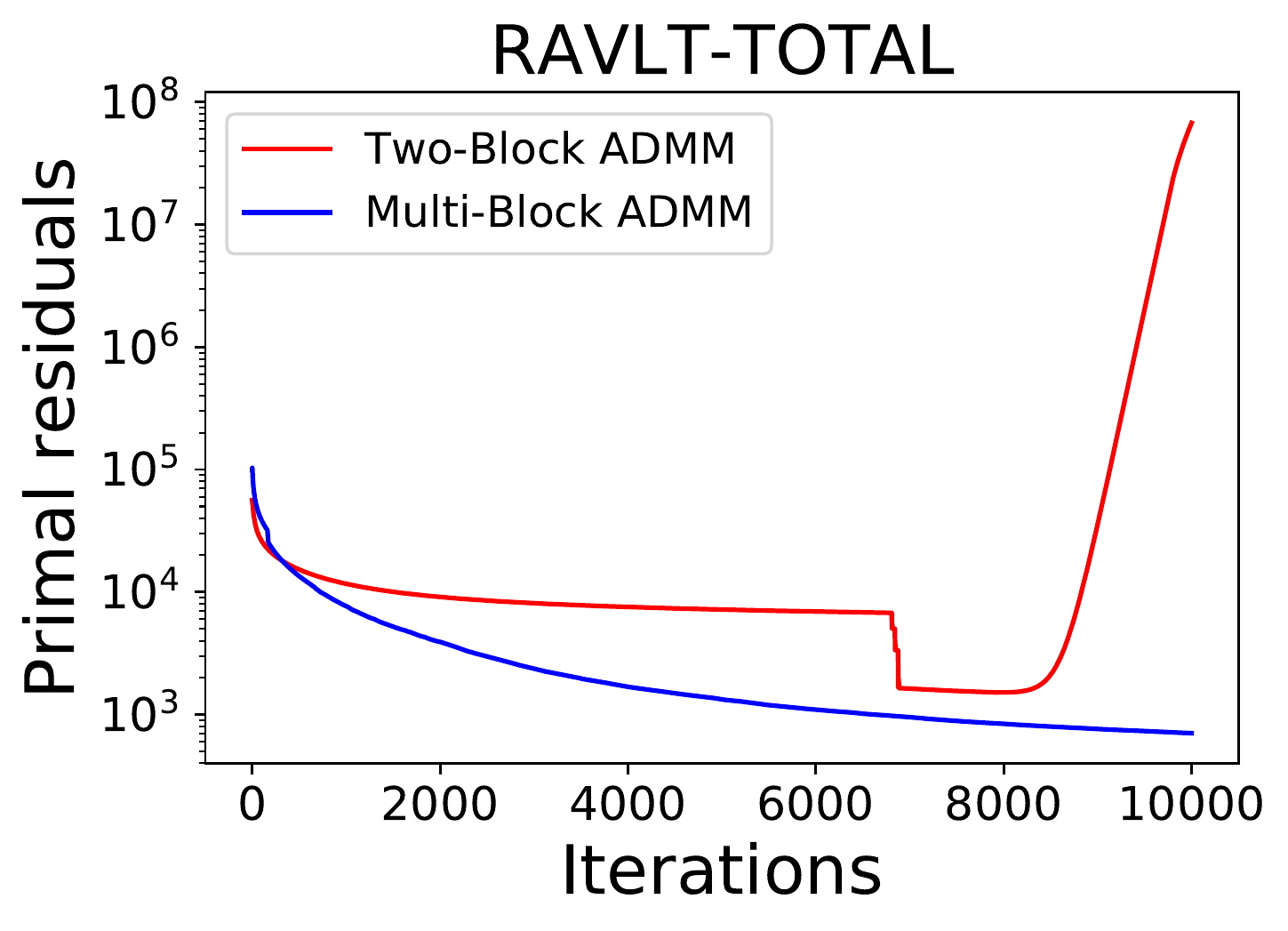}&
			\includegraphics[scale=0.25]{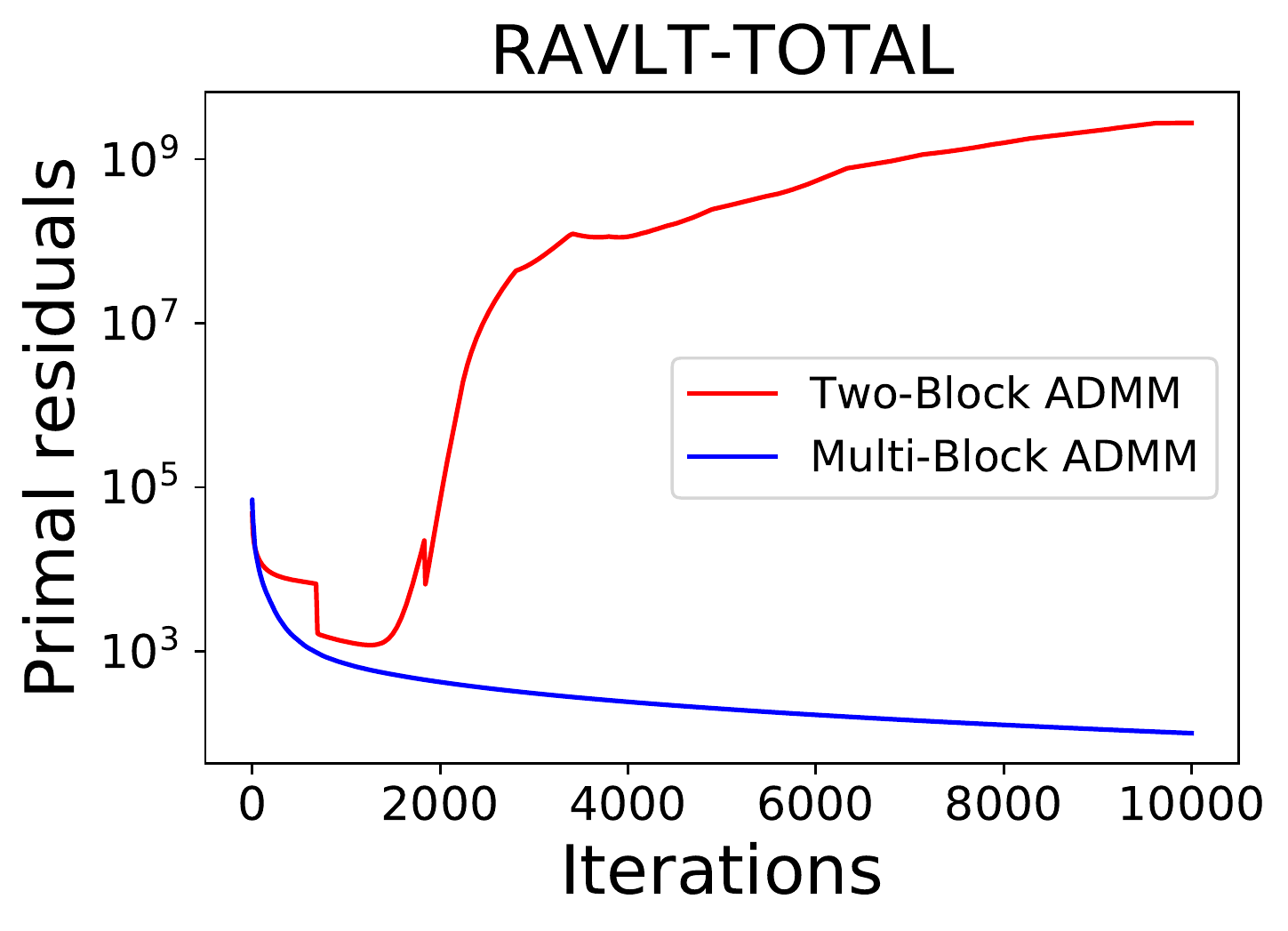}&
			\includegraphics[scale=0.25]{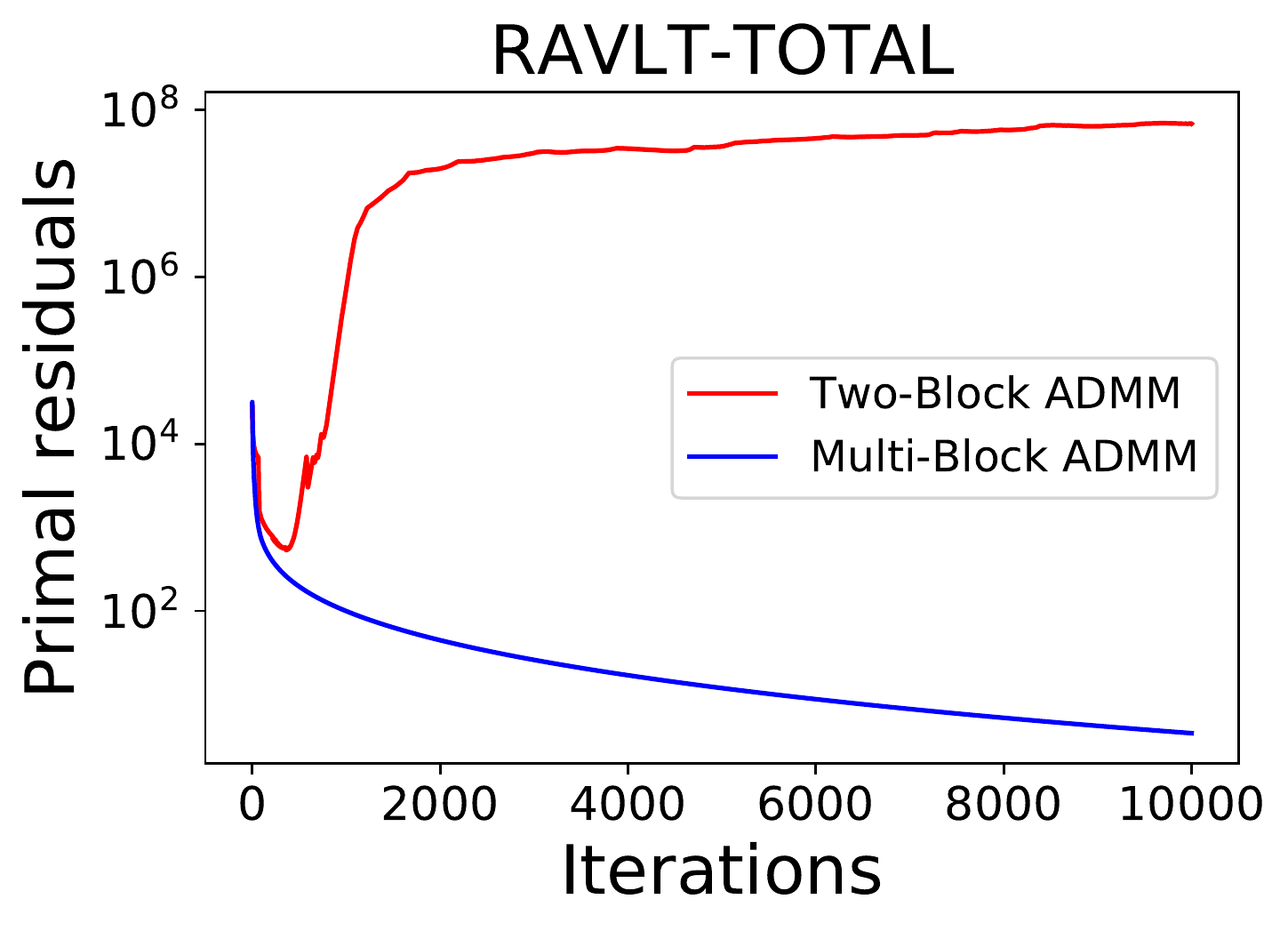}\\
			
	\end{tabular}}
	\caption{Average curves of primal residuals over 10 executions for all three cognitive scores with longer training time (10,000 iterations) for the ADNI dataset.  Multi-block shows a better convergence than two-block ADMM.}%
	\label{fig:convergence_adni1_longer_training}
\end{figure}

\noindent{\bf ADMM with longer training time:} A closer observation of Figure~\ref{fig:adni1_convergence}, particularly the method's convergence for smaller dual step sizes, $\rho \leq 0.01$, leads to the question whether the two-block ADMM would achieve similar performance to multi-block ADMM if we had continued the optimization process for a longer period of time. To answer this question, we extended the number of iterations for the ADMMs to 10,000 and generated the primal residual curves, as shown in Figure~\ref{fig:convergence_adni1_longer_training}. We observe that even for relatively small step sizes ($\rho=0.001$) the two-block ADMM still presents the same unstable behavior for MMSE and RAVLT-TOTAL cognitive scores. For ADAS, the two-block variant shows a slower convergence rate than the multi-block ADMM. \\

\noindent\textbf{Prediction Performance:} The final quantitative performance of the multitask learning regressors is measured by their performance on the held-out (test) data set. Table~\ref{tab:predictive_adni1} reports the rMSE per task (time point) and nMSE for the three cognitive scores investigated on ADNI dataset. To compute the prediction performance, we used the step size that led to the lowest  nMSE on the validation set.

\begin{table*}[htb]
	\caption{Prediction performance results of the three cognitive scores using the dataset ADNI. Multi-block ADMM produced more accurate predictions than the two-block version for all cognitive scores on both nMSE and rMSE.}
	\label{tab:predictive_adni1}
	\centering
	\scalebox{0.9}{
		\begin{tabular}{cccccccccc}
			\toprule[1.5pt]
			&& \multicolumn{2}{c}{ADAS} & & \multicolumn{2}{c}{MMSE} && \multicolumn{2}{c}{RAVLT-TOTAL} \\
			\cmidrule{3-4} \cmidrule{6-7}  \cmidrule{9-10}
			Time points && Two-Block & Multi-Block && Two-Block & Multi-Block && Two-Block & Multi-Block \\
			\cmidrule{1-1} \cmidrule{3-4} \cmidrule{6-7}  \cmidrule{9-10}
			nMSE  && 6.48$\pm$0.21 &\textbf{5.09$\pm$0.26} &&  2.81$\pm$0.08 & \textbf{2.24$\pm$0.08} && 8.99$\pm$0.25 & \bf{7.22$\pm$0.25}\\
			BL~~ rMSE  && 6.94$\pm$0.21 &\textbf{6.39$\pm$0.21} &&2.34$\pm$0.11 & \textbf{2.23$\pm$0.06} &&  9.66$\pm$0.34 & \textbf{9.20$\pm$0.41}\\
			M6~~ rMSE && 7.54$\pm$0.36 & \textbf{6.74$\pm$0.21} && 2.34$\pm$0.11 & \textbf{2.23$\pm$0.06} && 9.97$\pm$0.49 & \textbf{9.26$\pm$0.31}\\
			M12 rMSE && 8.71$\pm$0.60 & \textbf{7.64$\pm$0.52} && 3.41$\pm$0.20 & \textbf{2.97$\pm$0.18} && 10.92$\pm$0.36 & \textbf{9.62$\pm$0.37}\\
			M24 rMSE && 9.92$\pm$0.54 & \textbf{9.15$\pm$0.69} && 4.20$\pm$0.31 & \textbf{3.71$\pm$0.27} && 11.82$\pm$0.28 & \textbf{10.39$\pm$0.31}\\
			M36 rMSE && 9.24$\pm$0.35 & \textbf{7.94$\pm$0.44} && 3.38$\pm$0.29 & \textbf{3.11$\pm$0.19} && 10.58$\pm$0.62 & \textbf{9.03$\pm$0.40}\\
			M48 rMSE && 11.50$\pm$1.85 & \textbf{7.59$\pm$1.12} && 5.74$\pm$0.56 & \textbf{3.64$\pm$0.62} && 15.12$\pm$1.69 & \textbf{10.70$\pm$0.66}\\
			\bottomrule[1.5pt]
	\end{tabular}}
\end{table*}

Reflecting the performance on the validation set, the multitask learning formulation optimized by the multi-block ADMM produced lower error on the test set. The improvement is noticed for all cognitive scores and time points (tasks), but is more pronounced in the last time points (M36 and M48), which contains the smallest amount of samples. These results reaffirm the superior performance of the multi-block over the two-block ADMM in our multitask learning formulation for the Alzheimer's Disease progression problem.

\subsection{Parkinson's disease assessment}
\label{sec:parkinson}


For this experiment, we used data from the Parkinson's Progression Marker Initiative (PPMI)\footnote{https://www.ppmi-info.org} repository. The dataset includes information from healthy individuals and patients diagnosed with Parkinson's disease. Biospecimen and demographics information are collected from all patients at the beginning of the study, which is referred to \textit{baseline} (BL), and cognitive assessments are performed in each scheduled visit. Subject's cognitive function is measured by the MDS-UPDRS-Total score. Not all patients attended all visits, causing some time steps to have a smaller amount of patients. Table~\ref{tab:number_patients_visits_pd} shows the number of patients which had their cognitive state assessed at each visit.


\begin{table}[htb]
	\caption{Parkinson's dataset: Number of patients (samples) at each time step (task).  MX correspond to the number of months X after the baseline screening was taken.}
	\centering
	\begin{tabular}{ccccccccccccccc}
		\toprule
		\multirow{2}{*}{Baseline} & \multicolumn{14}{c}{Months after baseline} \\ \cline{2-15}
		& M3 & M6  & M9 & M12 & M18 & M24 & M30 & M36 & M42 & M48 & M54 & M60 & M72 & M84 \\ \hline
		185 & 99 &  99 & 90 & 199 &  97 & 195 &  92  & 209 & 92 & 216 & 80 & 214 & 121 & 82 \\		\bottomrule
	\end{tabular}
	\label{tab:number_patients_visits_pd}
\end{table}

A set of biological specimens are used to predict the cognitive state of the patient at a certain time in the future (visits). These variables have been identified in the literature as potential biomarkers for Parkinson's disease progression  \cite{Khoo2012} and \cite{Parnetti2013}. The variables used are listed below.
\begin{description}
	\item[Cerebrospinal Fluid:] ABeta 1-42, pTau, tTau, CSF Alpha-synuclein, p-tau/t-tau, p-tau/Abeta 1-42, and t-tau/Abeta 1-42.
	\item[Plasma:] Total Cholesterol, HDL, Triglycerides, Apolipoprotein A1, LDL, and EGF ELISA.
	\item[Serum:] Serum IGF-1.
	\item[RNA:]  UBC, DJ-1, SNCA-3UTR-2, RPL13, FBXO7-001, SNCA-3UTR-1, DHPR,
	GLT25D1, FBXO7-007, MON1B, SNCA-E4E6, SNCA-007, FBXO7-005,
	SRCAP, FBXO7-010, ZNF746, GUSB, SNCA-E3E4, and FBXO7-008.
\end{description}

Besides the biospecimen variables, we include demographics information which may be relevant for the prediction task. Four variables are incorporated: race, family history, gender, and age. Family history is a binary variable informing whether any family member has been diagnosed with PD.

Based on the patients information at the BL time step, the goal is to predict patient's cognitive function at the following time-steps (visits).  Therefore, a predictive model with sparsity structures will be able to identify the variables that most explain the evolution of patient's cognitive functions. From the multitask learning perspective, predicting the MDS-UPDRS-Total score for each time step (visit) is considered as a task.

\begin{figure}[htb]
	\centering
	\begin{tabular}{ll}
		\includegraphics[scale=0.3]{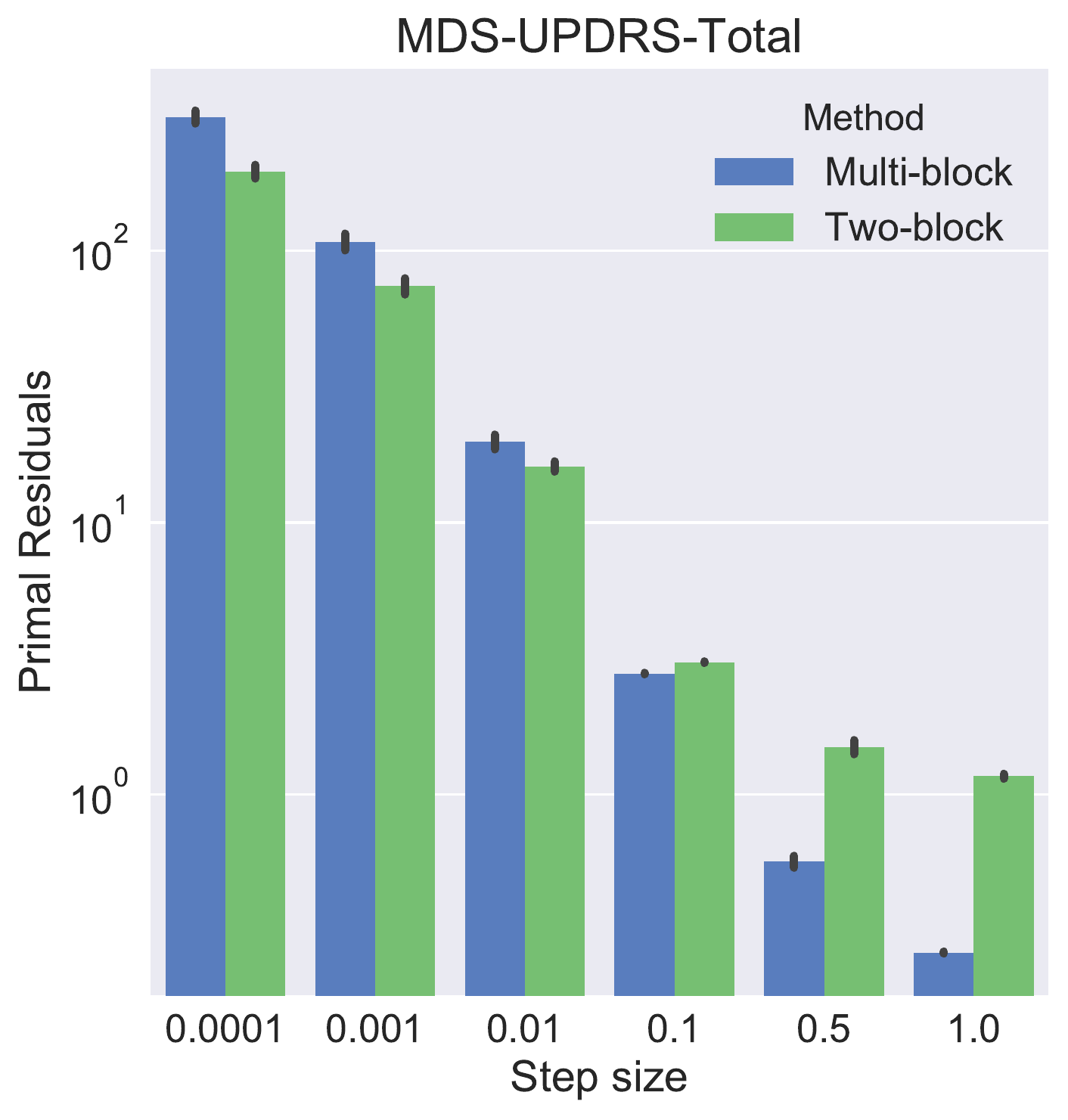}
		\includegraphics[scale=0.3]{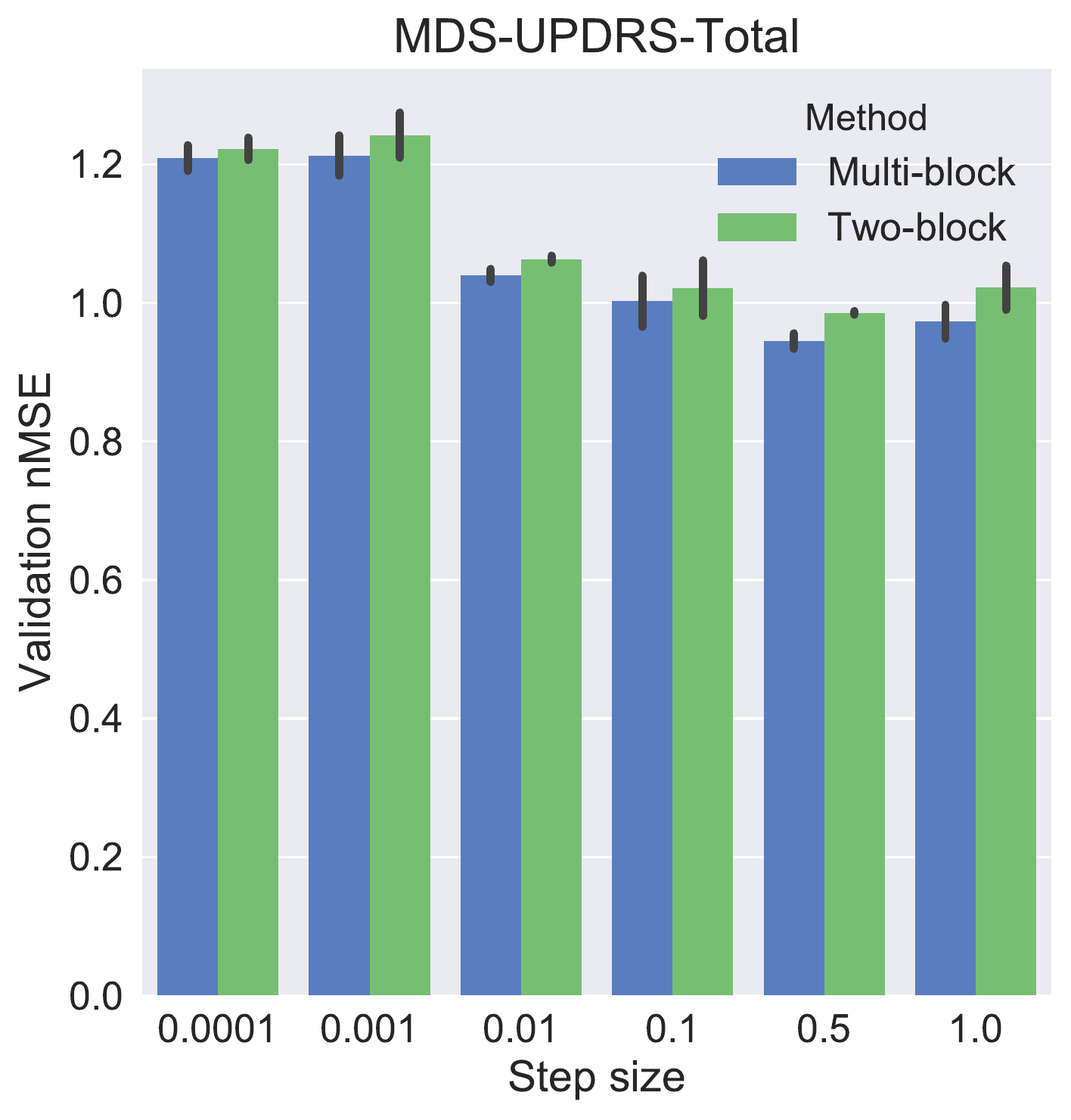}
	\end{tabular}
	\caption{Comparison of two-block and multi-block ADMM for TS-MTL formulation on Parkinson's disease dataset. Validation nMSE and primal residual are computed as the average over the last 100 iterations of the ADMM. Deviation bars show the variation over 10 independent executions of the methods with different train and test data split.}%
	\label{fig:concise_convergence_parkinson}
\end{figure}

We compared the performance of TS-MTL with two-block and multi-block ADMM for several values for the dual step size $\rho$ and the results are showed in Figure~\ref{fig:concise_convergence_parkinson}. It is worth mentioning that the two-block ADMM diverged for $\rho > 1 $, while multi-block ADMM converged for larger values of $\rho$. We notice that multi-block ADMM achieved consistently lower validation nMSE for the entire range for the dual step size $\rho$. Multi-block primal residuals are significantly smaller for $\rho$ that led to best validation nMSEs ($\rho = 0.5$ and $\rho=1$).

\begin{figure}[htb]
	\centering
	\scalebox{0.9}{
		\begin{tabular}{ccc}
			$\boldsymbol{\rho}=0.01$ & $\boldsymbol{\rho}=0.5$ & $\boldsymbol{\rho}=1$ \\
			\includegraphics[scale=0.25]{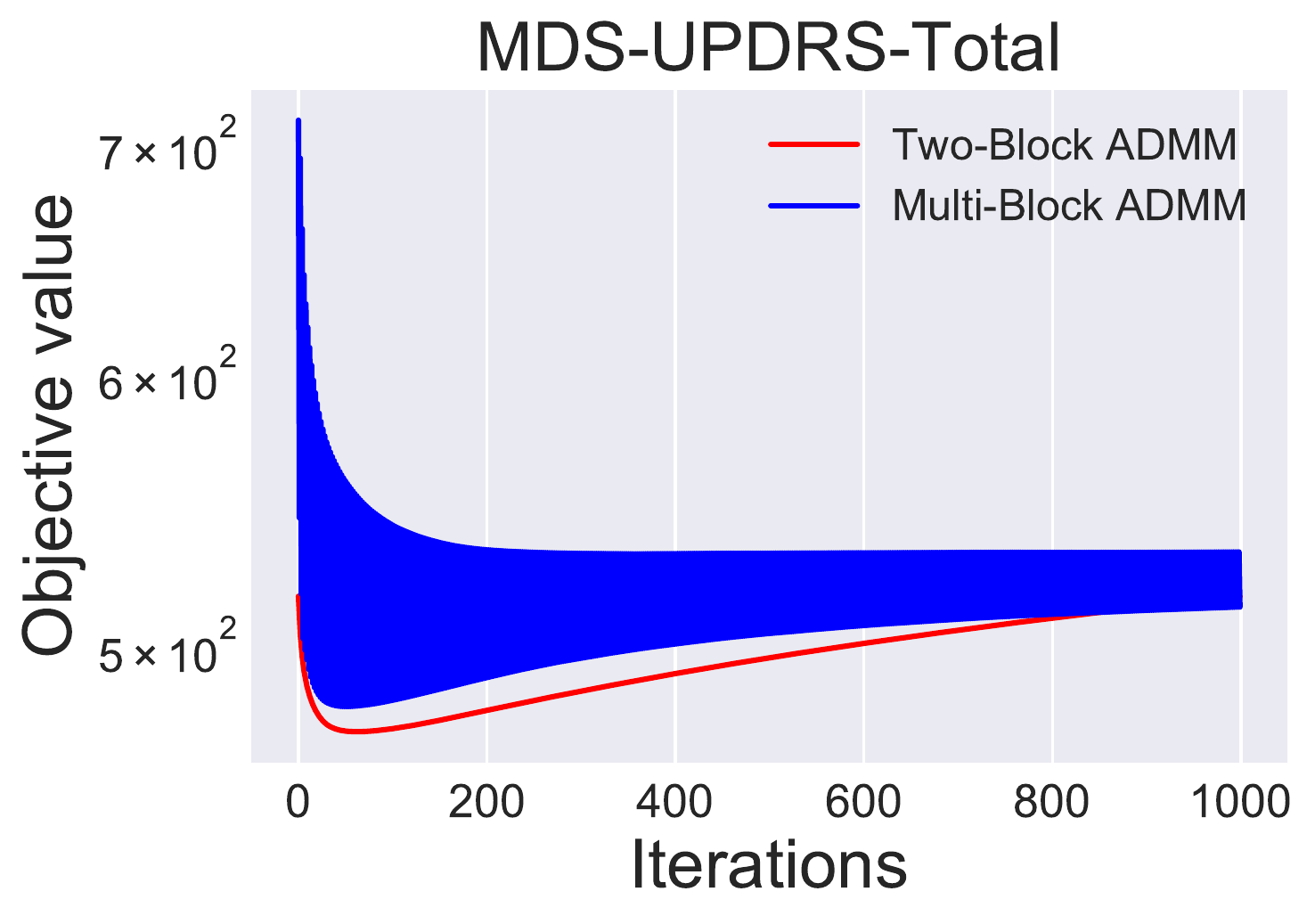}&
			\includegraphics[scale=0.25]{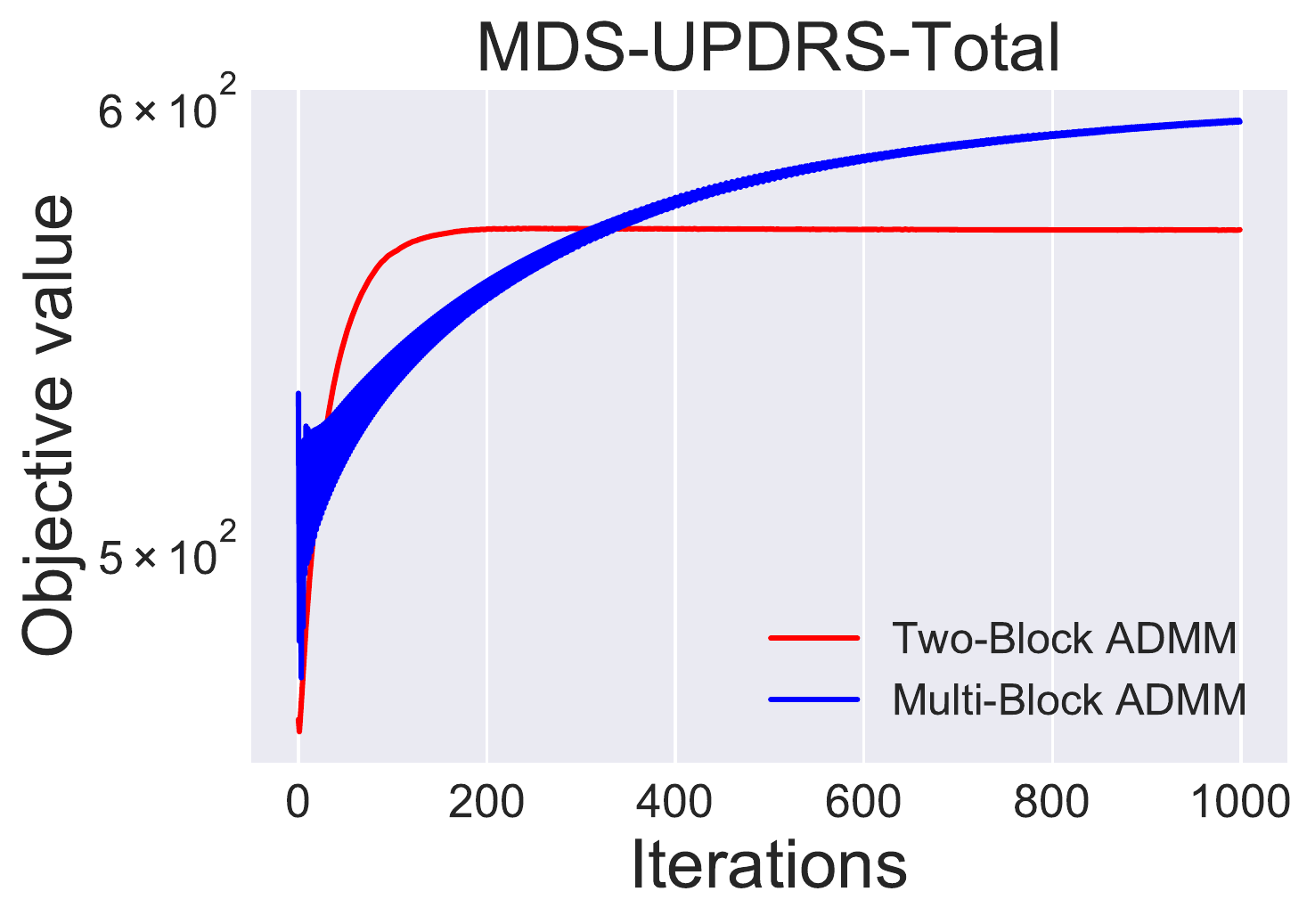}&
			\includegraphics[scale=0.25]{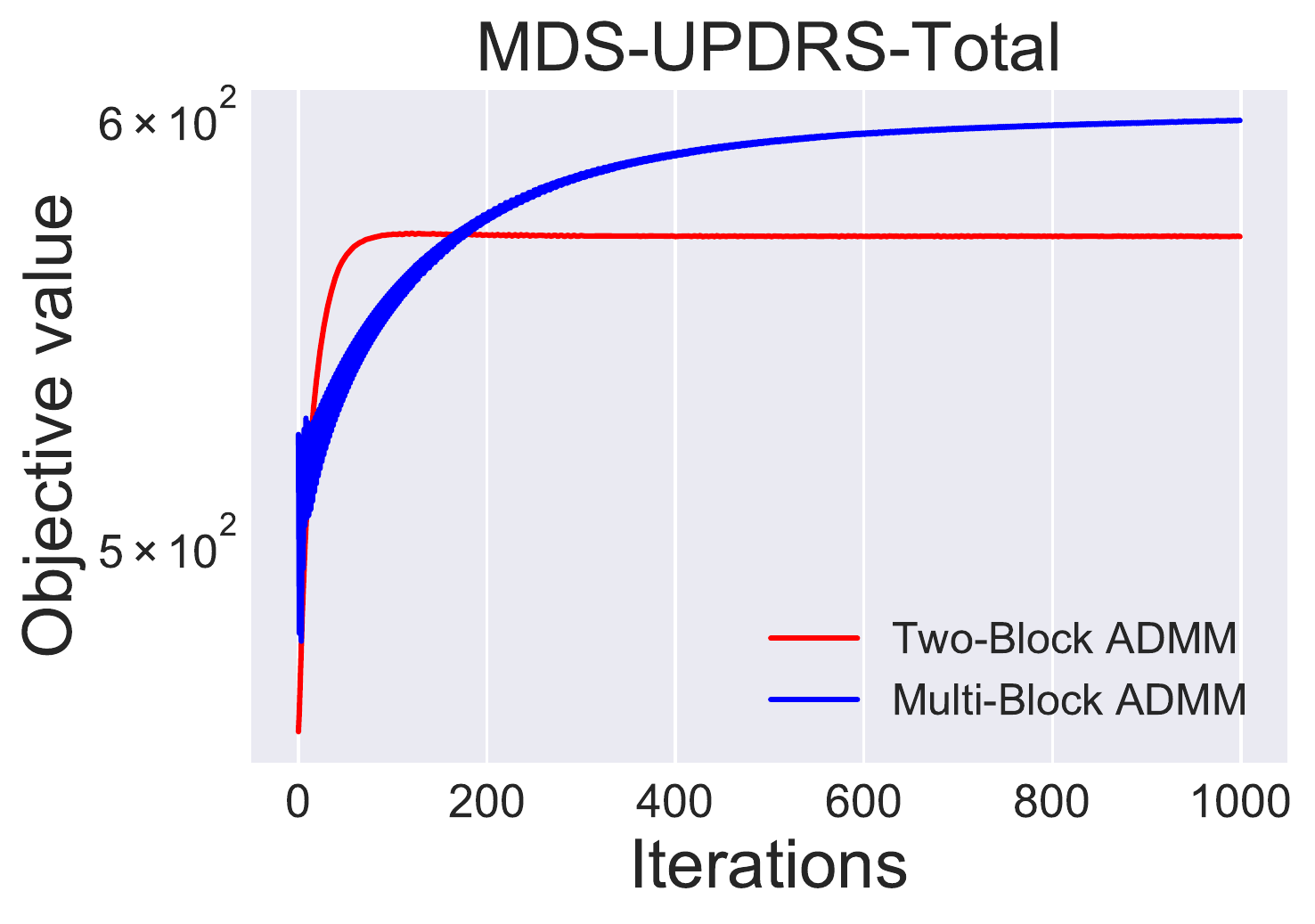}\\
			
			\includegraphics[scale=0.25]{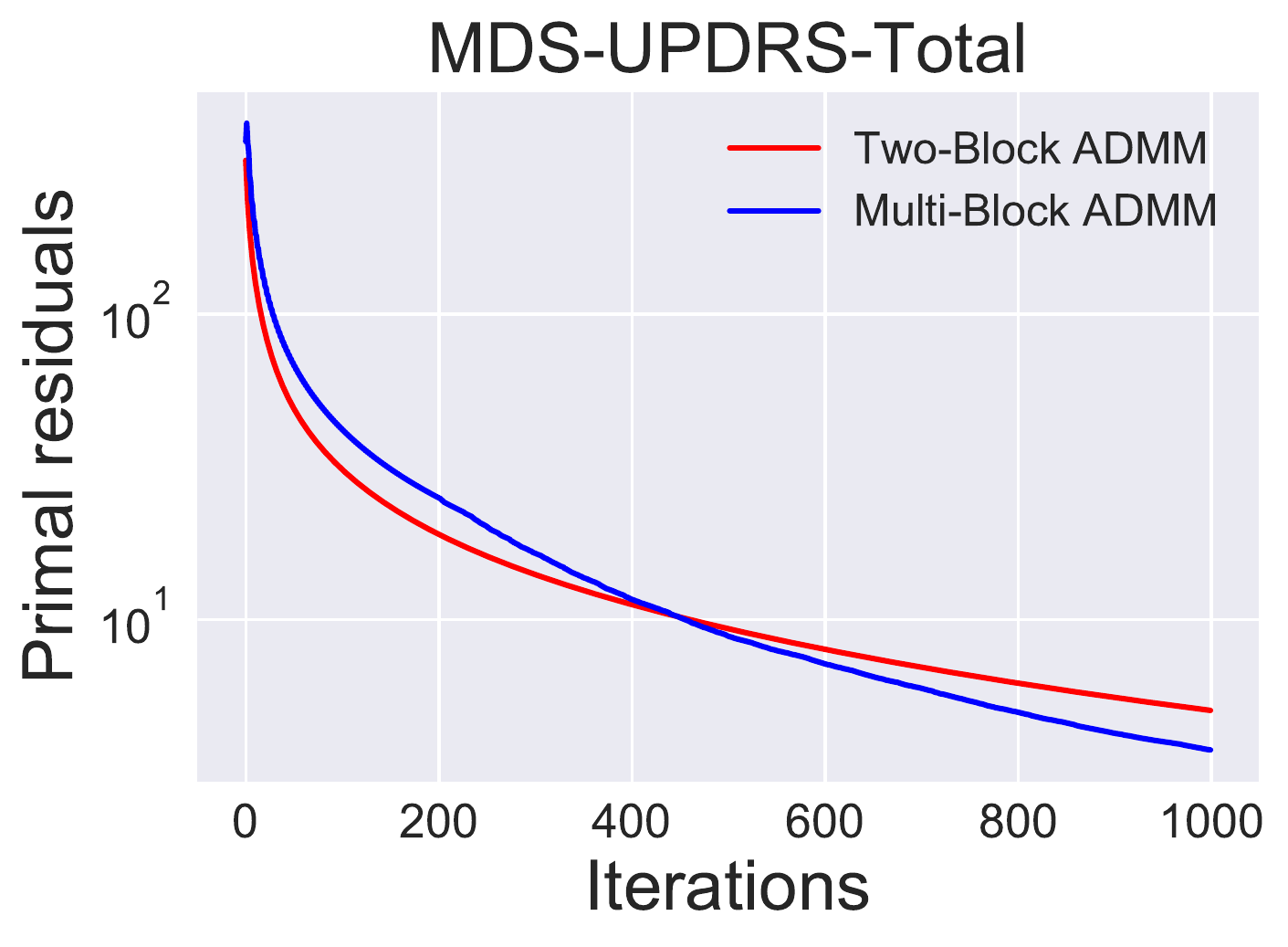}&
			\includegraphics[scale=0.25]{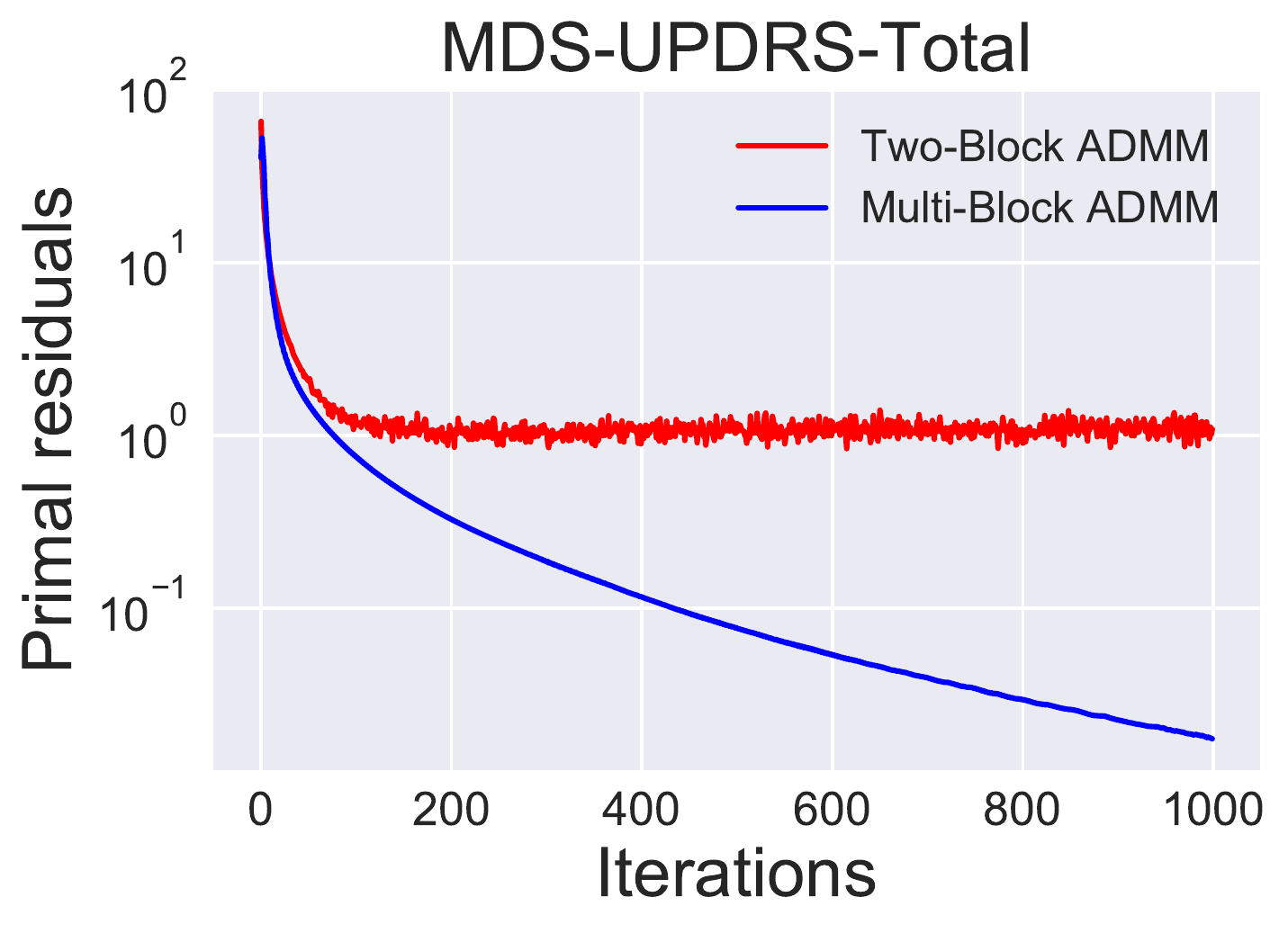}&
			\includegraphics[scale=0.25]{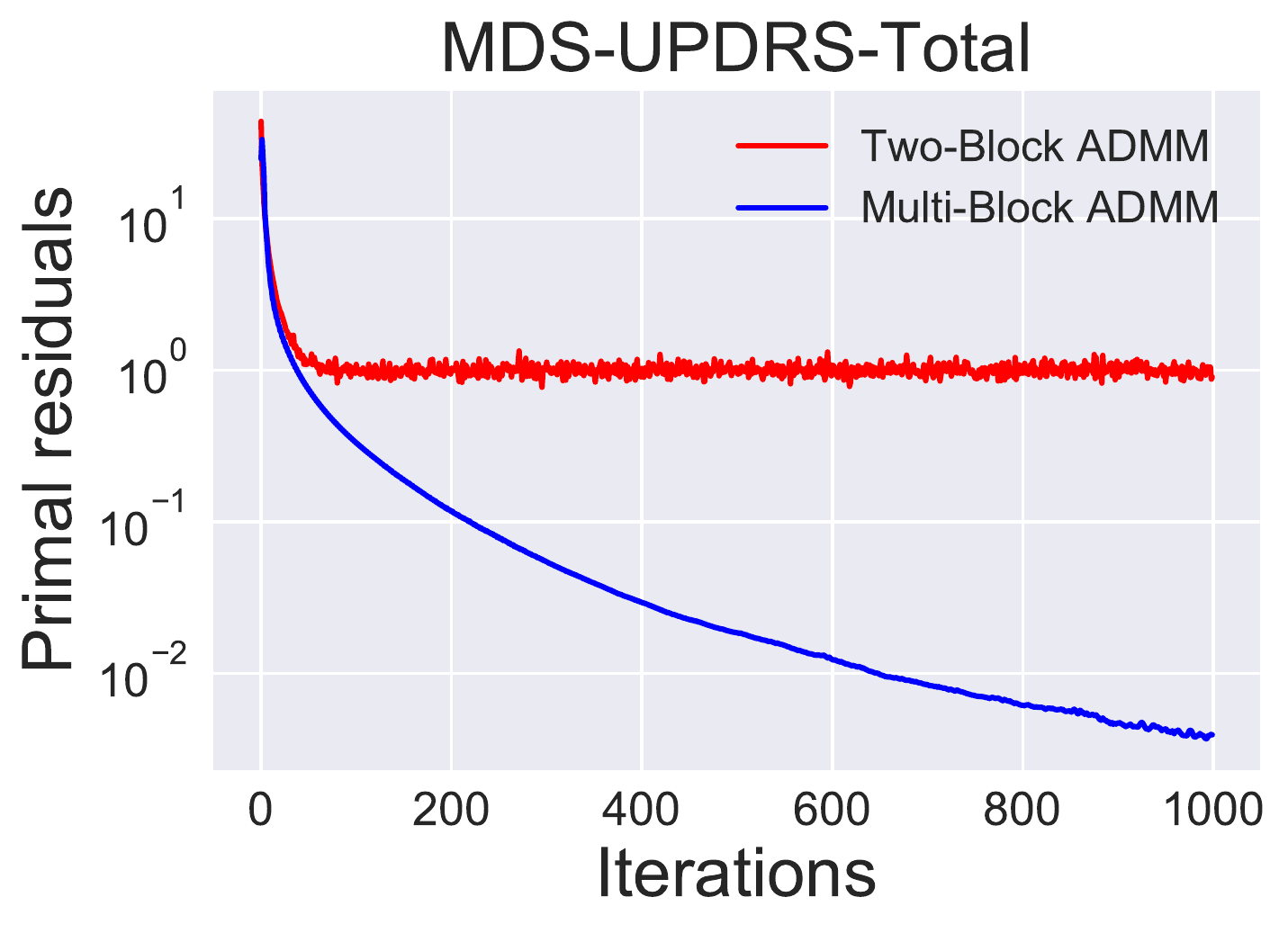}\\
			
			\includegraphics[scale=0.25]{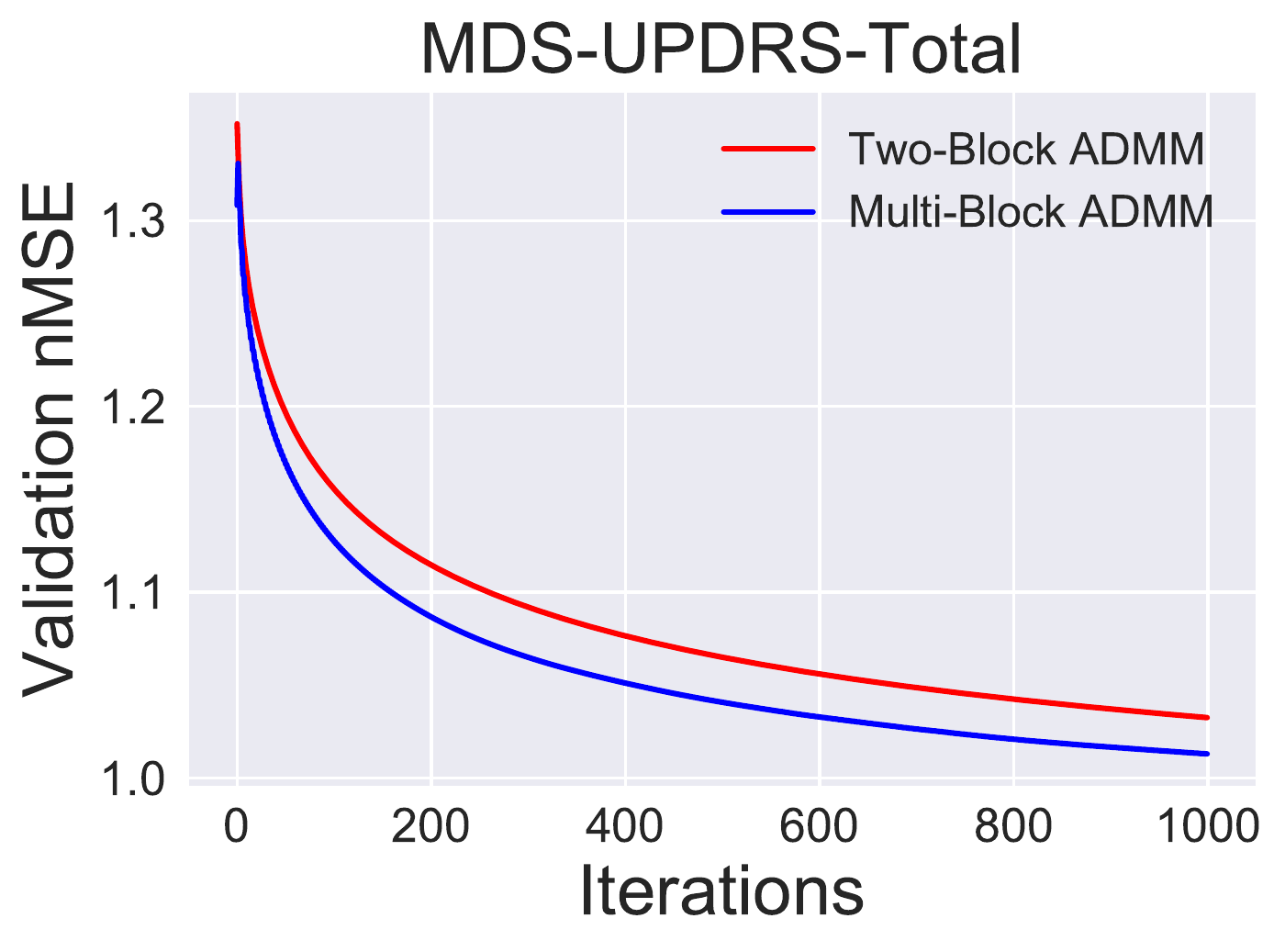}&
			\includegraphics[scale=0.25]{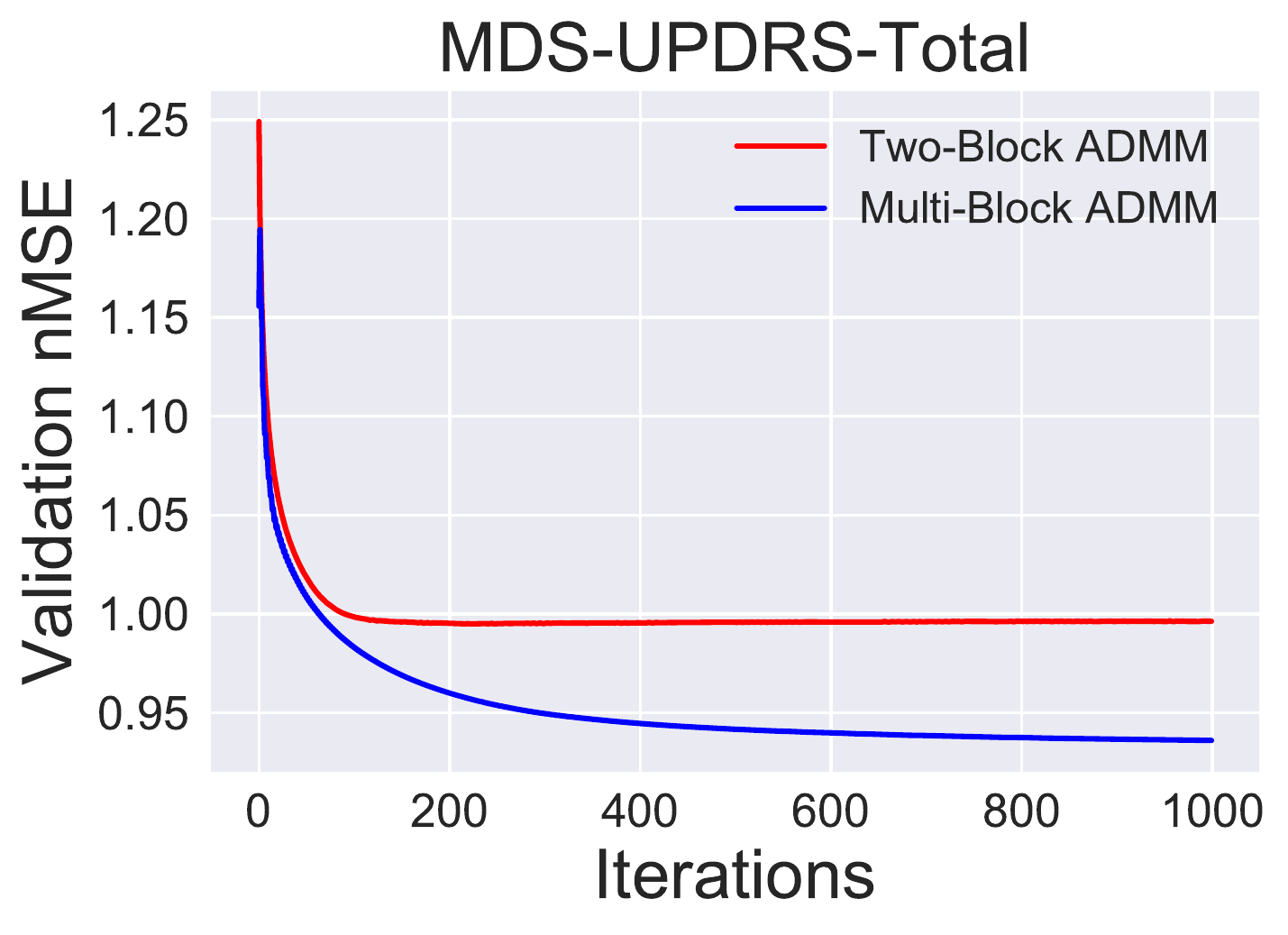}&
			\includegraphics[scale=0.25]{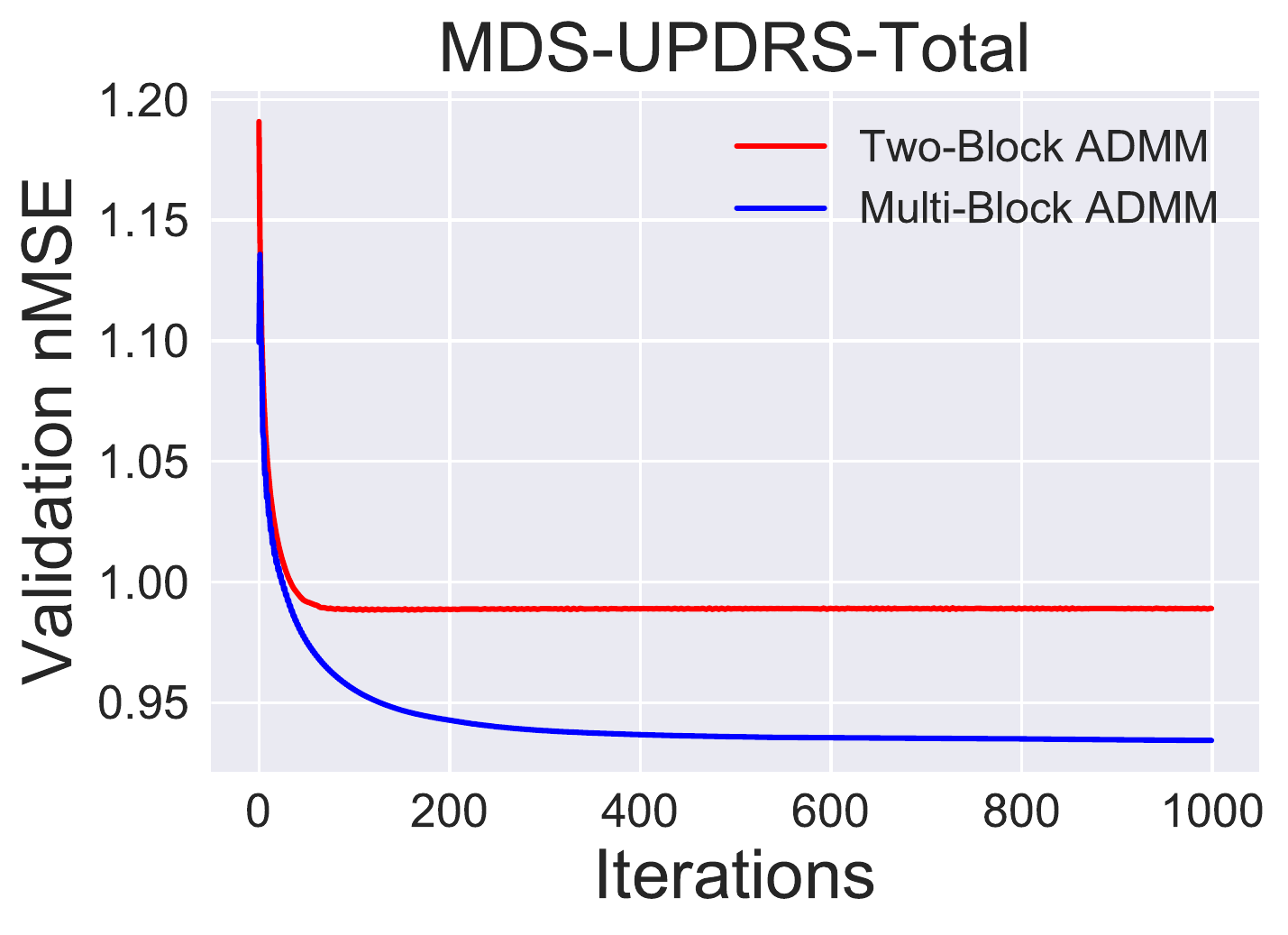}\\
			
	\end{tabular}}
	\caption{Average curves of objective function, primal residuals, and validation nMSE over 10 executions on the Parkinson's disease dataset. Many values for learning rate $\rho$ were investigated.  Multi-block shows a better convergence than two-block ADMM.}%
	\label{fig:convergence_pd}
\end{figure}

Convergence curves for both ADMM methods on PD dataset are shown in Figure~\ref{fig:convergence_pd}. Notably, multi-block ADMM achieved lower primal residuals and validation nMSE. For $\rho = 0.5$ and $\rho = 1$, two-block ADMM stagnates and could not longer reduce primal residual and validation nMSE. It is highly probable that even if more iterations were performed, the two-block ADMM would not obtain results comparable to the multi-block version.

\subsection{Air quality prediction}
\label{sec:air_quality}
For the Air quality dataset \footnote{https://archive.ics.uci.edu/ml/datasets/Air+quality} the goal is to predict CO concentration for a certain hour of the day given measurements from 5 distinct sensors + temperature and humidity information. In this setup, we train one regression model for each hour of the day. Although the 24-th hour is connected to 1st hour of the next day, we are not considering this circular dependence for this experiment.

\begin{figure}[htb]
	\centering
	\begin{tabular}{ll}
		\includegraphics[scale=0.3]{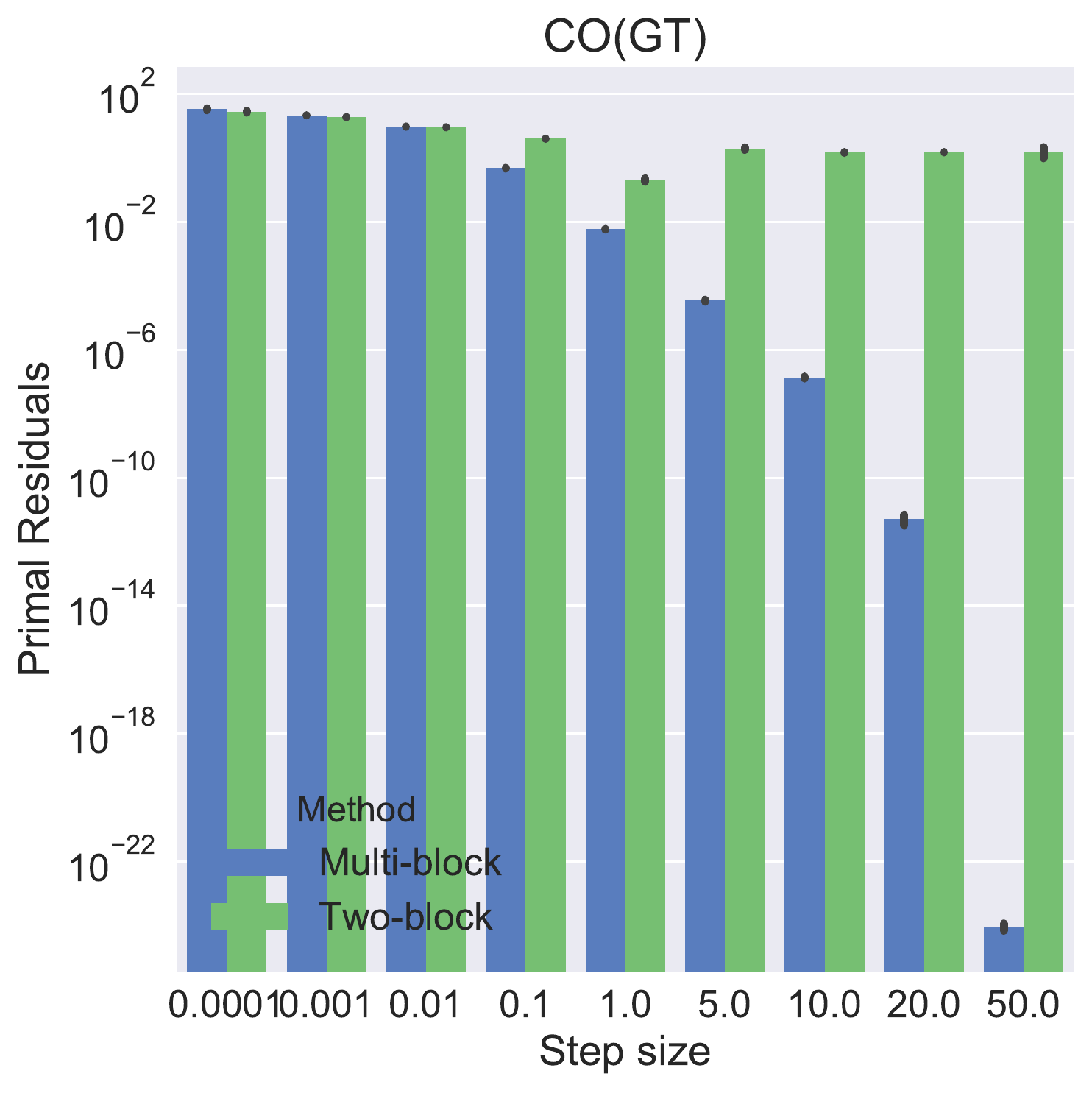}  \qquad
		\includegraphics[scale=0.3]{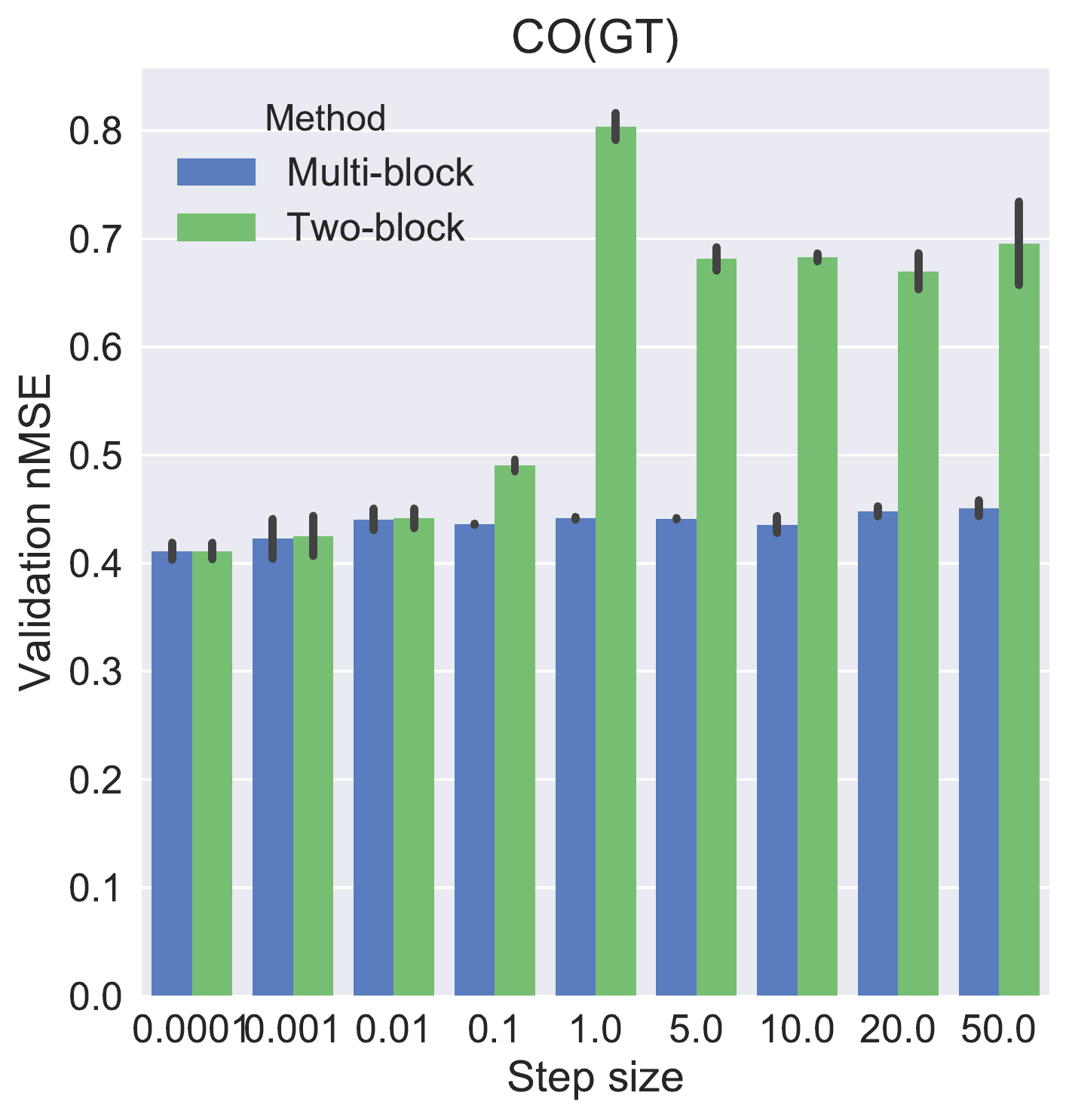}
	\end{tabular}
	\caption{Comparison of two-Block and multi-Block ADMM for TS-MTL formulation on Air Quality data. Multi-block shows similar or better performance than two-block ADMM.}%
	\label{fig:concise_air_quality}
\end{figure}

\begin{figure}[htb]
	\centering
	\begin{tabular}{ccc}
		$\boldsymbol{\rho}=0.001$ & $\boldsymbol{\rho}=0.01$ & $\boldsymbol{\rho}=1$ \\ \hline
		\includegraphics[scale=0.22]{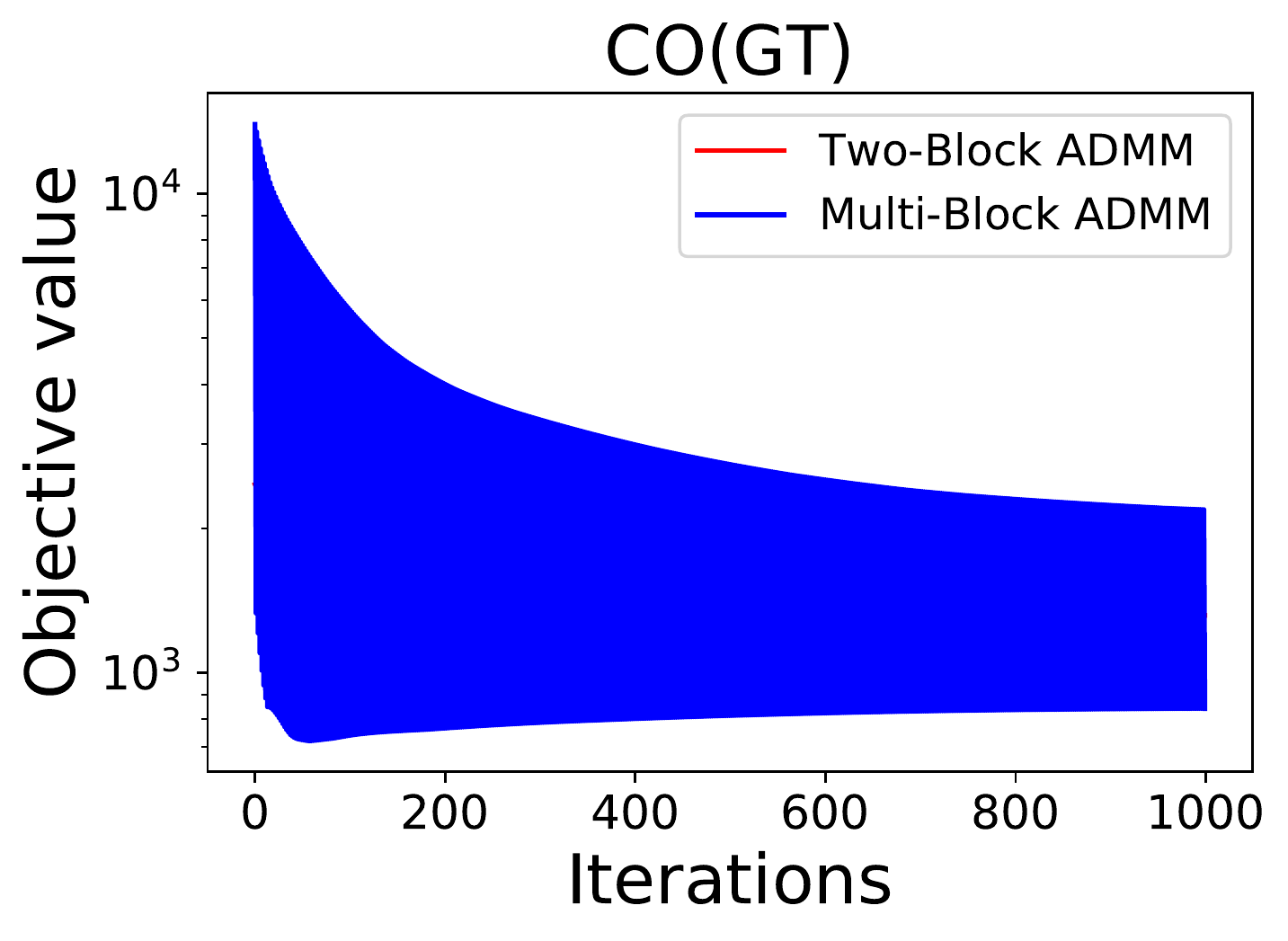}&
		\includegraphics[scale=0.22]{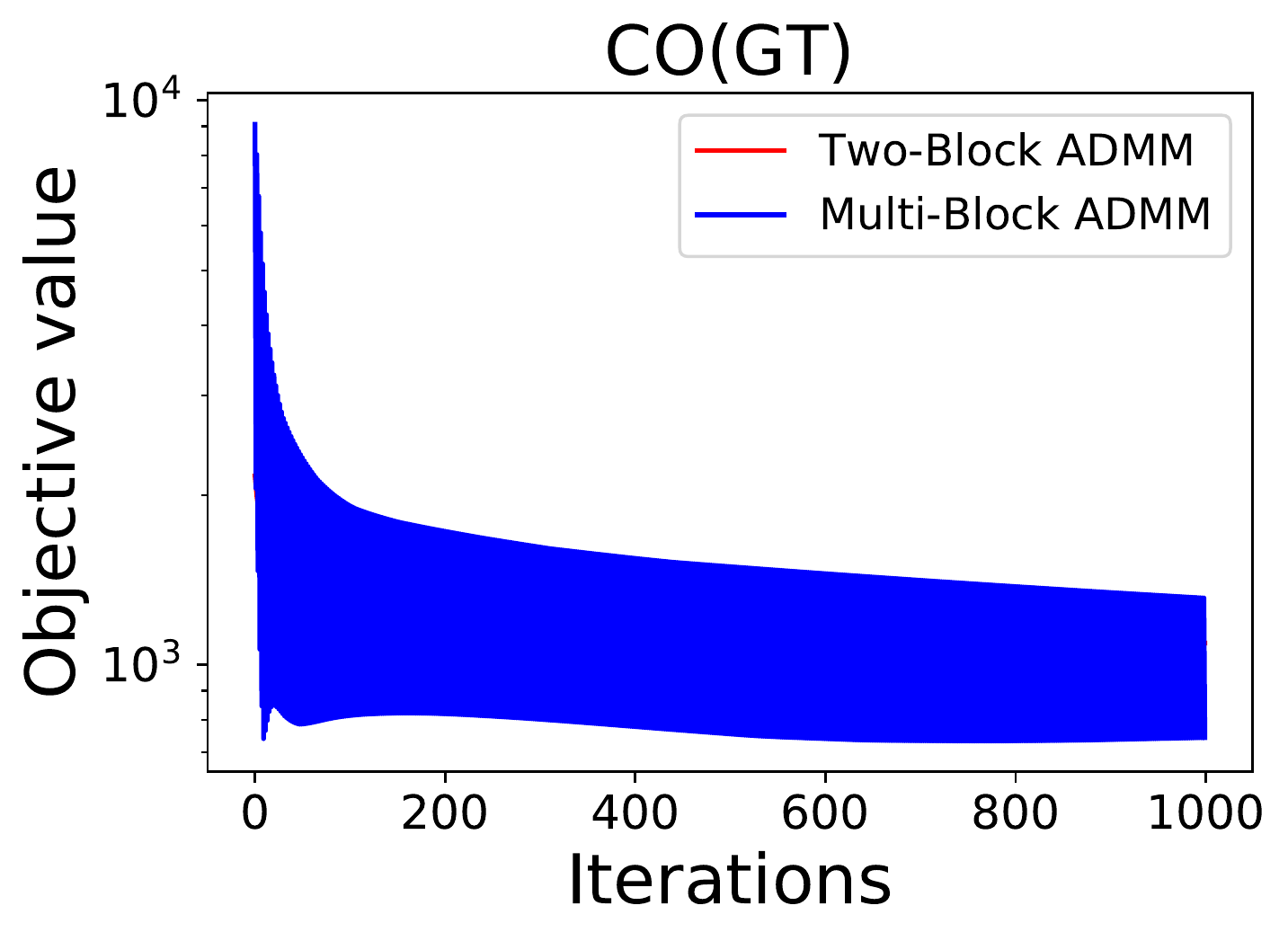}&
		\includegraphics[scale=0.22]{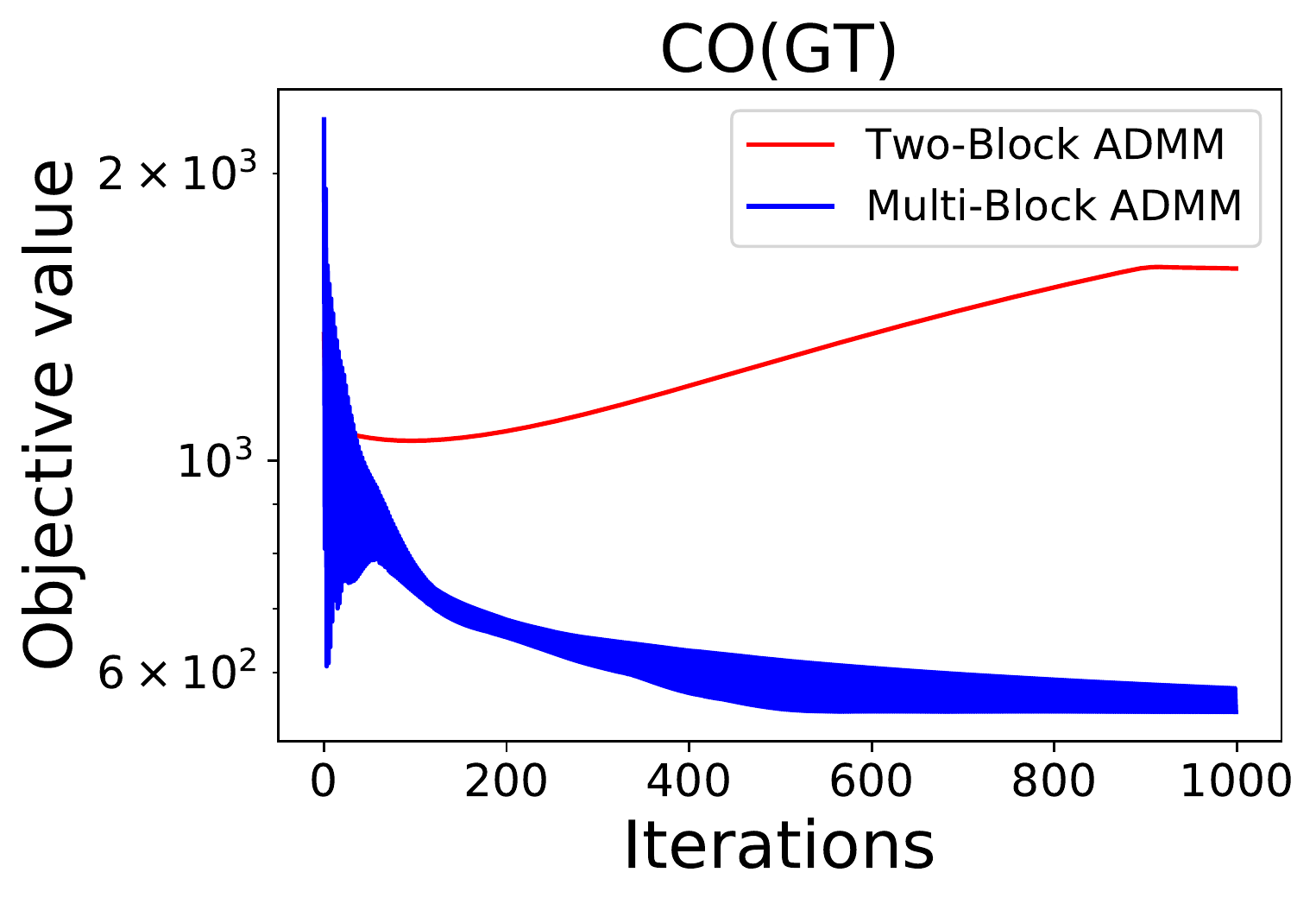}\\
		
		\includegraphics[scale=0.22]{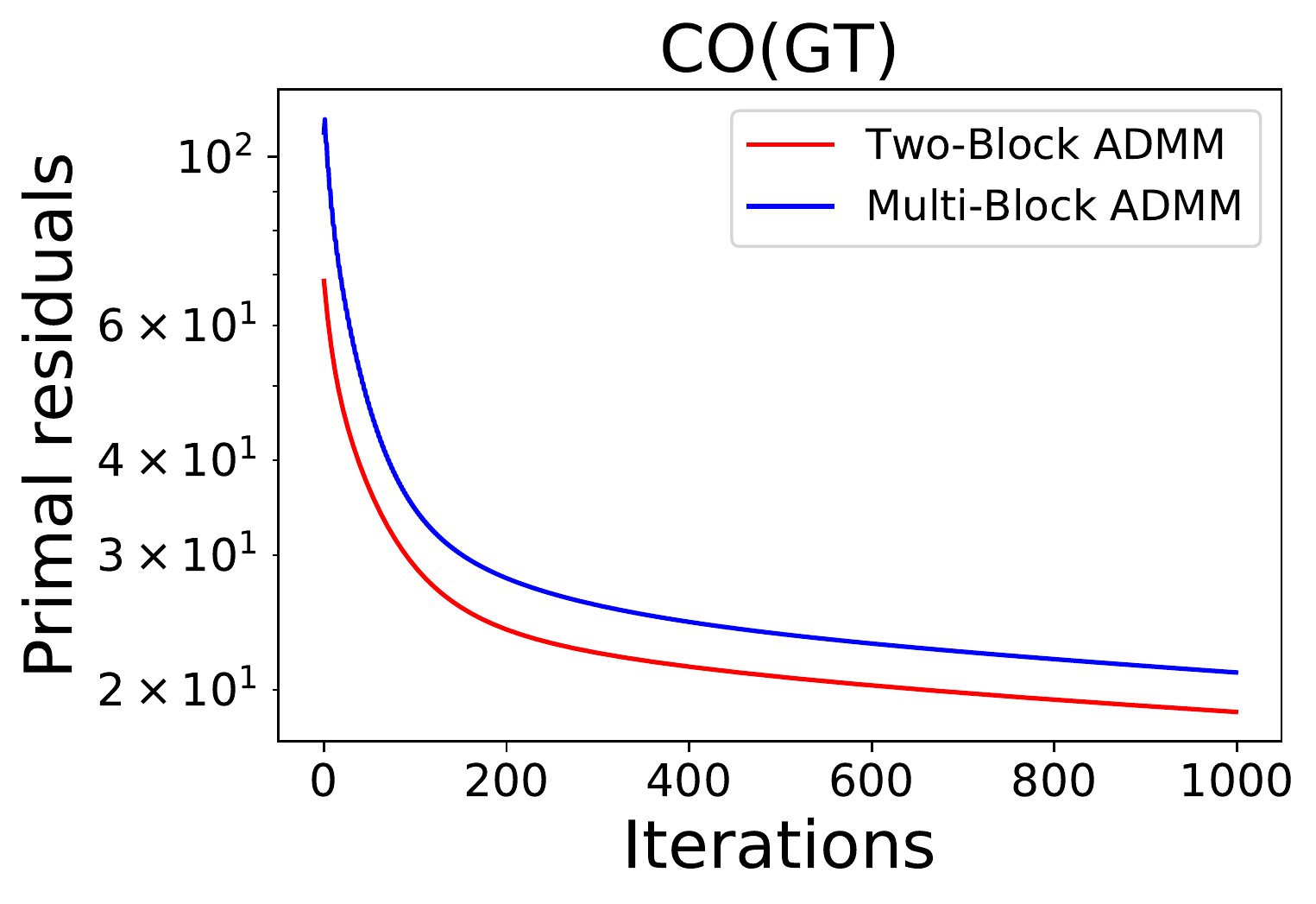}&
		\includegraphics[scale=0.22]{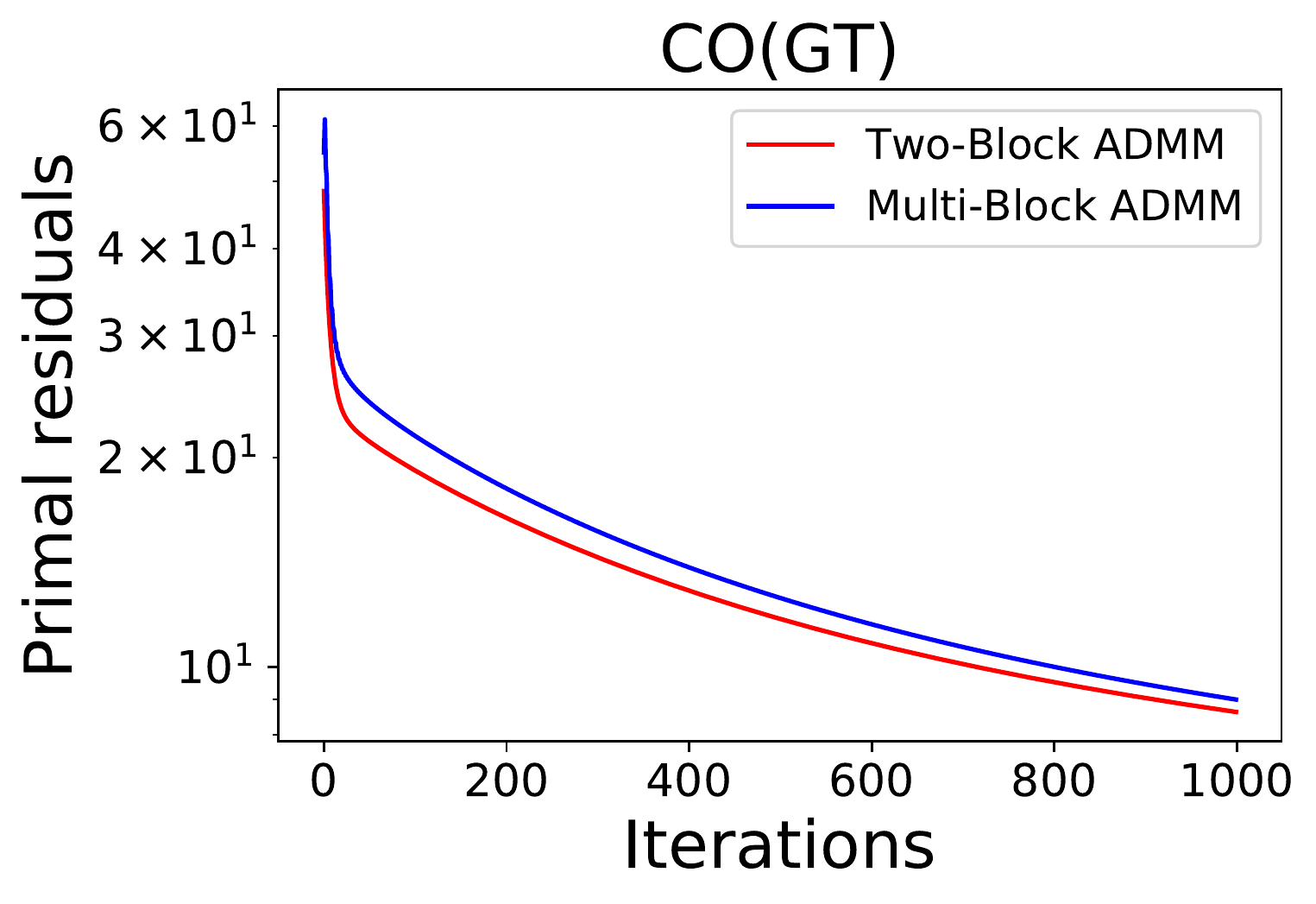}&
		\includegraphics[scale=0.22]{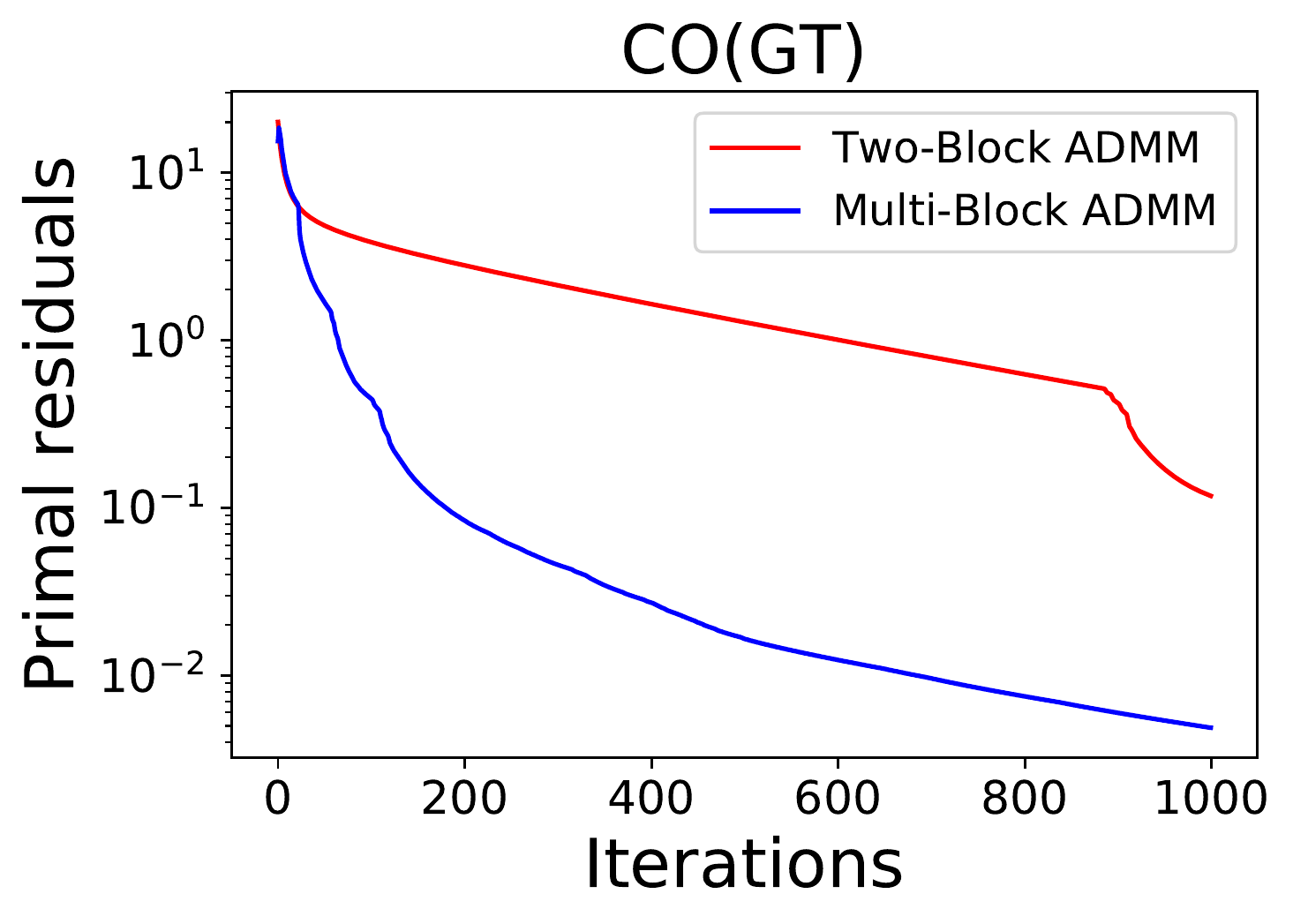}\\
		
		\includegraphics[scale=0.22]{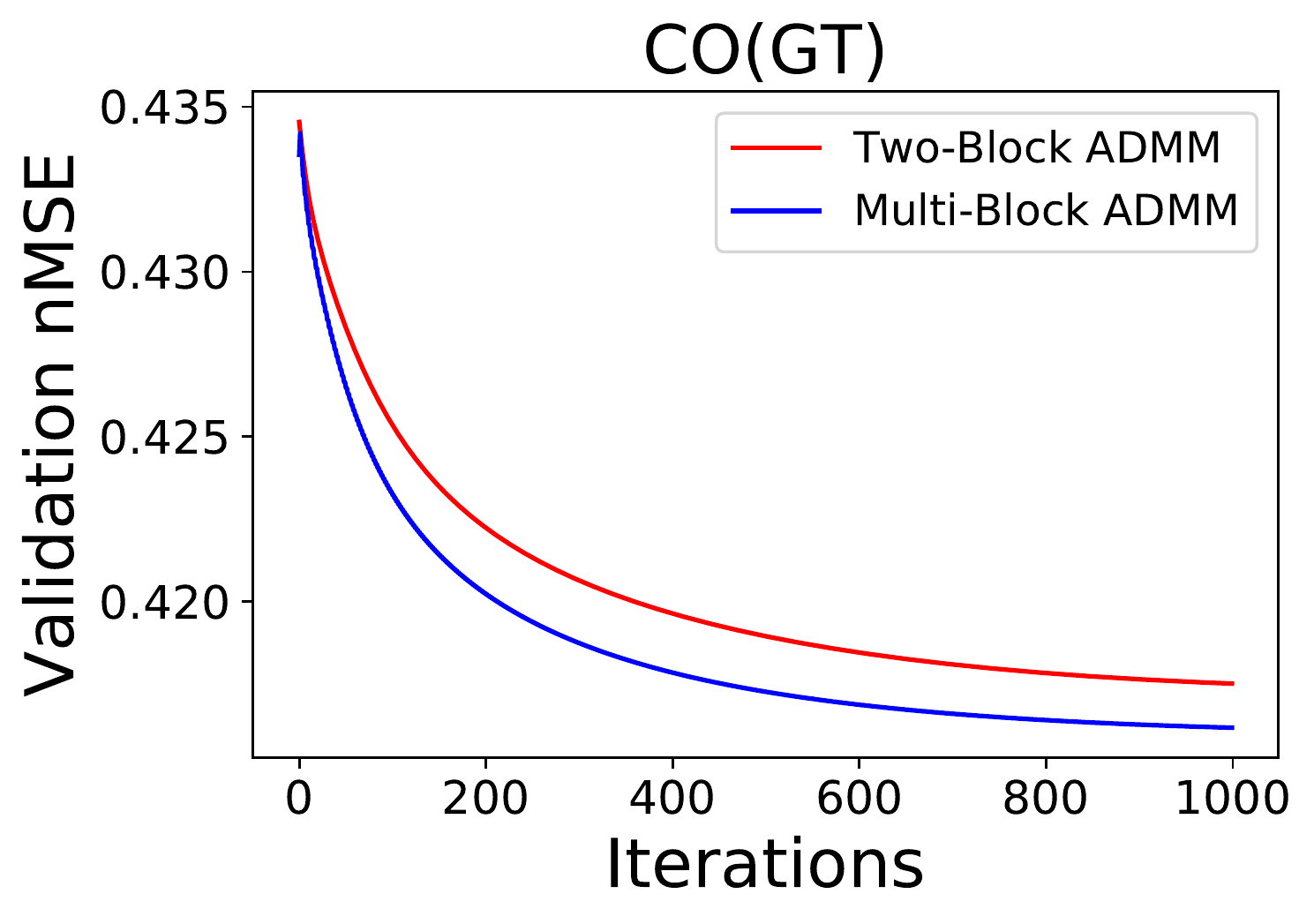}&
		\includegraphics[scale=0.22]{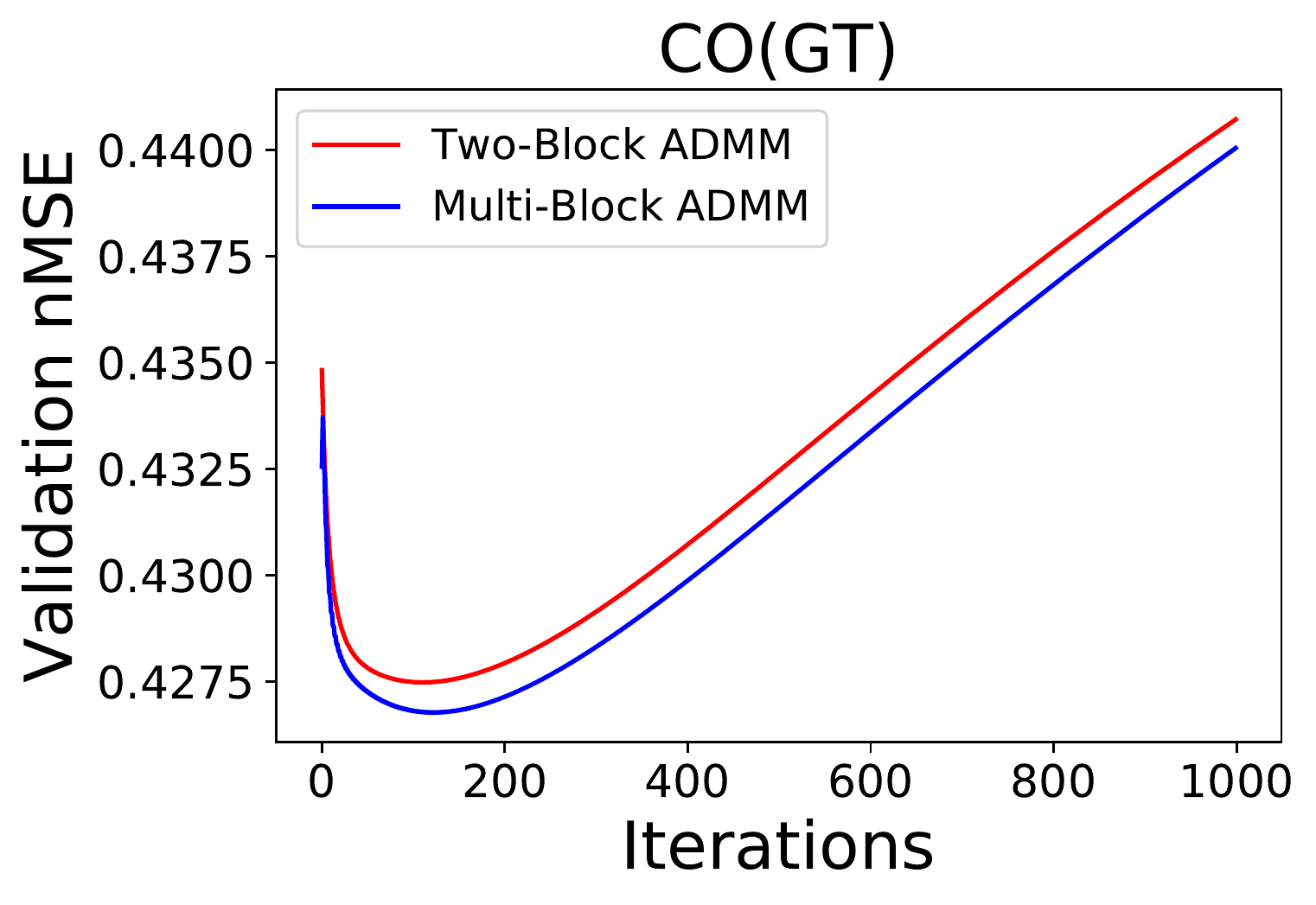}&
		\includegraphics[scale=0.22]{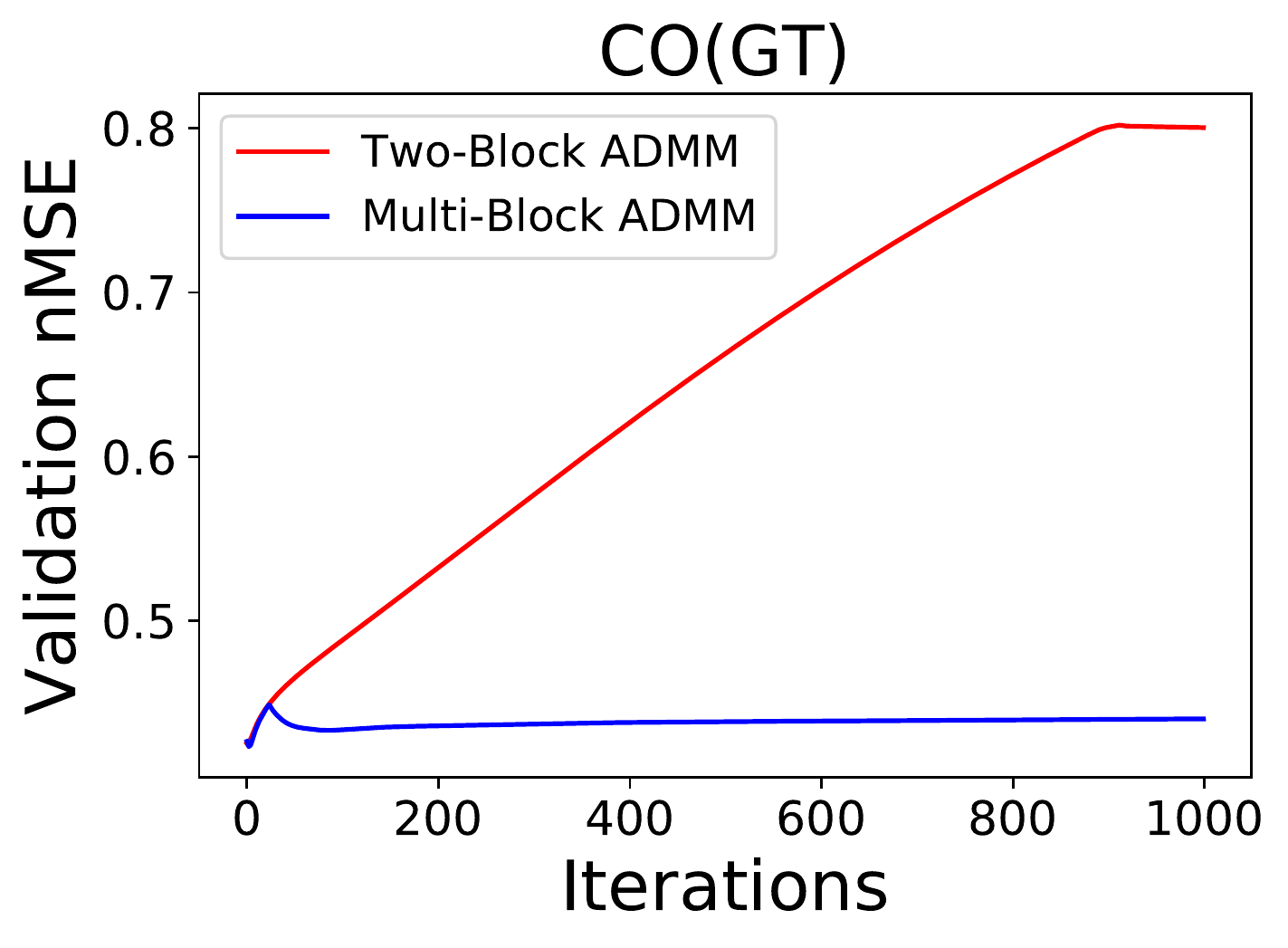}\\
	\end{tabular}
	\caption{Average curves of objective value, primal residuals, and validation nMSE over 10 executions for Air Quality dataset.  Multi-block shows a better convergence than two-block ADMM.}%
	\label{fig:convergence_air_quality}
\end{figure}

Similar to the other datasets, we conducted experiments with TS-MTL using both two-block and multi-block ADMM for several values of the dual step size $\rho$, and the results are presented in Figures~\ref{fig:concise_air_quality} and \ref{fig:convergence_air_quality}.
Again, multi-block shows a steady reduction of primal residuals during optimization, along with a more stable validation performance curves. Although multi-block does not present significant lower nMSE on the validation set in the range of values for $\rho \in [1e^{-4}, 1e^{-2}]$, it is clearly better for larger values (Figure~\ref{fig:concise_air_quality} bottom).

\section{Conclusions}
\label{sec:conc}

In this paper, we reported an experimental study on the convergence  aspects of two variants of Alternating Direction Method of Multipliers (ADMM): two-block and multi-block. For many optimization problems associated with data mining and machine learning models, for example,  both versions can be used. Given the fact that two-block ADMM has been widely used in the last decade, it has continued to be the method of choice without much thought regarding the use of alternative multi-block ADMM options. For convex problems, they are supposed to give the same solution after all, with suitable choices of (dual) step sizes. However, a question that arises is whether in practice the two-block ADMM is comparable/competitive with multi-block ADMM solving an equivalent problem.

We studied this question in the context of optimization problems arising from multitask learning (MTL) with focus on modeling Alzheimer's disease (AD) progression, Parkinson's disease assessment, and air quality prediction. For all three problems, multi-block
ADMM showed superior performance when compared to the two-block counterpart, both in terms of convergence stability and prediction power.

Although multi-block ADMM requires further theoretical investigations, we surprisingly showed that in practice it is indeed a candidate to be considered when solving large convex optimization problems if using ADMM. Therefore, we suggest that researchers and practitioners to try out the equivalent multi-block versions of the two-block ADMM and then compare their performance.

Additional, both theoretical and empirical, results for multi-block ADMM are still necessary to precisely characterize the convergence and performance of multi-block ADMM, and these are the future research steps.

\bibliographystyle{unsrt}
\bibliography{references}

\end{document}